\documentclass[10pt, twocolumn]{article}

\usepackage[verbose=true,letterpaper]{geometry}
\AtBeginDocument{
  \newgeometry{
    textheight=9in,
    textwidth=6in,
    top=1in,
    headheight=20pt,
    headsep=15pt,
    footskip=15pt
  }
}
\usepackage[compact]{titlesec}         
\titlespacing{\section}{4pt}{4pt}{4pt}

\usepackage{graphicx} 
\usepackage{comment}
\usepackage{subcaption}

\usepackage[utf8]{inputenc} 
\usepackage[T1]{fontenc}    
\usepackage{hyperref}       
\usepackage{url}            
\usepackage{booktabs}       
\usepackage{amsfonts}       
\usepackage{nicefrac}       
\usepackage{microtype}      
\usepackage{float}

\usepackage{authblk}
\usepackage{enumitem}
\usepackage{cuted}
\usepackage{graphicx}%
\usepackage{lmodern}
\usepackage{anyfontsize}
\usepackage{amsmath,amssymb,amsfonts}%
\usepackage{amsthm}%
\usepackage{mathrsfs}%
\usepackage[title]{appendix}%
\usepackage{textcomp}%
\usepackage{manyfoot}%
\usepackage{booktabs}%
\usepackage{algorithm}%
\usepackage{algorithmicx}%
\usepackage{algpseudocode}%
\usepackage{listings}%
\usepackage{booktabs}\let\cline\cmidrule
\usepackage{multirow}
\usepackage{adjustbox}
\usepackage{rotating}
\usepackage{tabularx}
\usepackage[table,xcdraw]{xcolor}
\usepackage{cleveref}
\crefformat{section}{\S#2#1#3} 
\crefformat{subsection}{\S#2#1#3}
\crefformat{subsubsection}{\S#2#1#3}

\makeatletter
\renewenvironment{abstract}{%
    \if@twocolumn
      \section*{\abstractname}%
    \else 
      \begin{center}%
        {\bfseries \Large\abstractname\vspace{\z@}}
      \end{center}%
      \quotation
    \fi}
    {\if@twocolumn\else\endquotation\fi}
\makeatother

\newcommand{\AM}{TSFM}
\newcommand{\Model}{MMRPhys}

\usepackage{environ}
\newcommand{\acksection}{\section*{Acknowledgments and Disclosure of Funding}}
\NewEnviron{ack}{%
  \acksection
  \BODY
}

\title{
\vspace{-20pt}
\noindent\makebox[\linewidth]{\rule{\linewidth}{2pt}}
\textbf{Efficient and Robust Multidimensional Attention in Remote Physiological Sensing through Target Signal Constrained Factorization}
\noindent\makebox[\linewidth]{\rule{\linewidth}{1pt}}
}

\author[1]{\textbf{Jitesh~Joshi}}
\author[1]{\textbf{Youngjun~Cho}}
\affil[1]{\normalsize{Department of Computer Science, University College London, UK}}
\affil[ ]{\{jitesh.joshi.20, youngjun.cho\}@ucl.ac.uk}
\date{}

\begin{document}
\setlength\abovedisplayskip{3pt}
\setlength\belowdisplayskip{3pt}

\maketitle

\begin{strip}
    \centering
    \begin{minipage}{0.85\textwidth}
        \begin{abstract}
            Remote physiological sensing using camera-based technologies offers transformative potential for non-invasive vital sign monitoring across healthcare and human-computer interaction domains. Although deep learning approaches have advanced the extraction of physiological signals from video data, existing methods have not been sufficiently assessed for their robustness to domain shifts. These shifts in remote physiological sensing include variations in ambient conditions, camera specifications, head movements, facial poses, and physiological states which often impact real-world performance significantly. Cross-dataset evaluation provides an objective measure to assess generalization capabilities across these domain shifts. We introduce Target Signal Constrained Factorization module (\AM{}), a novel multidimensional attention mechanism that explicitly incorporates physiological signal characteristics as factorization constraints, allowing more precise feature extraction. Building on this innovation, we present \Model{}, an efficient dual-branch 3D-CNN architecture designed for simultaneous multitask estimation of photoplethysmography (rPPG) and respiratory (rRSP) signals from multimodal RGB and thermal video inputs. Through comprehensive cross-dataset evaluation on five benchmark datasets, we demonstrate that \Model{} with \AM{} significantly outperforms state-of-the-art methods in generalization across domain shifts for rPPG and rRSP estimation, while maintaining a minimal inference latency suitable for real-time applications. Our approach establishes new benchmarks for robust multitask and multimodal physiological sensing and offers a computationally efficient framework for practical deployment in unconstrained environments. The web browser-based application featuring on-device real-time inference of \Model{} model is available at \url{https://physiologicailab.github.io/mmrphys-live/}.
        \end{abstract}
        \hspace{2cm}
    \end{minipage}
\end{strip}

\section{Introduction}\label{sec intro}
Remote photoplethysmography (rPPG), a technique that captures changes in peripheral blood volume using RGB video cameras, has emerged as a prominent method for contactless cardiac monitoring. Remote estimation of other physiological signals, such as respiratory rate (rRSP) \cite{cho2017robust, iozza2019monitoring, khanam2024integratingrgbt} and blood pressure (rSBP and rDBP) \cite{van2023improving, curran2023camera}, is gaining increasing interest due to its promising applications in healthcare, human-computer interaction, and remote monitoring, where wearable or contact-based devices may be uncomfortable or intrusive \cite{sureshkumar2021review, ding2022noncontact}. 

Recent rPPG approaches have seen significant advances, from EfficientPhys \cite{liu2023efficientphys} and PhysFormer \cite{yu2022physformer} to more recent innovations like PhysFormer++ \cite{yu2023physformer}, TransPhys \cite{wang2023transphys}, RADIANT \cite{gupta2023radiant} and Contrast-Phys+ \cite{sun2024contrast}. Although transformer-based architectures have gained prominence, comparative evaluations of Swin-transformer-based \cite{liu2021swin} and CNN-based EfficientPhys \cite{liu2023efficientphys} demonstrated that the latter achieves superior accuracy with significantly lower latency, making it more suitable for mobile deployment. 

For estimating rRSP, researchers have explored various approaches \cite{liu2019recent}, including RGB imaging-based methods that derive respiratory signals from modulation of rPPG \cite{van2016robust} or motion induced by breathing \cite{iozza2019monitoring, sanyal2018algorithms, wang2022algorithmic}. Thermal infrared imaging methods track airflow-induced temperature fluctuations around the nostril and mouth regions \cite{zhu2005tracking, cho2017deepbreath, cho2017robust}, while multispectral approaches combine near- and far-infrared imaging to track multiple modulations related to breathing \cite{scebba2020multispectral}. Although RGB methods offer a wider applicability \cite{van2016robust}, thermal approaches often demonstrate superior robustness \cite{cho2017robust, yang2022graph}. 

MTTS-CAN \cite{liu2020multitask} demonstrated effective multitask learning through an attention module that bridges appearance and motion branches for simultaneous estimation of rPPG and rRSP. Recent work introduced the Wrapping Temporal Shift Module (WTSM) \cite{narayanswamy2024bigsmall} for modeling time-variant signals with small window sizes, built on the Temporal Shift Module (TSM) \cite{lin2019tsm}. The BigSmall architecture, using WTSM blocks, validated a multibranch approach combining low-resolution temporal and high-resolution spatial branches for simultaneous estimation of rPPG, rRSP and facial actions \cite{narayanswamy2024bigsmall}.

Estimating physiological signals from video presents challenges that include changes in illumination, head movements, and variations in skin tone \cite{huang2023challenges}. Attention mechanisms have proven instrumental in mitigating these challenges by enhancing relevant spatial-temporal features \cite{chen2018deepphys}. In recent developments, the Factorized Self-Attention Module (FSAM) \cite{joshi2024factorizephys} has shown significant efficacy in cross-dataset generalization by employing Non-negative Matrix Factorization (NMF) for multidimensional attention within the context of rPPG estimation. Nevertheless, this methodology is not without its limitations: non-negativity constraints, in isolation, do not ensure uniqueness of solutions \cite{wang2012nonnegative} while it is important to consistently maintain an attention on desired set of spatial-temporal features. Also, NMF is predisposed to generating discontinuous factorization components \cite{yokota2015smooth}, thereby rendering it less effective for capturing the smooth temporal characteristics essential to physiological signals.

To overcome these limitations, this work introduces a novel approach that leverages the ground-truth signal as a factorization constraint. This constraint enforces reconstruction of low-rank embeddings to preserve spatial and channel features that demonstrate temporal correlation with the target physiological signal. We demonstrate that this target signal constraint enables attention mechanisms to adapt seamlessly to the characteristics of different physiological signals, eliminating the need for signal-specific attention adaptations. To enable systematic comparison across imaging modalities and spatial resolutions \cite{narayanswamy2024bigsmall, yu2023physformer, liu2020multitask, botina2022rtrppg}, we introduce \Model{}, a family of efficient dual-branch 3D-CNN networks for simultaneous rPPG and rRSP estimation. \Model{} supports single- and multi-task estimation from uni-modal (RGB/thermal) or multi-modal input across various spatial resolutions. Through comprehensive evaluation on five public datasets, we demonstrate exceptional cross-dataset generalization and state-of-the-art accuracy with significantly reduced computational complexity. In summary, we make the following contributions:
\begin{itemize}
    \item \textbf{Target Signal Constrained Factorization Module (\AM{}):} A novel multidimensional attention mechanism that leverages physiological signal characteristics as explicit constraints for non-negative matrix factorization of deep-layer embeddings, enabling more precise and signal-specific feature selection.
    
    \item \textbf{\Model{}:} A family of efficient dual-branch 3D-CNN architectures that integrates \AM{} and supports multi-task learning for simultaneous estimation of rPPG and respiratory signals. This framework accommodates both single-modal (RGB or thermal) and multimodal (combined RGB and thermal) video inputs across various spatial resolutions, optimizing performance while maintaining computational efficiency.
    
    \item \textbf{Comprehensive evaluation:} A rigorous assessment of \Model{} and \AM{} across five benchmark datasets, employing multiple performance metrics with standard error measurements to establish cross-dataset generalization capabilities compared to state-of-the-art methods for rPPG and respiratory signal estimation.
\end{itemize}

\section{Related Works}\label{sec related works}
Remote physiological sensing has seen substantial advances in recent years. This section focuses on attention mechanisms in end-to-end physiological sensing methods and the theoretical foundations of constrained nonnegative matrix factorization that underpin our proposed approach.

\subsection{Attentions Mechanisms in Remote Physiological Sensing}\label{subsec remote phys}
The extraction of physiological signals from facial video frames requires networks to effectively isolate specific temporal characteristics from a multidimensional feature space while mitigating variances associated with head motion, illumination, and skin pigmentation\cite{huang2023challenges, wang2024camerabased}. As highlighted in a recent review on contactless physiological monitoring in clinical settings \cite{huang2023challenges}, these systems remain vulnerable to various disturbances, particularly head movements.

DeepPhys \cite{chen2018deepphys} pioneered attention in rPPG by implementing a \textit{ appearance} branch to derive attention weights for the main \textit{motion} branch. Drawing inspiration from the CBAM attention mechanism \cite{woo2018cbam}, ST-Attention \cite{niu2019robust} extracted salient features from spatial-temporal maps to enhance remote heart rate estimation. A similar dual attention mechanism in the SMP-Net framework \cite{ding2022noncontact} simultaneously learned features from RGB and infrared spatial-temporal maps to estimate multiple physiological signals. Building on this foundation, JAMSNet \cite{zhao2022jamsnet} and GLISNet \cite{zhao2023learning} attempted to devise multidimensional attention by combining attention blocks computed separately across different dimensions. However, this approach still suffered from information loss when squeezing dimensions to compute attention.

In a recent work \cite{shao2024video}, the authors noted that breathing cycles typically span several frames, making it particularly challenging to model long-term spatial-temporal features. Their proposed long-term temporal attention mechanism, implemented in a TSM-based architecture \cite{lin2019tsm}, demonstrated effectiveness in capturing these extended temporal dependencies. Although MTTS-CAN \cite{liu2020multitask} demonstrated the effectiveness of attention as a bridge between appearance and motion branches, it did not implement signal-specific adaptations of the attention mechanism for different physiological signals (rPPG and rRSP). This highlights an important gap in current approaches—the need for attention mechanisms that can adapt to the unique spatial-temporal characteristics of different physiological signals.

More recently, FactorizePhys \cite{joshi2024factorizephys} introduced FSAM, which adapts Non-negative Matrix Factorization (NMF) \cite{lee1999learning} as a multidimensional attention mechanism for rPPG estimation. FSAM works by factorizing voxel embeddings without squeezing any dimension, enabling joint modeling of spatial, temporal, and channel attention. This approach demonstrated state-of-the-art cross-dataset generalization for rPPG estimation. The integration of NMF principles into deep learning architectures has gained increasing attention, as demonstrated by the Hamburger module \cite{geng2021attention}, which employed NMF for global context modeling in image generation and semantic segmentation tasks.

Although non-negativity constraints effectively reveal latent structures \cite{lee1999learning}, they do not guarantee unique solutions (not a single definite result) \cite{wang2012nonnegative}, as the reconstruction of the low-rank matrix depends significantly on the initialization of basis and coefficient matrices. When motion or illumination variations in the spatial-temporal features dominate the subtle modulations of the physiological signal, the low-rank approximation inherently favors these stronger artifacts because they contribute more to minimizing the overall approximation error. Furthermore, FSAM \cite{joshi2024factorizephys} fails to account for the inherently smooth temporal characteristics of physiological signals, and standard NMF frequently produces discontinuous factorization components \cite{yokota2015smooth}, compromising its effectiveness for reliable multidimensional attention computation. These fundamental limitations underscore the necessity for specialized constraints specifically designed to capture the temporal dynamics of physiological signals while deriving multidimensional attention through factorization. To provide context for FSAM \cite{joshi2024factorizephys} and our proposed \AM{}, we next examine the theoretical principles and inherent limitations of NMF that have informed our approach.

\subsection{Non-negative Matrix Factorization (NMF)}\label{subsec constrained nmf}
NMF \cite{paatero1994positive, lee1999learning} has proven to be a powerful dimensionality reduction method that can infer latent structures in multivariate data \cite{lee1999learning, du2023matrix}. The non-negativity constraint prevents mutual cancelation between basis functions, leading to interpretable parts-based representations \cite{lee1999learning}. NMF has been explored in different tasks including blind source separation (BSS) in images and nonnegative signals \cite{berry2007algorithms}, spectral recovery \cite{sajda2004nonnegative}, feature extraction and clustering \cite{ding2006orthogonal}. 

Existing surveys on NMF methods summarize and categorize various approaches, including Deep NMF methods, that have gained attention among researchers in machine learning, signal processing, and statistics \cite{wang2012nonnegative, du2023matrix}. These Deep NMF methods combine NMF and multilayer neural networks to use network forward propagation to derive the iterative formulas of basis and feature matrices \cite{chen2022survey}. As our goal is to use matrix factorization to compute attention, rather than evaluating or advancing the state of solving NMF through deep neural networks \cite{chen2022survey}, we use the multiplicative update method \cite{geng2021attention} to derive a low-rank matrix. This approach, also used in FSAM \cite{joshi2024factorizephys}, offers distinct advantages for physiological signal extraction: it maintains computational efficiency during both training and inference, allows for explicit incorporation of signal-specific constraints without requiring end-to-end retraining, and provides better interpretability of the attention mechanism through direct access to factorization components. Moreover, using multiplicative updates ensures stable convergence for the specific constraints we introduce, whereas Deep NMF methods might struggle with the integration of target signal constraints due to their more complex optimization dynamics and potential instability when backpropagating through the factorization process.

\subsubsection{Limitations of Standard NMF}\label{review nmf limitations}
Although the standard NMF is proficient in uncovering latent structures, it does not guarantee unique solutions \cite{wang2012nonnegative}, since the reconstructed low-rank matrix depends on varying initializations of the basis and coefficient matrices, which are necessary to resolve factorization using iterative updates. Furthermore, it may produce factorization components that lack continuity, a significant limitation in applications that require smooth transitions between data points \cite{yokota2015smooth}. These limitations are particularly relevant for the extraction of physiological signals, where temporal continuity is an essential characteristic.

\subsubsection{Constrained NMF to Address Limitations}
To remedy these limitations, researchers have introduced additional constraints such as regularization terms. These include sparseness constraints \cite{hoyer2004non}, orthogonality constraints \cite{ding2006orthogonal}, and smoothness constraints \cite{yokota2015smooth}. For example, Ye et al. \cite{ye2019blind} demonstrated improved heart rate extraction from continuous wave Doppler radar by applying sparseness constraints through sparse NMF (SPNMF) and weighted sparse NMF (WSPNMF) algorithms.

Smooth NMF (SNMF) \cite{yokota2015smooth} imposes smoothness constraints using Gaussian radial basis functions (GRBFs) and has shown significantly better blind source separation performance than conventional NMF methods across various noise levels in both synthetic and real-world data. This approach is particularly relevant for physiological signals, which inherently possess temporal smoothness characteristics. Although these constrained approaches offer improvements over standard NMF, they rely on generic constraints that are not specifically tailored to the characteristics of the signals being extracted. This presents an opportunity to explore signal-specific constraints that could enhance attention mechanisms for physiological signal extraction.

Although prior work has explored various applications of NMF in computer vision and signal processing, the potential of integrating physiological signal characteristics directly into factorization constraints for multidimensional attention mechanisms remains unexplored. Current SOTA methods such as FSAM \cite{joshi2024factorizephys} focus on general non-negativity constraints without considering the unique temporal and physiological characteristics of specific biosignals. Our proposed Target Signal Constrained Factorization module (\AM{}) addresses this gap by explicitly incorporating the properties of the physiological signal of the target as factorization constraints. This novel integration establishes a direct mathematical relationship between the attention mechanism and the physiological signal being extracted, enabling the network to be inherently more selective toward features that correlate with the underlying physiological processes rather than relying solely on general nonnegativity properties.

\section{Methods}\label{sec methods}
In contrast to the Transformers-based self-attention mechanism \cite{vaswani2017attention}, the NMF-based attention mechanism \cite{lee1999learning}, as implemented by FSAM \cite{joshi2024factorizephys}, can be regarded as a \textit{factorize and excite} approach, drawing an informal parallel from the \textit{squeeze and excite} attention mechanism \cite{hu2018squeeze}. For a comprehensive introduction to NMF, the reader is directed to this seminal work \cite{lee1999learning}, along with previous research on the Hamburger module \cite{geng2021attention}, which conceptualized NMF as a global context block and contributed to an efficient approximation method employing multiplicative updates to address the factorization problem. We further elucidate the formulation of FSAM \cite{joshi2024factorizephys}, which employs standard NMF to derive the multidimensional attention mechanism, in \cref{method primer fsam}. Moreover, given the potential of smoothness-constrained NMF \cite{yokota2015smooth, zdunek2012approximation} to address certain inherent limitations associated with NMF-based attention mechanisms (\cref{review nmf limitations}), we provide an overview of the formulation of GRBF-based smooth NMF in \cref{method primer smooth nmf}. Building upon these established attention mechanisms and their limitations, we now introduce our novel Target Signal Constrained Factorization (\AM{}) module, to enhance multidimensional attention by explicitly incorporating physiological signal characteristics into the factorization process.

\subsection{Target Signal Constrained Factorization (TSFM): Proposed Attention Module}\label{method tsfm}
\begin{figure}
    \centering
    \includegraphics[width=1.0\linewidth]{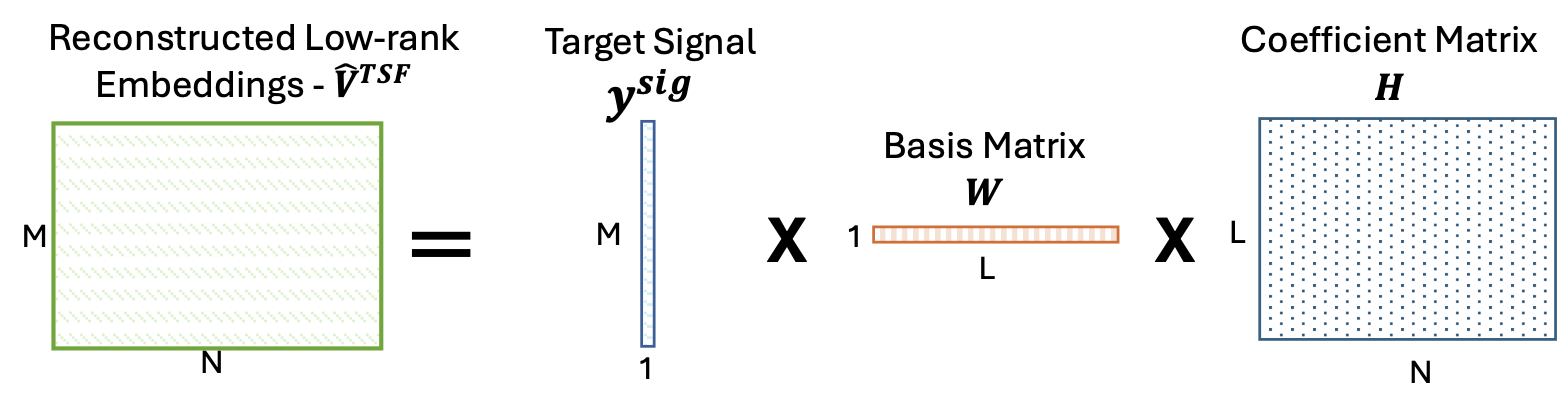}
    \caption{Proposed Target Signal Constrained Factorization}
    \label{fig:tsfm}
\end{figure}

\begin{figure*}[ht]
    \centering
    \includegraphics[width=1.0\linewidth]{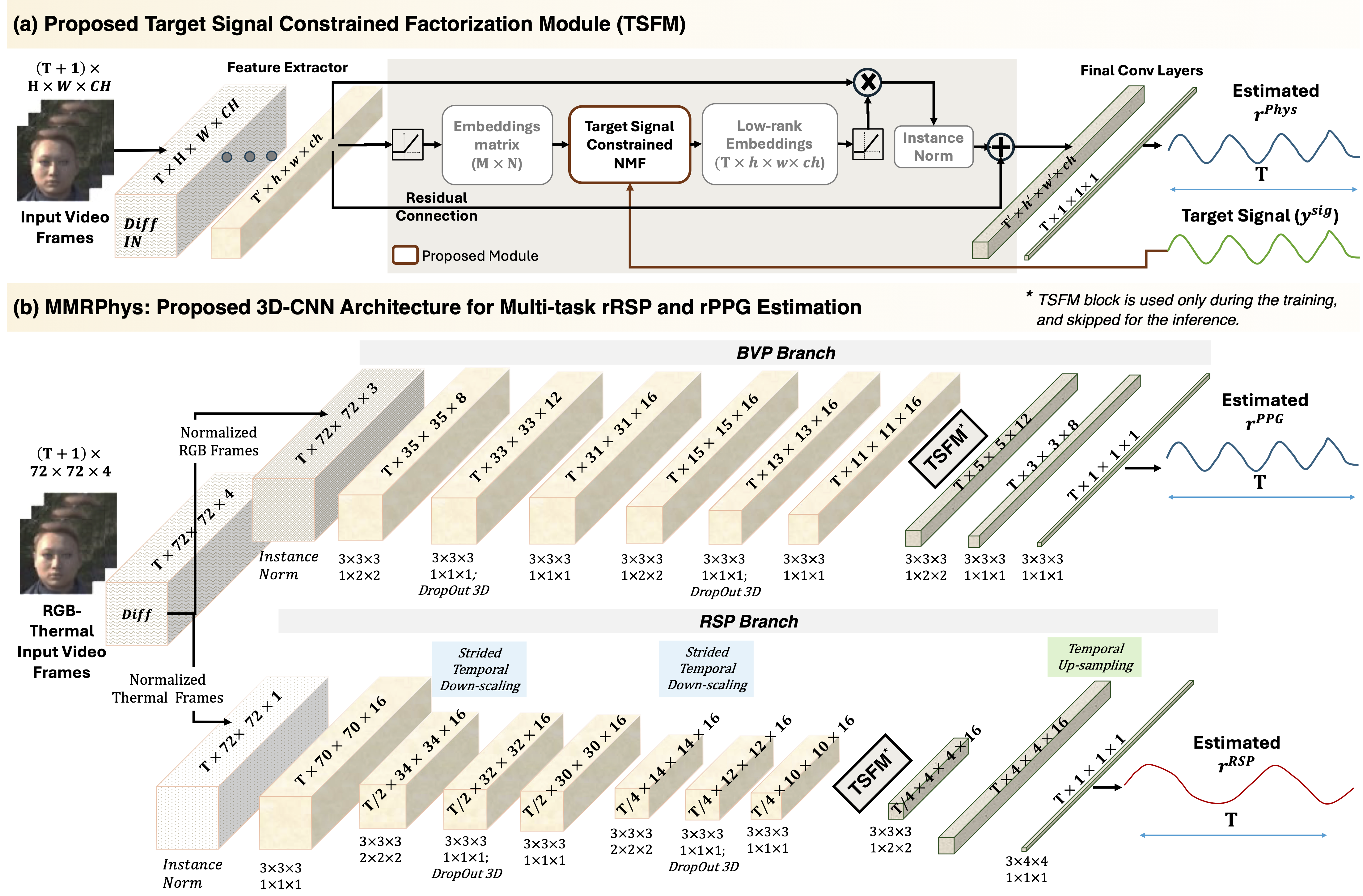}
    \caption{Overview of the proposed TSFM module and MMRPhys architecture. [a]: Deployment of the proposed Target Signal Constrained Factorization module for remote physiological sensing, [b]: MMRPhys: proposed dual-branch 3D-CNN architecture for simultaneous estimation of rPPG and rRSP.}
    \label{fig:MMRPhys}
\end{figure*}

First, we emphasize the importance of mapping the temporal features of \(\varepsilon \in \mathbb{R}^{\tau \times \kappa \times \alpha \times \beta}\) with the vector dimension of \(V^{st}\) (that is, \(\tau \mapsto M\)) (equation \ref{eq 3d mapping}). This is not only documented in FSAM \cite{joshi2024factorizephys} as the optimal mapping, but is also essential to implement suitable factorization constraints that reflect the temporal attributes of the physiological signal to be estimated. Moreover, we underscore previously reported results regarding multidimensional factorization-based attention, which facilitated the satisfactory performance of a 3D-CNN architecture, FactorizePhys \cite{joshi2024factorizephys}, even when the attention module was omitted during inference. In view of the observations and considerations mentioned above, we perform factorization of 2-D transformed embeddings \(V \in \mathbb{R}^{M \times N}\) (equation \ref{eq 3d mapping}), constrained by the target signal, to compute low-rank embeddings matrix, as depicted in Fig. \ref{fig:tsfm} (equation \ref{eq tsfm})

\begin{equation}\label{eq tsfm}
    \hat{V}^{TSF} = y^{sig} W H 
\end{equation}
where \(y^{sig}\) can be rPPG, rRSP or any physiological signal that is to be estimated as a downstream task. \(\hat{V}^{TSF}\) is transformed back to the embedding dimension (similar to equation \ref{eq appx embeddings}), which is then multiplied by the original embedding (equation \ref{eq appx embeddings TSFM b}) as depicted in Fig. \ref{fig:MMRPhys}[a].
\begin{align}
\label{eq appx embeddings TSFM a}
& \hat{\varepsilon}^{TSF} = \xi_{post}(\Gamma^{MN \mapsto \tau\kappa\alpha\beta}( \hat{V}^{TSF}  \in \mathbb{R}^{M \times N}))\\
\label{eq appx embeddings TSFM b}
& r^{phys} = \omega(\varepsilon + \mathcal{IN}(\varepsilon \odot \hat{\varepsilon}^{TSF}))
\end{align}

We refer to this novel multidimensional attention module as the target signal-constrained factorization module (\AM{}). The prevailing optimization algorithm to solve the factorization constraint by smoothness employs the iterative update rule \cite{yokota2015smooth}, which, as noted in \cite{geng2021attention}, can lead to an unstable gradient. In this work, we adapt the multiplicative update method, initially introduced to solve NMF within the Hamburger module \cite{geng2021attention}. This method functions as a linear approximation of the conventional backpropagation through time algorithm \cite{werbos1990backpropagation}, as elucidated within the framework of Proposition 2 when time \(t \rightarrow \infty\)  \cite{geng2021attention}. We implement this multiplicative update method to solve the constrained factorization in three to four iterations within \textit{no\_grad} block of PyTorch \cite{paszke2019pytorch}.

Figure \ref{fig:MMRPhys}[a] illustrates the deployment of \AM{} within a generic 3D-CNN framework that aims to estimate remote physiological signals from video frames. We integrate \AM{} and the GRBF constraint-based attention module \cite{yokota2015smooth} into an existing SOTA rPPG architecture, FactorizePhys \cite{joshi2024factorizephys}, to conduct a comprehensive comparison between FSAM \cite{joshi2024factorizephys}, \AM{} and the GRBF constraint-based attention module \cite{yokota2015smooth}. Furthermore, we integrate \AM{} into our proposed \Model{} model, which we elaborate in the following subsection.

\subsection{MMRPhys: Proposed Multi-task Architecture}\label{MMRPhys}
With our novel TSFM, this work introduces \Model{}, a suite of efficient 3D-CNN architectures, designed to support multi-task learning, which allows the simultaneous estimation of rPPG and rRSP signals. \Model{} builds upon the core 3D-CNN architecture of FactorizePhys \cite{joshi2024factorizephys}, maintaining its attention mechanism integration while optimizing the convolutional backbone for reduced parameters and inference latency. \Model{} extends this optimized foundation through a dual-branch design that enables processing of both single modal (either RGB or thermal infrared) and multimodal video frames that combine RGB and thermal data. Furthermore, \Model{} can be configured to support multiple spatial resolutions, allowing systematic investigation of performance between various types of input data.

Figure \ref{fig:MMRPhys}[b] introduces a configuration of \Model{} that utilizes both combined RGB and thermal data as input to estimate rPPG and rRSP signals within a multi-task learning framework. Alternative configurations of \Model{} can be realized by specifying the appropriate parameters in the configuration files included in the repository. Specifically, \Model{} supports three different spatial resolutions of the input video frames, which include \(T \times 72 \times 72 \times Ch\), \(T \times 36 \times 36 \times Ch\), and \(T \times 9 \times 9 \times Ch\), where \(T\) can be any number of frames (typically \(\geq 10s\), and \(Ch \in [1, 3, 4]\) to support RGB, thermal, and combined RGB-thermal input video frames. The support for various other spatial resolution can be easily extended with minor adaptations of the existing models.

The spatial distribution of rPPG and rRSP signals across facial video frames exhibits significant differences. Specifically, the rPPG signal is expected to be distributed across the entire skin surface in the facial regions, whereas the sources of the rRSP signal vary between RGB and thermal infrared modalities. In thermal infrared video frames, notable rRSP signals can be derived from the temperature modulation associated with respiration at the nostrils. In contrast, within RGB video frames, the rRSP signal can be obtained from head movements related to breathing or from the envelope of amplitude modulation of rPPG peaks.

In contrast to rPPG signals, the temporal modulation of rRSP signals manifests at a significantly slower rate, with typical respiratory rates ranging from 12 to 18 breaths per minute in healthy adults, as opposed to the normal resting heart rate, which ranges from 60 to 100 beats per minute for a healthy adult \cite{medlineplusvitalsigns}. In light of these distinctions, we have designed \Model{} as a dual-branch architecture comprising \textit{BVP} and \textit{RSP} branches, each dedicated to the estimation of rPPG and rRSP signals, respectively. Analogous to FactorizePhys \cite{joshi2024factorizephys}, the \textit{BVP} branch of \Model{} aggregates the spatial dimension, while maintaining the temporal dimension throughout the branch; however, we have optimized the model parameters and FLOPS to further minimize inference time latency.

The \textit{RSP} branch of \Model{} employs convolutional strides across both spatial and temporal dimensions, effectively capturing slow temporal modulations while providing an enhanced temporal receptive field. In the final layers, the temporal dimension is upscaled to ensure alignment with the temporal dimension of the estimated rRSP signal. Both the \textit{BVP} and \textit{RSP} branches of \Model{} are architected to be fully convolutional in the temporal dimension, allowing for a variable number of input frames during both model training and inference. 

\section{Experiments}\label{sec exp overview}

\subsection{Overview of Evaluation Strategy}\label{evaluation strategy}
We conduct an extensive evaluation of \Model{} and \AM{} using multiple metrics and associated standard error measurements to compare cross-dataset generalization performance against SOTA rPPG and rRSP methods, using five benchmarking datasets \cref{subsec exp dataset}. Our implementation uses PyTorch framework \cite{paszke2019pytorch} and builds upon rPPG-Toolbox \cite{liu2024rppg} . We adhere to the pre-processing, model training, and evaluation stages for the rPPG estimation task as specified in FactorizePhys \cite{joshi2024factorizephys}. Furthermore, our implementation of \AM{} and GRBF based attention module adapts FSAM \cite{joshi2024factorizephys}, to apply the respective constraints on the factorization. We further elaborate on model training for rPPG and rRSP, along with the optimization objective in \cref{subsec exp implementation}.

The evaluation is organized into three distinct parts: i) An ablation study aimed at providing a thorough comparison between NMF-based FSAM \cite{joshi2024factorizephys} and NMF-based attention modules incorporating target signal and GRBFs as constraints; ii) An evaluation under a single modality and single task setting employing rPPG datasets and benchmarks; and iii) A comprehensive assessment within a multimodal and multi-task context utilizing both rPPG and rRSP datasets and associated benchmarks.

Our ablation study emphasizes cross-dataset generalization performance, as disentangling the confounding effects of model overfitting poses significant challenges in within-dataset evaluations. In contrast, cross-dataset can reliably assess the effectiveness of multidimensional attention in learning the relevant spatial-temporal features, which we report in Table \ref{results ablation}.

For single-task and single-modality performance evaluation, we employ RGB-based rPPG estimation, a well-explored area which facilitates rigorous comparison with the established benchmarks. We meticulously select SOTA end-to-end rPPG methods that exemplify diverse architectural frameworks, including 2D-CNN, 3D-CNN, and Transformer-based models, alongside key attention mechanisms. The selected rPPG SOTA methods encompass PhysNet ~\cite{yu2019remote}, EfficientPhys~\cite{liu2023efficientphys}, PhysFormer~\cite{yu2022physformer}, and FactorizePhys \cite{joshi2024factorizephys}. We report this evaluation in Table \ref{results single modality single task}

Finally, a thorough evaluation is performed using multimodal video inputs comprising RGB and thermal frames within a multi-task framework to concurrently estimate rPPG and rRSP signals. The analysis primarily involves a comparison of the performance between \Model{} with \AM{} and the BigSmall \cite{narayanswamy2024bigsmall} model, a more recent state-of-the-art multi-task model that exclusively utilizes RGB frames for the estimation of rPPG and rRSP signals. To ensure a fair evaluation, we include the multi-task evaluation \Model{}, when it is trained exclusively with RGB video frames.

Additionally, considering that the BigSmall \cite{narayanswamy2024bigsmall} model is a dual-branch network, adaptations are made to enable it to accept RGB and thermal video frames interchangeably into its \textit{Small} and \textit{Big} branches. For comprehensiveness, \Model{} and BigSmall \cite{narayanswamy2024bigsmall} are trained for multi-task estimation of rPPG and rRSP signals when relying solely on thermal video frames as input. Furthermore, for all experiments, \Model{} is additionally trained utilizing FSAM \cite{joshi2024factorizephys} to compare the efficacy of our proposed approach \AM{} in the context of multi-task learning. 

Comprehensive results pertaining to both within-dataset and cross-dataset evaluations are presented in \cref{results multi modality multi task}. It is important to note that of the two datasets \cite{mcduff2022scamps, zhang2016multimodal} providing rPPG and rRSP ground-truth signals to facilitate multi-task learning, only one \cite{zhang2016multimodal} features multimodal input video frames. Consequently, our cross-dataset multi-task performance evaluation could only be conducted using a single modality (RGB). 

\subsection{Datasets}\label{subsec exp dataset}
We report our findings utilizing five benchmark datasets. Each of these datasets provides ground-truth rPPG signals. Additionally, two datasets also contain ground-truth rRSP, thereby facilitating multi-task evaluation. Moreover, all datasets comprise RGB video frames, while two of these datasets \cite{zhang2016multimodal,joshi2024ibvp} additionally provide thermal video frames. Subsequently, we provide a brief description of each dataset.

\textbf{BP4D+:} [RGB, Thermal] $\mid$ [rPPG, rRSP]\\
The BP4D+ dataset \cite{zhang2016multimodal, ertugrul2019cross} comprises 1400 RGB and thermal videos of varying duration, recorded at a rate of 25 frames per second (FPS) from 140 participants, each of whom engaged in 10 emotion-inducing tasks. The dataset offers ground-truth data encompassing blood pressure measurements, including waveforms, systolic, diastolic, and mean values, which were acquired using a finger unit of the non-invasive blood pressure monitoring system, Biopac NIBP100D. Additionally, it provides heart rate derived from blood pressure waveforms, respiration data encompassing both wave and rate, electrodermal activity, and Facial Action Coding System (FACS) encodings. This dataset is utilized for comprehensive multimodal and multi-task evaluation of \Model{}, facilitating concurrent estimation of rPPG and rRSP signals.

\textbf{iBVP:} [RGB, Thermal] $\mid$ [rPPG]\\
The iBVP dataset \cite{joshi2024ibvp} comprises 124 synchronized videos in RGB and thermal infrared modalities, featuring 31 subjects participating under four different conditions, which involve controlled breathing, mathematical tasks, and head movements. Each video spans 180 seconds, accompanied by corresponding ground-truth BVP signals captured from the participant's ear utilizing PhysioKit \cite{joshi2023physiokit}. Furthermore, the signal quality indicator for the ground-truth BVP signals facilitates the exclusion of noisy segments in both video frames and the BVP signal. Although this dataset offers synchronized thermal video frames, it does not contain ground-truth rRSP signals, owing to which we employ iBVP dataset primarily for the evaluation of RGB-based rPPG, while we use its combined RGB-Thermal frames for qualitative cross-dataset analysis of models trained with RGB-thermal frames from the BP4D + data set \cite{zhang2016multimodal, ertugrul2019cross}.

\textbf{PURE:} [RGB] $\mid$ [rPPG]\\
The PURE dataset \cite{stricker2014non} comprises 60 video recordings captured with diverse motion conditions, accompanied by ground-truth BVP signals and SpO\textsc{2} measurements from 10 subjects. This dataset serves the purpose of evaluating the single-task estimation of rPPG signals from RGB frames.

\textbf{SCAMPS:} [RGB] $\mid$ [rPPG, rRSP]\\
The SCAMPS dataset \cite{mcduff2022scamps} uniquely offers 2800, 60 seconds videos of synthetic avatars, generated using high-fidelity, quasi-photorealistic renderings. This dataset is utilized for comprehensive multi-task evaluation of \Model{}, facilitating concurrent estimation of rPPG and rRSP signals from RGB frames.

\textbf{UBFC-rPPG:} [RGB] $\mid$ [rPPG]\\
The UBFC-rPPG dataset \cite{bobbia2019unsupervised} provides video recordings alongside ground-truth BVP signals collected from 43 participants. The videos were captured in controlled indoor settings, utilizing a combination of natural sunlight and artificial lighting. This dataset serves the purpose of evaluating the single-task estimation of rPPG signals from RGB frames.

\subsection{Implementation Details}\label{subsec exp implementation}
For single-task estimation for rPPG signals, all models were meticulously trained for 10 epochs utilizing uniformly preprocessed video frames alongside their corresponding ground-truth signals. The input video dimension was maintained consistently as \(160\times72\times72\), representing 160 total frames, and a spatial resolution of \(72\times72\). In multi-task context, the BigSmall \cite{narayanswamy2024bigsmall} uses dual video streams at \(180\times9\times9\) and \(180\times144\times144\) resolutions to infer rRSP, rPPG, and FACS. As our work excludes FACS, we use \(180\times9\times9\) and \(180\times72\times72\) resolution frames for \textit{Small} and \textit{Big} branches, modifying the first layer of the \textit{Big} branch accordingly. 

Since our proposed method \Model{} necessitates only a single stream of video input, we conducted distinct experiments utilizing \(180\times9\times9\), \(180\times36\times36\), and \(180\times72\times72\). For evaluation purposes, the inferred signal segments were amalgamated to create an extended continuous segment, equating to the lesser of 30 seconds or the maximum length of an individual video clip (\(>> 180\) frames). This approach ensures sufficient frequency resolution for the derivation of FFT peak-based HR and RR metrics.

Similar to the training protocol of SOTA for the multi-task learning \cite{narayanswamy2024bigsmall}, we train all models on BP4D+ dataset \cite{zhang2016multimodal, ertugrul2019cross} with 5 epochs. In particular, for training models on the SCAMPS dataset \cite{mcduff2022scamps}, an adaptation to this protocol was made for single- and multi-task estimations, where training was confined to a single epoch to avert potential overfitting \cite{joshi2024factorizephys}. To optimize the models for a single rPPG estimation task, we use the negative Pearson correlation (\(\eta_{p}\)) as a loss function, which has been widely used for end-to-end remote physiological sensing tasks \cite{liu2024rppg, joshi2024factorizephys}.

In the context of optimizing models for multi-task learning, the application of \(\eta_{p}\) as the optimization objective led to a numerical imbalance between the rPPG and rRSP signals. This is attributable to the distinct prominence of one signal over the other within each modality. To address these empirical observations, it is crucial to allocate weights judiciously to the individual loss terms associated with the estimation of RPPG and rRSP. 

Given the complexity of determining optimal weights, prior research on multi-task learning \cite{narayanswamy2024bigsmall} has employed mean-squared-error (\(MSE\)) loss. Our empirical analysis compared \(MSE\) and \(Smooth~L1\) loss, with the latter demonstrating superior convergence across all multi-task models. Therefore, we adopted \(Smooth~L1\) loss, as expressed in equation \ref{smooth l1}, for the training of all multi-task models.
\begin{equation}
\label{smooth l1}
l_i = \begin{cases}
        0.5 (x_i - y_i)^2, & \text{if } |x_i - y_i| < 1.0 \\
        |x_i - y_i| - 0.5, & \text{otherwise }
        \end{cases}    
\end{equation}

All models were trained on a quad-GPU workstation with 4 \(\times\) NVIDIA GeForce RTX 3090 24GB XC3 Ampere Graphics Cards, 8 \(\times\) 16GB ECC DDR4 3200 MHz Server RAM and AMD Ryzen Threadripper PRO 3975WX 32 Core WRX8 CPU, hosting Ubuntu 20.04.6 OS. To assess the latency for model inference, we employed a laptop with single NVIDIA GeForce RTX 3070 GPU, 16GB RAM, and an Intel Core i7-10870H processor, hosting Ubuntu 22.04 OS.

\section{Results and Discussion}\label{sec results}
To assess the performance of estimated rPPG signals, we utilize heart rate (HR)~\cite{xiao2024remote} in conjunction with blood volume pulse (BVP) metrics, specifically the signal-to-noise ratio (SNR) and the maximum amplitude of cross-correlation (MACC) \cite{joshi2024ibvp, cho2017robust}. For the assessment of rRSP signal estimation, respiratory rate (RR) is employed alongside SNR and MACC metrics particularly adapted for the RSP signals. SNR and MACC constitute direct measures for comparing estimated rPPG and rRSP signals to the respective ground-truth signals. The metrics that we report for HR and RR include the mean absolute error (MAE), the root mean square error (RMSE), the mean absolute percentage error (MAPE), and Pearson's correlation coefficient (Corr)~\cite{xiao2024remote} in relation to estimated HR and RR. Estimates of uncertainty, which reflect the variability associated with signal estimation, have been shown to correlate strongly with the absolute error of the estimated HR \cite{song2023uncertainty}. In addition, for each of these metrics, the standard error is provided to assess the variability of each model.

\subsection{Ablation to Evaluate NMF Constraints}\label{results ablation}
We perform ablation with FactorizePhys \cite{joshi2024factorizephys}, using only rPPG datasets, to fairly compare the performance of the baseline model and different types of constraints applied on NMF. For this, we train FactorizePhys\cite{joshi2024factorizephys} model on UBFC-rPPG \cite{bobbia2019unsupervised} dataset and report the cross-dataset performance on two different datasets, viz. PURE \cite{stricker2014non} and iBVP \cite{joshi2024ibvp} in Table \ref{tab:results ablation}. 

\begin{table*}[ht]
\caption{Ablation study with rPPG datasets to compare different constraints on NMF}
\label{tab:results ablation}
\centering
\fontsize{6}{4}\selectfont
\setlength{\tabcolsep}{1pt}
\renewcommand{\arraystretch}{1.8}
\begin{tabular*}{\textwidth}{@{\extracolsep\fill}ccclcccccccccccc}
\hline
 &
   &
   &
  \multicolumn{1}{c}{} &
  \multicolumn{2}{c}{\textbf{MAE ↓}} &
  \multicolumn{2}{c}{\textbf{RMSE ↓}} &
  \multicolumn{2}{c}{\textbf{MAPE ↓}} &
  \multicolumn{2}{c}{\textbf{Corr ↑}} &
  \multicolumn{2}{c}{\textbf{SNR ↑}} &
  \multicolumn{2}{c}{\textbf{MACC ↑}} \\ \cline{5-16} 
\multirow{-2}{*}{\textbf{\begin{tabular}[c]{@{}c@{}}Training\\ Dataset\end{tabular}}} &
  \multirow{-2}{*}{\textbf{\begin{tabular}[c]{@{}c@{}}Testing\\ Dataset\end{tabular}}} &
  \multirow{-2}{*}{\textbf{\begin{tabular}[c]{@{}c@{}}Attention\\ Block\end{tabular}}} &
  \multicolumn{1}{c}{\multirow{-2}{*}{\textbf{\begin{tabular}[c]{@{}c@{}}Constraint\\ Type\end{tabular}}}} &
  \textbf{Avg} &
  {\color[HTML]{808080} \textbf{SE}} &
  \textbf{Avg} &
  {\color[HTML]{808080} \textbf{SE}} &
  \textbf{Avg} &
  {\color[HTML]{808080} \textbf{SE}} &
  \textbf{Avg} &
  {\color[HTML]{808080} \textbf{SE}} &
  \textbf{Avg} &
  {\color[HTML]{808080} \textbf{SE}} &
  \textbf{Avg} &
  {\color[HTML]{808080} \textbf{SE}} \\ \hline \hline
 &
   &
  - &
  NA &
  1.37 &
  {\color[HTML]{808080} 1.02} &
  7.97 &
  {\color[HTML]{808080} 2.84} &
  2.55 &
  {\color[HTML]{808080} 2.07} &
  0.94 &
  {\color[HTML]{808080} 0.04} &
  13.74 &
  {\color[HTML]{808080} 0.81} &
  0.77 &
  {\color[HTML]{808080} 0.02} \\ \cline{3-16} 
 &
   &
  \checkmark &
  None &
  0.48 &
  {\color[HTML]{808080} 0.17} &
  1.39 &
  {\color[HTML]{808080} 0.35} &
  0.72 &
  {\color[HTML]{808080} 0.28} &
  1.00 &
  {\color[HTML]{808080} 0.01} &
  14.16 &
  {\color[HTML]{808080} 0.83} &
  \textbf{0.78} &
  {\color[HTML]{808080} 0.02} \\ \cline{3-16} 
 &
   &
  \checkmark &
  GRBF &
  0.69 &
  {\color[HTML]{808080} 0.18} &
  1.56 &
  {\color[HTML]{808080} 0.34} &
  0.84 &
  {\color[HTML]{808080} 0.22} &
  1.00 &
  {\color[HTML]{808080} 0.01} &
  \textbf{17.15} &
  {\color[HTML]{808080} 0.97} &
  0.77 &
  {\color[HTML]{808080} 0.02} \\ \cline{3-16} 
\multirow{-4}{*}{UBFC-rPPG} &
  \multirow{-4}{*}{PURE} &
  \textbf{\checkmark} &
  \textbf{Target Signal} &
  \textbf{0.30} &
  {\color[HTML]{808080} 0.12} &
  \textbf{1.00} &
  {\color[HTML]{808080} 0.29} &
  \textbf{0.41} &
  {\color[HTML]{808080} 0.18} &
  \textbf{1.00} &
  {\color[HTML]{808080} 0.01} &
  15.06 &
  {\color[HTML]{808080} 0.90} &
  \textbf{0.78} &
  {\color[HTML]{808080} 0.02} \\ \hline \hline
 &
   &
  - &
  NA &
  1.99 &
  {\color[HTML]{808080} 0.42} &
  4.82 &
  {\color[HTML]{808080} 1.03} &
  2.89 &
  {\color[HTML]{808080} 0.69} &
  0.87 &
  {\color[HTML]{808080} 0.05} &
  5.88 &
  {\color[HTML]{808080} 0.57} &
  0.54 &
  {\color[HTML]{808080} 0.01} \\ \cline{3-16} 
 &
   &
  \checkmark &
  None &
  \textbf{1.74} &
  {\color[HTML]{808080} 0.39} &
  4.39 &
  {\color[HTML]{808080} 1.06} &
  \textbf{2.42} &
  {\color[HTML]{808080} 0.57} &
  0.90 &
  {\color[HTML]{808080} 0.04} &
  \textbf{6.59} &
  {\color[HTML]{808080} 0.57} &
  \textbf{0.56} &
  {\color[HTML]{808080} 0.01} \\ \cline{3-16} 
 &
   &
  \checkmark &
  GRBF &
  1.84 &
  {\color[HTML]{808080} 0.36} &
  4.21 &
  {\color[HTML]{808080} 0.71} &
  2.43 &
  {\color[HTML]{808080} 0.47} &
  0.91 &
  {\color[HTML]{808080} 0.04} &
  6.21 &
  {\color[HTML]{808080} 0.63} &
  0.55 &
  {\color[HTML]{808080} 0.01} \\ \cline{3-16} 
\multirow{-4}{*}{UBFC-rPPG} &
  \multirow{-4}{*}{iBVP} &
  \textbf{\checkmark} &
  \textbf{Target Signal} &
  1.82 &
  {\color[HTML]{808080} 0.34} &
  \textbf{3.99} &
  {\color[HTML]{808080} 0.76} &
  2.48 &
  {\color[HTML]{808080} 0.48} &
  \textbf{0.92} &
  {\color[HTML]{808080} 0.04} &
  6.26 &
  {\color[HTML]{808080} 0.58} &
  0.55 &
  {\color[HTML]{808080} 0.01} \\ \hline \hline
\end{tabular*}
    \footnotesize
    \begin{flushleft}
    AVG: Average; SE: Standard Error; Double-horizontal lines to delineate the comparison group. \end{flushleft}
\end{table*}

As presented in Table \ref{tab:results ablation}, the implementation of an NMF-based attention block significantly enhances the estimation of the rPPG signal when compared with the baseline FactorizePhys \cite{joshi2024factorizephys} model that does not implement an attention mechanism. The imposition of GRBF \cite{yokota2015smooth} as constraints fails to consistently enhance performance. This phenomenon can be attributed to the generic temporal properties of GRBF, which do not provide effective constraints capable of augmenting the attention mechanism for the task of rPPG signal estimation. 

In contrast, the proposed \textit{Target Signal} constraint consistently enhances performance, achieving superior results across most evaluation metrics in both testing datasets. Despite variations in performance due to different testing datasets, the proposed approach \AM{} displays superior generalizability and robustness. The principal aim of this ablation study is to discern between the proposed \textit{Target Signal} and GRBF as a viable constraint to solve factorization of multidimensional embeddings. For our main evaluation on single-task and multi-task signal estimation, we conduct an exhaustive evaluation of \AM{} in contrast to FSAM \cite{joshi2024factorizephys}, which does not impose any constraints on NMF.

\subsection{Single-Task Evaluation: rPPG}\label{results single modality single task}
\subsubsection{Within-dataset Performance}\label{results within bvp}
Within-dataset performance refers to the ability of a model to accurately represent and fit data derived from the same distribution, thereby serving as a critical criterion. The state-of-the-art rPPG methods are trained and evaluated on three benchmark datasets with a 70\%-30\% split by participants. Although FactorizePhys \cite{joshi2024factorizephys} is employed to conduct a thorough comparison between FSAM \cite{joshi2024factorizephys} and our proposed attention module \AM{}, we also assess the \textit{BVP} branch of our proposed framework \Model{}, which is trained using \AM{}.
\begin{table*}[ht]
\caption{Within-dataset Performance Evaluation for rPPG Estimation}
\label{tab:results within all}
\centering
\fontsize{6}{4}\selectfont
\setlength{\tabcolsep}{1pt}
\renewcommand{\arraystretch}{2.25}
\begin{tabular*}{\textwidth}{@{\extracolsep\fill}llcccccccccccc}\hline
\multicolumn{1}{c}{} &
  \multicolumn{1}{c}{} &
  \multicolumn{2}{c}{\textbf{MAE ↓}} &
  \multicolumn{2}{c}{\textbf{RMSE   ↓}} &
  \multicolumn{2}{c}{\textbf{MAPE   ↓}} &
  \multicolumn{2}{c}{\textbf{Corr   ↑}} &
  \multicolumn{2}{c}{\textbf{SNR ↑}} &
  \multicolumn{2}{c}{\textbf{MACC   ↑}} \\ \cline{3-14} 
\multicolumn{1}{c}{\multirow{-2}{*}{\textbf{Model}}} &
  \multicolumn{1}{c}{\multirow{-2}{*}{\textbf{\begin{tabular}[c]{@{}c@{}}Attention\\ Module\end{tabular}}}} &
  \textbf{Avg} &
  {\color[HTML]{808080} \textbf{SE}} &
  \textbf{Avg} &
  {\color[HTML]{808080} \textbf{SE}} &
  \textbf{Avg} &
  {\color[HTML]{808080} \textbf{SE}} &
  \textbf{Avg} &
  {\color[HTML]{808080} \textbf{SE}} &
  \textbf{Avg} &
  {\color[HTML]{808080} \textbf{SE}} &
  \textbf{Avg} &
  {\color[HTML]{808080} \textbf{SE}} \\ \hline \hline
\multicolumn{14}{c}{\textbf{iBVP Dataset \cite{joshi2024ibvp}, Subject-wise Split: Train: 70,   Test: 30}} \\ \hline \hline
PhysNet &
  - &
  1.18 &
  {\color[HTML]{808080} 0.29} &
  \textbf{2.10} &
  {\color[HTML]{808080} 0.51} &
  1.64 &
  {\color[HTML]{808080} 0.42} &
  0.98 &
  {\color[HTML]{808080} 0.03} &
  10.63 &
  {\color[HTML]{808080} 1.05} &
  \textbf{0.68} &
  {\color[HTML]{808080} 0.02} \\ \hline
PhysFormer &
  MHSA * &
  1.96 &
  {\color[HTML]{808080} 0.63} &
  4.22 &
  {\color[HTML]{808080} 1.47} &
  2.49 &
  {\color[HTML]{808080} 0.72} &
  0.91 &
  {\color[HTML]{808080} 0.07} &
  \textbf{10.72} &
  {\color[HTML]{808080} 1.04} &
  0.66 &
  {\color[HTML]{808080} 0.03} \\ \hline
EfficientPhys &
  SASN \dag &
  2.74 &
  {\color[HTML]{808080} 0.96} &
  6.28 &
  {\color[HTML]{808080} 2.14} &
  3.56 &
  {\color[HTML]{808080} 1.13} &
  0.81 &
  {\color[HTML]{808080} 0.10} &
  7.01 &
  {\color[HTML]{808080} 1.03} &
  0.58 &
  {\color[HTML]{808080} 0.03} \\ \hline
FactorizePhys &
  FSAM \ddag &
  1.13 &
  {\color[HTML]{808080} 0.36} &
  2.42 &
  {\color[HTML]{808080} 0.77} &
  1.52 &
  {\color[HTML]{808080} 0.50} &
  0.97 &
  {\color[HTML]{808080} 0.04} &
  9.75 &
  {\color[HTML]{808080} 1.05} &
  0.65 &
  {\color[HTML]{808080} 0.02} \\ \hline
FactorizePhys &
  \textbf{TSFM} &
  0.90 &
  {\color[HTML]{808080} 0.33} &
  2.20 &
  {\color[HTML]{808080} 0.67} &
  1.18 &
  {\color[HTML]{808080} 0.43} &
  0.98 &
  {\color[HTML]{808080} 0.04} &
  10.10 &
  {\color[HTML]{808080} 1.09} &
  0.64 &
  {\color[HTML]{808080} 0.02} \\ \hline
\textbf{MMRPhys} &
  \textbf{TSFM} &
  \textbf{0.67} &
  {\color[HTML]{808080} 0.27} &
  \textbf{1.73} &
  {\color[HTML]{808080} 0.61} &
  \textbf{0.89} &
  {\color[HTML]{808080} 0.36} &
  \textbf{0.99} &
  {\color[HTML]{808080} 0.03} &
  9.54 &
  {\color[HTML]{808080} 1.02} &
  0.64 &
  {\color[HTML]{808080} 0.02} \\ \hline \hline
\multicolumn{14}{c}{\textbf{PURE Dataset \cite{stricker2014non}, Subject-wise Split: Train: 70,   Test: 30}} \\ \hline \hline
PhysNet &
  - &
  0.59 &
  {\color[HTML]{808080} 0.27} &
  1.28 &
  {\color[HTML]{808080} 0.46} &
  0.92 &
  {\color[HTML]{808080} 0.44} &
  0.9970 &
  {\color[HTML]{808080} 0.02} &
  19.66 &
  {\color[HTML]{808080} 1.18} &
  \textbf{0.90} &
  {\color[HTML]{808080} 0.01} \\ \hline
PhysFormer &
  MHSA * &
  0.68 &
  {\color[HTML]{808080} 0.26} &
  1.31 &
  {\color[HTML]{808080} 0.46} &
  1.08 &
  {\color[HTML]{808080} 0.43} &
  0.9968 &
  {\color[HTML]{808080} 0.02} &
  19.05 &
  {\color[HTML]{808080} 1.07} &
  0.87 &
  {\color[HTML]{808080} 0.01} \\ \hline
EfficientPhys &
  SASN \dag &
  0.49 &
  {\color[HTML]{808080} 0.26} &
  1.21 &
  {\color[HTML]{808080} 0.46} &
  0.73 &
  {\color[HTML]{808080} 0.42} &
  0.9975 &
  {\color[HTML]{808080} 0.02} &
  15.25 &
  {\color[HTML]{808080} 1.20} &
  0.80 &
  {\color[HTML]{808080} 0.02} \\ \hline
FactorizePhys &
  FSAM \ddag &
  0.49 &
  {\color[HTML]{808080} 0.26} &
  1.21 &
  {\color[HTML]{808080} 0.46} &
  0.73 &
  {\color[HTML]{808080} 0.42} &
  0.9974 &
  {\color[HTML]{808080} 0.02} &
  19.63 &
  {\color[HTML]{808080} 1.40} &
  0.86 &
  {\color[HTML]{808080} 0.01} \\ \hline
FactorizePhys &
  \textbf{TSFM} &
  0.48 &
  {\color[HTML]{808080} 0.18} &
  0.91 &
  {\color[HTML]{808080} 0.27} &
  0.72 &
  {\color[HTML]{808080} 0.29} &
  0.9986 &
  {\color[HTML]{808080} 0.01} &
  \textbf{23.56} &
  {\color[HTML]{808080} 1.59} &
  0.87 &
  {\color[HTML]{808080} 0.01} \\ \hline
\textbf{MMRPhys} &
  \textbf{TSFM} &
  \textbf{0.28} &
  {\color[HTML]{808080} 0.15} &
  \textbf{0.69} &
  {\color[HTML]{808080} 0.25} &
  \textbf{0.44} &
  {\color[HTML]{808080} 0.24} &
  \textbf{0.9991} &
  {\color[HTML]{808080} 0.01} &
  23.02 &
  {\color[HTML]{808080} 1.40} &
  0.88 &
  {\color[HTML]{808080} 0.01} \\ \hline \hline
\multicolumn{14}{c}{\textbf{UBFC-rPPG Dataset \cite{bobbia2019unsupervised}, Subject-wise Split:   Train: 70, Test: 30}} \\ \hline \hline
PhysNet &
  - &
  1.62 &
  {\color[HTML]{808080} 0.73} &
  3.08 &
  {\color[HTML]{808080} 1.16} &
  1.46 &
  {\color[HTML]{808080} 0.68} &
  0.98 &
  {\color[HTML]{808080} 0.06} &
  5.21 &
  {\color[HTML]{808080} 1.97} &
  0.90 &
  {\color[HTML]{808080} 0.01} \\ \hline
PhysFormer &
  MHSA * &
  1.76 &
  {\color[HTML]{808080} 0.79} &
  3.36 &
  {\color[HTML]{808080} 1.30} &
  1.60 &
  {\color[HTML]{808080} 0.74} &
  0.96 &
  {\color[HTML]{808080} 0.08} &
  6.10 &
  {\color[HTML]{808080} 1.86} &
  0.90 &
  {\color[HTML]{808080} 0.01} \\ \hline
EfficientPhys &
  SASN \dag &
  2.30 &
  {\color[HTML]{808080} 1.40} &
  5.54 &
  {\color[HTML]{808080} 2.53} &
  2.28 &
  {\color[HTML]{808080} 1.44} &
  0.90 &
  {\color[HTML]{808080} 0.13} &
  6.75 &
  {\color[HTML]{808080} 1.76} &
  0.87 &
  {\color[HTML]{808080} 0.01} \\ \hline
FactorizePhys &
  FSAM \ddag &
  2.84 &
  {\color[HTML]{808080} 1.42} &
  5.87 &
  {\color[HTML]{808080} 2.52} &
  2.73 &
  {\color[HTML]{808080} 1.46} &
  0.88 &
  {\color[HTML]{808080} 0.14} &
  6.33 &
  {\color[HTML]{808080} 2.00} &
  \textbf{0.91} &
  {\color[HTML]{808080} 0.01} \\ \hline
FactorizePhys &
  \textbf{TSFM} &
  2.30 &
  {\color[HTML]{808080} 1.26} &
  5.10 &
  {\color[HTML]{808080} 2.16} &
  2.24 &
  {\color[HTML]{808080} 1.27} &
  0.91 &
  {\color[HTML]{808080} 0.12} &
  8.16 &
  {\color[HTML]{808080} 2.28} &
  \textbf{0.91} &
  {\color[HTML]{808080} 0.01} \\ \hline
\textbf{MMRPhys} &
  \textbf{TSFM} &
  \textbf{0.00} &
  {\color[HTML]{808080} 0.00} &
  \textbf{0.00} &
  {\color[HTML]{808080} 0.00} &
  \textbf{0.00} &
  {\color[HTML]{808080} 0.00} &
  \textbf{1.00} &
  {\color[HTML]{808080} 0.00} &
  \textbf{9.07} &
  {\color[HTML]{808080} 2.03} &
  \textbf{0.91} &
  {\color[HTML]{808080} 0.01} \\ \hline \hline
  \end{tabular*}
    \footnotesize
    \begin{flushleft}
    \ddag FSAM: Factorized Self-Attention Module \cite{joshi2024factorizephys}; \dag SASN: Self-Attention Shifted Network \cite{liu2023efficientphys}; * MHSA: Temporal Difference Multi-Head Self-Attention \cite{yu2022physformer};  TSFM: Proposed Attention Module; AVG: Average; SE: Standard Error; Double-horizontal lines to delineate the comparison group.
    \end{flushleft}
\end{table*}

The findings reported in Table \ref{tab:results within all} indicate that \AM{} typically enhances the performance of the FactorizePhys model \cite{joshi2024factorizephys} across a majority of metrics and on all three datasets. Furthermore, it is observed that the optimal performance on all datasets is consistently attained through the integration of our proposed methods, specifically \Model{} trained with \AM{}.

\subsubsection{Cross-dataset Generalization}\label{results bvp cross dataset}
The evaluation of cross-dataset generalization is a crucial task that provides a thorough analysis and reliable estimates of model performance on data that are unseen or are outside the original distribution. Table \ref{tab:results cross rPPG averaged} presents the consolidated cross-dataset evaluation of the SOTA rPPG methods and our proposed methods, trained on the iBVP \cite{joshi2024ibvp}, PURE \cite{stricker2014non}, SCAMPS \cite{mcduff2022scamps}, and UBFC-rPPG \cite{bobbia2019unsupervised} datasets, and evaluated on the remaining three datasets. The outcomes reported for each model in Table \ref{tab:results cross rPPG averaged}, trained on the specified datasets, are derived by averaging their performance across multiple test datasets, as outlined in the \textit{Datasets} column. As an exception, the SCAMPS dataset \cite{mcduff2022scamps}, is used exclusively for training purposes due to its composition of synthesized videos.
\begin{table*}[ht]
\caption{Average Cross-dataset Performance Evaluation for rPPG Estimation}
\label{tab:results cross rPPG averaged}
\centering
\fontsize{6}{4}\selectfont
\setlength{\tabcolsep}{1pt}
\renewcommand{\arraystretch}{1.9}
\begin{tabular*}{\textwidth}{@{\extracolsep\fill}lllcccccccccccc}\hline
\multicolumn{1}{c}{} &
  \multicolumn{1}{c}{} &
  \multicolumn{1}{c}{} &
  \multicolumn{2}{c}{\textbf{MAE ↓}} &
  \multicolumn{2}{c}{\textbf{RMSE ↓}} &
  \multicolumn{2}{c}{\textbf{MAPE↓}} &
  \multicolumn{2}{c}{\textbf{Corr ↑}} &
  \multicolumn{2}{c}{\textbf{SNR ↑}} &
  \multicolumn{2}{c}{\textbf{MACC ↑}} \\ \cline{4-15} 
\multicolumn{1}{c}{\multirow{-2}{*}{\textbf{Datasets}}} &
  \multicolumn{1}{c}{\multirow{-2}{*}{\textbf{Model}}} &
  \multicolumn{1}{c}{\multirow{-2}{*}{\textbf{\begin{tabular}[c]{@{}c@{}}Attention\\      Module\end{tabular}}}} &
  \textbf{Avg} &
  {\color[HTML]{808080} \textbf{SE}} &
  \textbf{Avg} &
  {\color[HTML]{808080} \textbf{SE}} &
  \textbf{Avg} &
  {\color[HTML]{808080} \textbf{SE}} &
  \textbf{Avg} &
  {\color[HTML]{808080} \textbf{SE}} &
  \textbf{Avg} &
  {\color[HTML]{808080} \textbf{SE}} &
  \textbf{Avg} &
  {\color[HTML]{808080} \textbf{SE}} \\ \hline \hline   
 &
  PhysNet &
  - &
  \cellcolor[HTML]{E8E7E4}5.44 &
  {\color[HTML]{808080} 2.03} &
  \cellcolor[HTML]{F1E5DD}14.92 &
  {\color[HTML]{808080} 4.09} &
  \cellcolor[HTML]{ECE6E0}5.89 &
  {\color[HTML]{808080} 2.08} &
  \cellcolor[HTML]{EFE4DE}0.70 &
  {\color[HTML]{808080} 0.11} &
  \cellcolor[HTML]{E5E7E6}8.52 &
  {\color[HTML]{808080} 1.51} &
  \cellcolor[HTML]{E2E7E9}0.76 &
  {\color[HTML]{808080} 0.03} \\ \cline{2-15} 
 &
  PhysFormer &
  MHSA* &
  \cellcolor[HTML]{FBE2D5}8.23 &
  {\color[HTML]{808080} 2.47} &
  \cellcolor[HTML]{FBE2D5}18.07 &
  {\color[HTML]{808080} 4.48} &
  \cellcolor[HTML]{FBE2D5}7.83 &
  {\color[HTML]{808080} 2.16} &
  \cellcolor[HTML]{FBE2D5}0.60 &
  {\color[HTML]{808080} 0.13} &
  \cellcolor[HTML]{FBE2D5}6.28 &
  {\color[HTML]{808080} 2.09} &
  \cellcolor[HTML]{FBE2D5}0.71 &
  {\color[HTML]{808080} 0.03} \\ \cline{2-15} 
 &
  EfficientPhys &
  SASN \dag &
  \cellcolor[HTML]{CAEDFB}0.85 &
  {\color[HTML]{808080} 0.31} &
  \cellcolor[HTML]{C9EEFA}2.13 &
  {\color[HTML]{808080} 0.61} &
  \cellcolor[HTML]{CAEDFB}1.15 &
  {\color[HTML]{808080} 0.43} &
  \cellcolor[HTML]{C9EDF4}0.993 &
  {\color[HTML]{808080} 0.02} &
  \cellcolor[HTML]{D3EBF4}10.34 &
  {\color[HTML]{808080} 1.04} &
  \cellcolor[HTML]{D2EBF4}0.79 &
  {\color[HTML]{808080} 0.02} \\ \cline{2-15} 
 &
  FactorizePhys &
  FSAM \ddag &
  \cellcolor[HTML]{C9EEF7}0.82 &
  {\color[HTML]{808080} 0.30} &
  \cellcolor[HTML]{C9EEF5}2.05 &
  {\color[HTML]{808080} 0.56} &
  \cellcolor[HTML]{C9EEF5}1.05 &
  {\color[HTML]{808080} 0.39} &
  \cellcolor[HTML]{C8EDEF}0.994 &
  {\color[HTML]{808080} 0.02} &
  \cellcolor[HTML]{C5EEDC}12.02 &
  {\color[HTML]{808080} 1.11} &
  \cellcolor[HTML]{C5EEDE}0.815 &
  {\color[HTML]{808080} 0.02} \\ \cline{2-15} 
 &
  FactorizePhys &
  \textbf{TSFM} &
  \cellcolor[HTML]{C9EEFA}0.84 &
  {\color[HTML]{808080} 0.31} &
  \cellcolor[HTML]{CAEDFB}2.14 &
  {\color[HTML]{808080} 0.61} &
  \cellcolor[HTML]{C9EEF7}1.07 &
  {\color[HTML]{808080} 0.42} &
  \cellcolor[HTML]{CBECFA}0.991 &
  {\color[HTML]{808080} 0.02} &
  \cellcolor[HTML]{C1F0C8}\textbf{12.55} &
  {\color[HTML]{808080} 1.12} &
  \cellcolor[HTML]{C5EEDA}0.818 &
  {\color[HTML]{808080} 0.02} \\ \cline{2-15} 
\multirow{-9}{*}{\begin{tabular}[c]{@{}l@{}}Train:\\ ~~*~iBVP\\ \\ Test:\\ ~~*~PURE,\\ ~~*~UBFC-rPPG\end{tabular}} &
  \textbf{\Model{}} &
  \textbf{TSFM} &
  \cellcolor[HTML]{C1F0C8}\textbf{0.43} &
  {\color[HTML]{808080} 0.19} &
  \cellcolor[HTML]{C1F0C8}\textbf{1.32} &
  {\color[HTML]{808080} 0.42} &
  \cellcolor[HTML]{C1F0C8}\textbf{0.53} &
  {\color[HTML]{808080} 0.23} &
  \cellcolor[HTML]{C1F0C8}\textbf{0.998} &
  {\color[HTML]{808080} 0.01} &
  \cellcolor[HTML]{C3EFD0}12.36 &
  {\color[HTML]{808080} 1.08} &
  \cellcolor[HTML]{C1F0C8}\textbf{0.826} &
  {\color[HTML]{808080} 0.02} \\ \hline \hline
 &
  PhysNet &
  - &
  \cellcolor[HTML]{CBEDFA}1.43 &
  {\color[HTML]{808080} 0.37} &
  \cellcolor[HTML]{C9EEF8}3.21 &
  {\color[HTML]{808080} 0.72} &
  \cellcolor[HTML]{CAEDFB}1.80 &
  {\color[HTML]{808080} 0.46} &
  \cellcolor[HTML]{C5EEDC}0.95 &
  {\color[HTML]{808080} 0.04} &
  \cellcolor[HTML]{D0EBF6}7.21 &
  {\color[HTML]{808080} 0.92} &
  \cellcolor[HTML]{D1EBF6}0.70 &
  {\color[HTML]{808080} 0.01} \\ \cline{2-15} 
 &
  PhysFormer &
  MHSA* &
  \cellcolor[HTML]{D8EAF0}1.76 &
  {\color[HTML]{808080} 0.51} &
  \cellcolor[HTML]{D6EBF2}4.75 &
  {\color[HTML]{808080} 1.10} &
  \cellcolor[HTML]{D9EAEF}2.30 &
  {\color[HTML]{808080} 0.65} &
  \cellcolor[HTML]{DAE9EE}0.89 &
  {\color[HTML]{808080} 0.05} &
  \cellcolor[HTML]{D7EAF1}6.82 &
  {\color[HTML]{808080} 0.92} &
  \cellcolor[HTML]{D9E9EF}0.69 &
  {\color[HTML]{808080} 0.01} \\ \cline{2-15} 
 &
  EfficientPhys &
  SASN \dag &
  \cellcolor[HTML]{FBE2D5}2.61 &
  {\color[HTML]{808080} 0.94} &
  \cellcolor[HTML]{FBE2D5}8.99 &
  {\color[HTML]{808080} 2.34} &
  \cellcolor[HTML]{FBE2D5}3.42 &
  {\color[HTML]{808080} 1.26} &
  \cellcolor[HTML]{FBE2D5}0.77 &
  {\color[HTML]{808080} 0.06} &
  \cellcolor[HTML]{FBE2D5}4.90 &
  {\color[HTML]{808080} 0.82} &
  \cellcolor[HTML]{FBE2D5}0.62 &
  {\color[HTML]{808080} 0.02} \\ \cline{2-15} 
 &
  FactorizePhys &
  FSAM \ddag &
  \cellcolor[HTML]{C8EEF2}1.35 &
  {\color[HTML]{808080} 0.34} &
  \cellcolor[HTML]{C6EFE8}3.00 &
  {\color[HTML]{808080} 0.67} &
  \cellcolor[HTML]{C9EEF9}1.77 &
  {\color[HTML]{808080} 0.47} &
  \cellcolor[HTML]{C1F0C8}\textbf{0.960} &
  {\color[HTML]{808080} 0.04} &
  \cellcolor[HTML]{C2EFCB}7.83 &
  {\color[HTML]{808080} 0.94} &
  \cellcolor[HTML]{C2EFCB}0.725 &
  {\color[HTML]{808080} 0.01} \\ \cline{2-15} 
 &
  FactorizePhys &
  \textbf{TSFM} &
  \cellcolor[HTML]{C8EEF5}1.37 &
  {\color[HTML]{808080} 0.36} &
  \cellcolor[HTML]{CAEDFB}3.28 &
  {\color[HTML]{808080} 0.79} &
  \cellcolor[HTML]{C8EEF3}1.73 &
  {\color[HTML]{808080} 0.45} &
  \cellcolor[HTML]{CDECF9}0.937 &
  {\color[HTML]{808080} 0.03} &
  \cellcolor[HTML]{C3EFD1}7.79 &
  {\color[HTML]{808080} 0.93} &
  \cellcolor[HTML]{C3EFD2}0.723 &
  {\color[HTML]{808080} 0.01} \\ \cline{2-15} 
\multirow{-9}{*}{\begin{tabular}[c]{@{}l@{}}Train:\\ ~~*~PURE\\ \\ Test:\\ ~~*~iBVP,\\ ~~*~UBFC-rPPG\end{tabular}} &
  \textbf{\Model{}} &
  \textbf{TSFM} &
  \cellcolor[HTML]{C1F0C8}\textbf{1.12} &
  {\color[HTML]{808080} 0.27} &
  \cellcolor[HTML]{C1F0C8}\textbf{2.53} &
  {\color[HTML]{808080} 0.55} &
  \cellcolor[HTML]{C1F0C8}\textbf{1.42} &
  {\color[HTML]{808080} 0.31} &
  \cellcolor[HTML]{C2EFCE}0.958 &
  {\color[HTML]{808080} 0.02} &
  \cellcolor[HTML]{C1F0C8}\textbf{7.85} &
  {\color[HTML]{808080} 0.88} &
  \cellcolor[HTML]{C1F0C8}\textbf{0.726} &
  {\color[HTML]{808080} 0.01} \\ \hline \hline
 &
  PhysNet &
  - &
  \cellcolor[HTML]{FBE2D5}23.28 &
  {\color[HTML]{808080} 2.56} &
  \cellcolor[HTML]{FBE2D5}30.80 &
  {\color[HTML]{808080} 4.42} &
  \cellcolor[HTML]{FBE2D5}35.30 &
  {\color[HTML]{808080} 4.14} &
  \cellcolor[HTML]{FBE2D5}0.24 &
  {\color[HTML]{808080} 0.13} &
  \cellcolor[HTML]{FBE2D5}-2.80 &
  {\color[HTML]{808080} 0.63} &
  \cellcolor[HTML]{FBE2D5}0.32 &
  {\color[HTML]{808080} 0.02} \\ \cline{2-15} 
 &
  PhysFormer &
  MHSA* &
  \cellcolor[HTML]{F7E3D8}22.26 &
  {\color[HTML]{808080} 2.33} &
  \cellcolor[HTML]{F8E3D8}29.92 &
  {\color[HTML]{808080} 4.40} &
  \cellcolor[HTML]{F7E3D8}33.47 &
  {\color[HTML]{808080} 4.11} &
  \cellcolor[HTML]{ECE5E0}0.38 &
  {\color[HTML]{808080} 0.12} &
  \cellcolor[HTML]{F6E3D9}-2.00 &
  {\color[HTML]{808080} 0.95} &
  \cellcolor[HTML]{EDE5E0}0.39 &
  {\color[HTML]{808080} 0.02} \\ \cline{2-15} 
 &
  EfficientPhys &
  SASN \dag &
  \cellcolor[HTML]{D6EBF1}11.53 &
  {\color[HTML]{808080} 2.16} &
  \cellcolor[HTML]{E0E8EA}22.61 &
  {\color[HTML]{808080} 4.05} &
  \cellcolor[HTML]{D6EBF2}17.54 &
  {\color[HTML]{808080} 3.51} &
  \cellcolor[HTML]{DDE8EC}0.53 &
  {\color[HTML]{808080} 0.08} &
  \cellcolor[HTML]{DBE9ED}2.14 &
  {\color[HTML]{808080} 0.73} &
  \cellcolor[HTML]{D9E9EF}0.49 &
  {\color[HTML]{808080} 0.01} \\ \cline{2-15} 
 &
  FactorizePhys &
  FSAM \ddag &
  \cellcolor[HTML]{C3F0D5}3.10 &
  {\color[HTML]{808080} 0.96} &
  \cellcolor[HTML]{C3F0D5}8.19 &
  {\color[HTML]{808080} 1.89} &
  \cellcolor[HTML]{C4EFD9}5.44 &
  {\color[HTML]{808080} 1.78} &
  \cellcolor[HTML]{C3EFD2}0.87 &
  {\color[HTML]{808080} 0.06} &
  \cellcolor[HTML]{C5EEDB}7.39 &
  {\color[HTML]{808080} 0.81} &
  \cellcolor[HTML]{C5EEDD}0.64 &
  {\color[HTML]{808080} 0.01} \\ \cline{2-15} 
 &
  FactorizePhys &
  \textbf{TSFM} &
  \cellcolor[HTML]{C1F0C8}1.542 &
  {\color[HTML]{808080} 0.66} &
  \cellcolor[HTML]{C1F0C8}\textbf{5.71} &
  {\color[HTML]{808080} 1.76} &
  \cellcolor[HTML]{C1F0C8}2.20 &
  {\color[HTML]{808080} 1.03} &
  \cellcolor[HTML]{C1F0C8}\textbf{0.906} &
  {\color[HTML]{808080} 0.04} &
  \cellcolor[HTML]{C1F0C8}\textbf{8.91} &
  {\color[HTML]{808080} 0.88} &
  \cellcolor[HTML]{C1F0C8}\textbf{0.689} &
  {\color[HTML]{808080} 0.01} \\ \cline{2-15} 
\multirow{-9}{*}{\begin{tabular}[c]{@{}l@{}}Train:\\ ~~*~SCAMPS\\ \\ Test:\\ ~~*~iBVP,\\ ~~*~PURE,\\ ~~*~UBFC-rPPG\end{tabular}} &
  \textbf{\Model{}} &
  \textbf{TSFM} &
  \cellcolor[HTML]{C1F0C8}\textbf{1.539} &
  {\color[HTML]{808080} 0.66} &
  \cellcolor[HTML]{C1F0C8}5.75 &
  {\color[HTML]{808080} 1.76} &
  \cellcolor[HTML]{C1F0C8}\textbf{2.18} &
  {\color[HTML]{808080} 1.02} &
  \cellcolor[HTML]{C2EFCA}0.900 &
  {\color[HTML]{808080} 0.04} &
  \cellcolor[HTML]{C2EFCE}8.50 &
  {\color[HTML]{808080} 0.83} &
  \cellcolor[HTML]{C2EFCB}0.682 &
  {\color[HTML]{808080} 0.01} \\ \hline \hline
 &
  PhysNet &
  - &
  \cellcolor[HTML]{EDE5E0}6.78 &
  {\color[HTML]{808080} 1.54} &
  \cellcolor[HTML]{EBE6E2}14.40 &
  {\color[HTML]{808080} 2.68} &
  \cellcolor[HTML]{F3E4DB}12.88 &
  {\color[HTML]{808080} 3.06} &
  \cellcolor[HTML]{E6E6E5}0.68 &
  {\color[HTML]{808080} 0.09} &
  \cellcolor[HTML]{D4EAF3}8.28 &
  {\color[HTML]{808080} 0.79} &
  \cellcolor[HTML]{CEECF8}0.64 &
  {\color[HTML]{808080} 0.02} \\ \cline{2-15} 
 &
  PhysFormer &
  MHSA* &
  \cellcolor[HTML]{FBE2D5}8.38 &
  {\color[HTML]{808080} 1.81} &
  \cellcolor[HTML]{FBE2D5}17.95 &
  {\color[HTML]{808080} 3.18} &
  \cellcolor[HTML]{FBE2D5}14.56 &
  {\color[HTML]{808080} 3.39} &
  \cellcolor[HTML]{FBE2D5}0.55 &
  {\color[HTML]{808080} 0.09} &
  \cellcolor[HTML]{FBE2D5}5.22 &
  {\color[HTML]{808080} 0.79} &
  \cellcolor[HTML]{FBE2D5}0.55 &
  {\color[HTML]{808080} 0.02} \\ \cline{2-15} 
 &
  EfficientPhys &
  SASN \dag &
  \cellcolor[HTML]{D4EBF3}3.73 &
  {\color[HTML]{808080} 1.21} &
  \cellcolor[HTML]{DBEAEE}10.80 &
  {\color[HTML]{808080} 2.73} &
  \cellcolor[HTML]{D3EBF4}5.83 &
  {\color[HTML]{808080} 2.03} &
  \cellcolor[HTML]{D8E9F0}0.77 &
  {\color[HTML]{808080} 0.08} &
  \cellcolor[HTML]{ECE5E0}6.40 &
  {\color[HTML]{808080} 0.78} &
  \cellcolor[HTML]{EDE5E0}0.58 &
  {\color[HTML]{808080} 0.02} \\ \cline{2-15} 
 &
  FactorizePhys &
  FSAM \ddag &
  \cellcolor[HTML]{C1F0C9}1.110 &
  {\color[HTML]{808080} 0.28} &
  \cellcolor[HTML]{C1F0CC}2.89 &
  {\color[HTML]{808080} 0.71} &
  \cellcolor[HTML]{C1F0CA}1.57 &
  {\color[HTML]{808080} 0.43} &
  \cellcolor[HTML]{C2EFCD}0.949 &
  {\color[HTML]{808080} 0.03} &
  \cellcolor[HTML]{C3EFD2}10.38 &
  {\color[HTML]{808080} 0.70} &
  \cellcolor[HTML]{C1F0C8}\textbf{0.670} &
  {\color[HTML]{808080} 0.02} \\ \cline{2-15} 
 &
  FactorizePhys &
  \textbf{TSFM} &
  \cellcolor[HTML]{C1F0C8}\textbf{1.061} &
  {\color[HTML]{808080} 0.23} &
  \cellcolor[HTML]{C1F0C8}\textbf{2.50} &
  {\color[HTML]{808080} 0.53} &
  \cellcolor[HTML]{C1F0C8}\textbf{1.45} &
  {\color[HTML]{808080} 0.33} &
  \cellcolor[HTML]{C1F0C8}\textbf{0.957} &
  {\color[HTML]{808080} 0.02} &
  \cellcolor[HTML]{C1F0C8}\textbf{10.66} &
  {\color[HTML]{808080} 0.74} &
  \cellcolor[HTML]{C4EFD4}0.665 &
  {\color[HTML]{808080} 0.02} \\ \cline{2-15} 
\multirow{-9}{*}{\begin{tabular}[c]{@{}l@{}}Train:\\ ~~*~UBFC-rPPG\\ \\ Test:\\ ~~*~iBVP,\\ ~~*~PURE\end{tabular}} &
  \textbf{\Model{}} &
  \textbf{TSFM} &
  \cellcolor[HTML]{C1F0CA}1.129 &
  {\color[HTML]{808080} 0.24} &
  \cellcolor[HTML]{C1F0C9}2.60 &
  {\color[HTML]{808080} 0.53} &
  \cellcolor[HTML]{C1F0CA}1.55 &
  {\color[HTML]{808080} 0.34} &
  \cellcolor[HTML]{C2EFCB}0.953 &
  {\color[HTML]{808080} 0.03} &
  \cellcolor[HTML]{C6EEE3}9.83 &
  {\color[HTML]{808080} 0.73} &
  \cellcolor[HTML]{C7EEE9}0.656 &
  {\color[HTML]{808080} 0.02} \\ \hline \hline \end{tabular*}
    \footnotesize
    \begin{flushleft}
    \ddag FSAM: Factorized Self-Attention Module \cite{joshi2024factorizephys}; \dag SASN: Self-Attention Shifted Network \cite{liu2023efficientphys}; \\
    * MHSA: Temporal Difference Multi-Head Self-Attention \cite{yu2022physformer};  TSFM: Proposed Attention Module; SE: Standard Error; Cell color scale: Within each comparison group (delineated by double horizontal lines), values are highlighted using 3-color scale (green = best, orange = worst, blue = midpoint; Microsoft Excel conditional formatting) \\
    \end{flushleft}
\end{table*}
For a detailed examination of the individual cross-dataset results for each method, please refer to the Supplementary Table \ref{tab:results cross all}. The comprehensive findings underscore the efficacy of the proposed \AM{}, highlighting its consistent performance across diverse training datasets for both FactorizePhys \cite{joshi2024factorizephys} and \Model{}, in comparison to established architectures. Notably, sustained performance enhancements are observed for FactorizePhys \cite{joshi2024factorizephys} when trained with \AM{}, compared to its training with FSAM \cite{joshi2024factorizephys}. Importantly, \Model{} trained with \AM{} achieves state-of-the-art results, especially when utilizing the iBVP \cite{joshi2024ibvp}, PURE \cite{stricker2014non}, and SCAMPS \cite{mcduff2022scamps} datasets, and exhibits comparable performance when trained with the UBFC-rPPG \cite{bobbia2019unsupervised} dataset. 

Furthermore, Table \ref{tab:results cross rPPG averaged} illustrates the dependence of model performance on the training datasets employed; for instance, all models trained with the PURE dataset \cite{stricker2014non} consistently exhibit superior performance across multiple test datasets, while training with the SCAMPS dataset \cite{mcduff2022scamps} significantly impairs the performance of SOTA models. As SCAMPS is a synthetic dataset, it represents a significant distribution shift from real-world datasets. The performance of \Model{} and FactorizePhys \cite{joshi2024factorizephys} trained with \AM{} remains consistent across a variety of training datasets, demonstrating robust generalization between different distribution shifts compared to the existing SOTA methods.

\subsection{Multi-task Evaluation: rRSP, rPPG}\label{results multi modality multi task}
Analogous to the evaluation of single-task estimation, we conduct both within-dataset and cross-dataset evaluations for the multi-task estimation of rRSP and rPPG, facilitating a comparative analysis with the methodologies proposed by BigSmall \cite{narayanswamy2024bigsmall}. Of the datasets employed in this investigation, only two are amenable to training for multi-task estimation of rRSP and rPPG. Among these, the SCAMPS dataset \cite{mcduff2022scamps}, which consists of videos featuring synthesized avatars, is utilized exclusively for training purposes. 

Although the SOTA method \cite{narayanswamy2024bigsmall} did not explore the thermal modality, drawing inspiration from prior research on the efficacy of thermal frames in the estimation of rRSP \cite{cho2017robust}, we undertake a comprehensive evaluation utilizing thermal video frames. This involves training both BigSmall \cite{narayanswamy2024bigsmall} and \Model{} on the thermal frames of the BP4D+ dataset \cite{zhang2016multimodal}. In the context of cross-dataset evaluation incorporating thermal frames, qualitative evaluations are carried out using the iBVP dataset \cite{joshi2024ibvp}, given the absence of ground-truth rRSP signals.

\begin{sidewaystable*}[ht]
\caption{Multitask Performance Evaluation for Simultaneous rPPG and rRSP Estimation}
\label{tab:results multi-task}
\centering
\fontsize{6}{4}\selectfont
\setlength{\tabcolsep}{0.5pt}
\renewcommand{\arraystretch}{2.5}
\begin{tabular*}{\textwidth}{@{\extracolsep\fill}lllcccccccccccccccccccccccc}
\hline
\multicolumn{1}{c}{} &
  \multicolumn{1}{c}{} &
  \multicolumn{1}{c|}{} &
  \multicolumn{12}{c|}{\textbf{rRSP, RR}} &
  \multicolumn{12}{c}{\textbf{rPPG, HR}} \\ \cline{4-27} 
\multicolumn{1}{c}{} &
  \multicolumn{1}{c}{} &
  \multicolumn{1}{c|}{} &
  \multicolumn{2}{c}{\textbf{MAE ↓}} &
  \multicolumn{2}{c}{\textbf{RMSE ↓}} &
  \multicolumn{2}{c}{\textbf{MAPE ↓}} &
  \multicolumn{2}{c}{\textbf{Corr ↑}} &
  \multicolumn{2}{c}{\textbf{SNR ↑}} &
  \multicolumn{2}{c|}{\textbf{MACC ↑}} &
  \multicolumn{2}{c}{\textbf{MAE ↓}} &
  \multicolumn{2}{c}{\textbf{RMSE ↓}} &
  \multicolumn{2}{c}{\textbf{MAPE ↓}} &
  \multicolumn{2}{c}{\textbf{Corr ↑}} &
  \multicolumn{2}{c}{\textbf{SNR ↑}} &
  \multicolumn{2}{c}{\textbf{MACC ↑}} \\ \cline{4-27} 
\multicolumn{1}{c}{\multirow{-3}{*}{\textbf{Model}}} &
  \multicolumn{1}{c}{\multirow{-3}{*}{\textbf{\begin{tabular}[c]{@{}c@{}}Image\\ Modality\end{tabular}}}} &
  \multicolumn{1}{c|}{\multirow{-3}{*}{\textbf{\begin{tabular}[c]{@{}c@{}}Spatial \\ Resolution\end{tabular}}}} &
  \textbf{Avg} &
  {\color[HTML]{808080} \textbf{SE}} &
  \textbf{Avg} &
  {\color[HTML]{808080} \textbf{SE}} &
  \textbf{Avg} &
  {\color[HTML]{808080} \textbf{SE}} &
  \textbf{Avg} &
  {\color[HTML]{808080} \textbf{SE}} &
  \textbf{Avg} &
  {\color[HTML]{808080} \textbf{SE}} &
  \textbf{Avg} &
  \multicolumn{1}{c|}{{\color[HTML]{808080} \textbf{SE}}} &
  \textbf{Avg} &
  {\color[HTML]{808080} \textbf{SE}} &
  \textbf{Avg} &
  {\color[HTML]{808080} \textbf{SE}} &
  \textbf{Avg} &
  {\color[HTML]{808080} \textbf{SE}} &
  \textbf{Avg} &
  {\color[HTML]{808080} \textbf{SE}} &
  \textbf{Avg} &
  {\color[HTML]{808080} \textbf{SE}} &
  \textbf{Avg} &
  {\color[HTML]{808080} \textbf{SE}} \\ \hline 
\multicolumn{27}{l}{\begin{tabular}[c]{@{}l@{}}Within Dataset Performance Evaluation on BP4D+ Dataset.\\ Results reported as an average of three fold cross-validation based on subject-wise split.\end{tabular}} \\ \hline \hline
 &
  \cellcolor[HTML]{F2F2F2}\begin{tabular}[c]{@{}l@{}}Big: RGB \\ Small: RGB\end{tabular} &
  \multicolumn{1}{l|}{\cellcolor[HTML]{F2F2F2}\begin{tabular}[c]{@{}l@{}}Big: \(72\times72\)\\ Small: \(9\times9\)\end{tabular}} &
  \cellcolor[HTML]{F9E3D7}4.964 &
  {\color[HTML]{808080} 0.20} &
  \cellcolor[HTML]{FAE3D7}6.389 &
  {\color[HTML]{808080} 2.68} &
  \cellcolor[HTML]{F9E3D7}30.460 &
  {\color[HTML]{808080} 1.37} &
  \cellcolor[HTML]{F9E2D7}0.048 &
  {\color[HTML]{808080} 0.05} &
  \cellcolor[HTML]{FBE2D5}4.040 &
  {\color[HTML]{808080} 0.38} &
  \cellcolor[HTML]{FBE2D5}0.533 &
  \multicolumn{1}{c|}{{\color[HTML]{808080} 0.01}} &
  \cellcolor[HTML]{DCE9F6}2.861 &
  {\color[HTML]{808080} 0.29} &
  \cellcolor[HTML]{DCE9F6}6.583 &
  {\color[HTML]{808080} 0.73} &
  \cellcolor[HTML]{DCE9F6}3.494 &
  {\color[HTML]{808080} 0.37} &
  \cellcolor[HTML]{DDE8F5}0.884 &
  {\color[HTML]{808080} 0.02} &
  \cellcolor[HTML]{E0E7F1}6.600 &
  {\color[HTML]{808080} 0.38} &
  \cellcolor[HTML]{E3E7EF}0.668 &
  {\color[HTML]{808080} 0.01} \\ \cline{2-27} 
 &
  \cellcolor[HTML]{D9D9D9}\begin{tabular}[c]{@{}l@{}}Big: Thermal   \\ Small: Thermal\end{tabular} &
  \multicolumn{1}{l|}{\cellcolor[HTML]{D9D9D9}\begin{tabular}[c]{@{}l@{}}Big: \(72\times72\)\\ Small: \(9\times9\)\end{tabular}} &
  \cellcolor[HTML]{E4E7ED}3.745 &
  {\color[HTML]{808080} 0.20} &
  \cellcolor[HTML]{E9E6E8}5.617 &
  {\color[HTML]{808080} 2.67} &
  \cellcolor[HTML]{EEE5E3}27.552 &
  {\color[HTML]{808080} 1.79} &
  \cellcolor[HTML]{E1E7F0}0.296 &
  {\color[HTML]{808080} 0.05} &
  \cellcolor[HTML]{DBE8F7}8.818 &
  {\color[HTML]{808080} 0.44} &
  \cellcolor[HTML]{E1E7F1}0.663 &
  \multicolumn{1}{c|}{{\color[HTML]{808080} 0.01}} &
  \cellcolor[HTML]{FBE2D5}18.646 &
  {\color[HTML]{808080} 0.83} &
  \cellcolor[HTML]{FAE3D6}25.248 &
  {\color[HTML]{808080} 1.52} &
  \cellcolor[HTML]{FBE2D5}21.797 &
  {\color[HTML]{808080} 0.94} &
  \cellcolor[HTML]{F7E2D9}0.146 &
  {\color[HTML]{808080} 0.05} &
  \cellcolor[HTML]{FBE2D5}-6.391 &
  {\color[HTML]{808080} 0.28} &
  \cellcolor[HTML]{FBE2D5}0.314 &
  {\color[HTML]{808080} 0.01} \\ \cline{2-27} 
\multirow{-3}{*}{BigSmall \cite{narayanswamy2024bigsmall}} &
  \cellcolor[HTML]{BFBFBF}\begin{tabular}[c]{@{}l@{}}Big: Thermal   \\ Small: RGB\end{tabular} &
  \multicolumn{1}{l|}{\cellcolor[HTML]{BFBFBF}\begin{tabular}[c]{@{}l@{}}Big: \(72\times72\)\\ Small: \(9\times9\)\end{tabular}} &
  \cellcolor[HTML]{FBE2D5}5.044 &
  {\color[HTML]{808080} 0.20} &
  \cellcolor[HTML]{FBE2D5}6.432 &
  {\color[HTML]{808080} 2.64} &
  \cellcolor[HTML]{FBE2D5}30.790 &
  {\color[HTML]{808080} 1.34} &
  \cellcolor[HTML]{FBE2D5}0.018 &
  {\color[HTML]{808080} 0.05} &
  \cellcolor[HTML]{FBE2D5}3.921 &
  {\color[HTML]{808080} 0.39} &
  \cellcolor[HTML]{FBE2D5}0.535 &
  \multicolumn{1}{c|}{{\color[HTML]{808080} 0.01}} &
  \cellcolor[HTML]{DBE9F7}2.730 &
  {\color[HTML]{808080} 0.29} &
  \cellcolor[HTML]{DCE9F6}6.580 &
  {\color[HTML]{808080} 0.75} &
  \cellcolor[HTML]{DCE9F6}3.377 &
  {\color[HTML]{808080} 0.39} &
  \cellcolor[HTML]{DDE8F5}0.881 &
  {\color[HTML]{808080} 0.02} &
  \cellcolor[HTML]{E0E7F1}6.641 &
  {\color[HTML]{808080} 0.37} &
  \cellcolor[HTML]{E3E7EF}0.668 &
  {\color[HTML]{808080} 0.01} \\ \hline
 &
  \cellcolor[HTML]{F2F2F2} &
  \multicolumn{1}{l|}{\cellcolor[HTML]{F2F2F2}\(72\times72\)} &
  \cellcolor[HTML]{F1E5E0}4.487 &
  {\color[HTML]{808080} 0.20} &
  \cellcolor[HTML]{F1E4DF}6.015 &
  {\color[HTML]{808080} 0.35} &
  \cellcolor[HTML]{EFE5E2}27.777 &
  {\color[HTML]{808080} 1.40} &
  \cellcolor[HTML]{F4E3DC}0.092 &
  {\color[HTML]{808080} 0.05} &
  \cellcolor[HTML]{F7E2D9}4.612 &
  {\color[HTML]{808080} 0.36} &
  \cellcolor[HTML]{F5E3DC}0.565 &
  \multicolumn{1}{c|}{{\color[HTML]{808080} 0.01}} &
  \cellcolor[HTML]{CBEEDC}1.721 &
  {\color[HTML]{808080} 0.22} &
  \cellcolor[HTML]{D5EBEF}4.739 &
  {\color[HTML]{808080} 0.55} &
  \cellcolor[HTML]{CAEEDA}2.001 &
  {\color[HTML]{808080} 0.25} &
  \cellcolor[HTML]{D4EAED}0.942 &
  {\color[HTML]{808080} 0.02} &
  \cellcolor[HTML]{C5EFCF}9.535 &
  {\color[HTML]{808080} 0.45} &
  \cellcolor[HTML]{C6EED2}\textbf{0.792} &
  {\color[HTML]{808080} 0.01} \\
 &
  \cellcolor[HTML]{F2F2F2} &
  \multicolumn{1}{l|}{\cellcolor[HTML]{F2F2F2}\(36\times36\)} &
  \cellcolor[HTML]{F1E4DF}4.515 &
  {\color[HTML]{808080} 0.20} &
  \cellcolor[HTML]{F3E4DD}6.091 &
  {\color[HTML]{808080} 0.36} &
  \cellcolor[HTML]{F0E5E1}28.103 &
  {\color[HTML]{808080} 1.44} &
  \cellcolor[HTML]{F5E3DB}0.084 &
  {\color[HTML]{808080} 0.05} &
  \cellcolor[HTML]{F6E3DA}4.727 &
  {\color[HTML]{808080} 0.35} &
  \cellcolor[HTML]{F4E3DC}0.567 &
  \multicolumn{1}{c|}{{\color[HTML]{808080} 0.01}} &
  \cellcolor[HTML]{D6EAF2}1.772 &
  {\color[HTML]{808080} 0.22} &
  \cellcolor[HTML]{D9EAF7}4.789 &
  {\color[HTML]{808080} 0.55} &
  \cellcolor[HTML]{D4EBED}2.055 &
  {\color[HTML]{808080} 0.25} &
  \cellcolor[HTML]{D7E9F3}0.941 &
  {\color[HTML]{808080} 0.02} &
  \cellcolor[HTML]{C9EDD7}9.513 &
  {\color[HTML]{808080} 0.46} &
  \cellcolor[HTML]{C4EFCC}\textbf{0.792} &
  {\color[HTML]{808080} 0.01} \\
 &
  \multirow{-3}{*}{\cellcolor[HTML]{F2F2F2}\begin{tabular}[c]{@{}l@{}}RSP: RGB\\ BVP: RGB\end{tabular}} &
  \multicolumn{1}{l|}{\cellcolor[HTML]{F2F2F2}\(9\times9\)} &
  \cellcolor[HTML]{EBE6E6}4.133 &
  {\color[HTML]{808080} 0.20} &
  \cellcolor[HTML]{EBE6E5}5.745 &
  {\color[HTML]{808080} 0.35} &
  \cellcolor[HTML]{E9E6E8}26.453 &
  {\color[HTML]{808080} 1.44} &
  \cellcolor[HTML]{EEE4E3}0.158 &
  {\color[HTML]{808080} 0.05} &
  \cellcolor[HTML]{EFE4E2}5.803 &
  {\color[HTML]{808080} 0.38} &
  \cellcolor[HTML]{F0E4E1}0.589 &
  \multicolumn{1}{c|}{{\color[HTML]{808080} 0.01}} &
  \cellcolor[HTML]{C9EED7}1.711 &
  {\color[HTML]{808080} 0.20} &
  \cellcolor[HTML]{C1F0C8}\textbf{4.471} &
  {\color[HTML]{808080} 0.48} &
  \cellcolor[HTML]{CAEEDA}2.002 &
  {\color[HTML]{808080} 0.24} &
  \cellcolor[HTML]{CCEDDD}0.944 &
  {\color[HTML]{808080} 0.02} &
  \cellcolor[HTML]{D2EBE8}9.470 &
  {\color[HTML]{808080} 0.46} &
  \cellcolor[HTML]{DBE8F7}0.789 &
  {\color[HTML]{808080} 0.01} \\ \cline{2-27} 
 &
  \cellcolor[HTML]{D9D9D9} &
  \multicolumn{1}{l|}{\cellcolor[HTML]{D9D9D9}\(72\times72\)} &
  \cellcolor[HTML]{C7EFD5}2.994 &
  {\color[HTML]{808080} 0.19} &
  \cellcolor[HTML]{CDEDE0}4.840 &
  {\color[HTML]{808080} 0.33} &
  \cellcolor[HTML]{C6EFD2}21.500 &
  {\color[HTML]{808080} 1.57} &
  \cellcolor[HTML]{CBEDDA}0.386 &
  {\color[HTML]{808080} 0.05} &
  \cellcolor[HTML]{D0ECE4}9.022 &
  {\color[HTML]{808080} 0.44} &
  \cellcolor[HTML]{CAEDD8}0.701 &
  \multicolumn{1}{c|}{{\color[HTML]{808080} 0.01}} &
  \cellcolor[HTML]{F7E3D9}16.825 &
  {\color[HTML]{808080} 0.93} &
  \cellcolor[HTML]{FAE3D6}25.353 &
  {\color[HTML]{808080} 1.69} &
  \cellcolor[HTML]{FAE3D6}21.411 &
  {\color[HTML]{808080} 1.28} &
  \cellcolor[HTML]{FBE2D5}0.035 &
  {\color[HTML]{808080} 0.05} &
  \cellcolor[HTML]{F7E2D9}-4.167 &
  {\color[HTML]{808080} 0.26} &
  \cellcolor[HTML]{F9E2D8}0.355 &
  {\color[HTML]{808080} 0.01} \\
 &
  \cellcolor[HTML]{D9D9D9} &
  \multicolumn{1}{l|}{\cellcolor[HTML]{D9D9D9}\(36\times36\)} &
  \cellcolor[HTML]{C1F0C8}\textbf{2.950} &
  {\color[HTML]{808080} 0.18} &
  \cellcolor[HTML]{C1F0C8}\textbf{4.751} &
  {\color[HTML]{808080} 0.32} &
  \cellcolor[HTML]{C1F0C8}\textbf{21.244} &
  {\color[HTML]{808080} 1.54} &
  \cellcolor[HTML]{C2EFCA}0.397 &
  {\color[HTML]{808080} 0.04} &
  \cellcolor[HTML]{CBEDDB}9.057 &
  {\color[HTML]{808080} 0.45} &
  \cellcolor[HTML]{C8EED5}0.702 &
  \multicolumn{1}{c|}{{\color[HTML]{808080} 0.01}} &
  \cellcolor[HTML]{F5E4DC}15.608 &
  {\color[HTML]{808080} 0.90} &
  \cellcolor[HTML]{F8E3D8}24.060 &
  {\color[HTML]{808080} 1.65} &
  \cellcolor[HTML]{F7E3D9}19.800 &
  {\color[HTML]{808080} 1.23} &
  \cellcolor[HTML]{FAE2D6}0.087 &
  {\color[HTML]{808080} 0.05} &
  \cellcolor[HTML]{F7E3DA}-4.086 &
  {\color[HTML]{808080} 0.26} &
  \cellcolor[HTML]{F9E2D8}0.357 &
  {\color[HTML]{808080} 0.01} \\
 &
  \multirow{-3}{*}{\cellcolor[HTML]{D9D9D9}\begin{tabular}[c]{@{}l@{}}RSP: Thermal\\ BVP: Thermal\end{tabular}} &
  \multicolumn{1}{l|}{\cellcolor[HTML]{D9D9D9}\(9\times9\)} &
  \cellcolor[HTML]{D0ECE5}3.046 &
  {\color[HTML]{808080} 0.18} &
  \cellcolor[HTML]{CBEEDC}4.826 &
  {\color[HTML]{808080} 0.32} &
  \cellcolor[HTML]{CAEEDA}21.693 &
  {\color[HTML]{808080} 1.54} &
  \cellcolor[HTML]{C9EDD7}0.388 &
  {\color[HTML]{808080} 0.05} &
  \cellcolor[HTML]{DAE9F8}8.937 &
  {\color[HTML]{808080} 0.43} &
  \cellcolor[HTML]{D9E9F6}0.694 &
  \multicolumn{1}{c|}{{\color[HTML]{808080} 0.01}} &
  \cellcolor[HTML]{F5E4DB}15.888 &
  {\color[HTML]{808080} 0.92} &
  \cellcolor[HTML]{F9E3D7}24.586 &
  {\color[HTML]{808080} 1.69} &
  \cellcolor[HTML]{F8E3D8}20.190 &
  {\color[HTML]{808080} 1.26} &
  \cellcolor[HTML]{F9E2D7}0.092 &
  {\color[HTML]{808080} 0.05} &
  \cellcolor[HTML]{F7E3DA}-4.011 &
  {\color[HTML]{808080} 0.27} &
  \cellcolor[HTML]{F8E2D8}0.360 &
  {\color[HTML]{808080} 0.01} \\ \cline{2-27} 
 &
  \cellcolor[HTML]{BFBFBF} &
  \multicolumn{1}{l|}{\cellcolor[HTML]{BFBFBF}\(72\times72\)} &
  \cellcolor[HTML]{D5EBEE}3.076 &
  {\color[HTML]{808080} 0.19} &
  \cellcolor[HTML]{D3EBEB}4.880 &
  {\color[HTML]{808080} 0.33} &
  \cellcolor[HTML]{D4EBEC}22.145 &
  {\color[HTML]{808080} 1.58} &
  \cellcolor[HTML]{DAE9F8}0.365 &
  {\color[HTML]{808080} 0.05} &
  \cellcolor[HTML]{C8EED4}9.086 &
  {\color[HTML]{808080} 0.44} &
  \cellcolor[HTML]{C8EED5}0.702 &
  \multicolumn{1}{c|}{{\color[HTML]{808080} 0.01}} &
  \cellcolor[HTML]{C3F0CD}1.687 &
  {\color[HTML]{808080} 0.21} &
  \cellcolor[HTML]{C9EED8}4.585 &
  {\color[HTML]{808080} 0.53} &
  \cellcolor[HTML]{C6EFD2}1.980 &
  {\color[HTML]{808080} 0.25} &
  \cellcolor[HTML]{C8EED5}0.945 &
  {\color[HTML]{808080} 0.02} &
  \cellcolor[HTML]{C2EFC9}\textbf{9.551} &
  {\color[HTML]{808080} 0.45} &
  \cellcolor[HTML]{C5EFCF}\textbf{0.792} &
  {\color[HTML]{808080} 0.01} \\
 &
  \cellcolor[HTML]{BFBFBF} &
  \multicolumn{1}{l|}{\cellcolor[HTML]{BFBFBF}\(36\times36\)} &
  \cellcolor[HTML]{C6EFD3}2.988 &
  {\color[HTML]{808080} 0.18} &
  \cellcolor[HTML]{C3F0CB}4.765 &
  {\color[HTML]{808080} 0.32} &
  \cellcolor[HTML]{C3F0CD}21.374 &
  {\color[HTML]{808080} 1.52} &
  \cellcolor[HTML]{C9EDD7}0.388 &
  {\color[HTML]{808080} 0.05} &
  \cellcolor[HTML]{C3EFCC}9.122 &
  {\color[HTML]{808080} 0.44} &
  \cellcolor[HTML]{C7EED4}0.702 &
  \multicolumn{1}{c|}{{\color[HTML]{808080} 0.01}} &
  \cellcolor[HTML]{CCEDDE}1.727 &
  {\color[HTML]{808080} 0.21} &
  \cellcolor[HTML]{D3EBEC}4.713 &
  {\color[HTML]{808080} 0.55} &
  \cellcolor[HTML]{C9EED8}1.997 &
  {\color[HTML]{808080} 0.25} &
  \cellcolor[HTML]{D0ECE4}0.943 &
  {\color[HTML]{808080} 0.02} &
  \cellcolor[HTML]{C8EED4}9.521 &
  {\color[HTML]{808080} 0.45} &
  \cellcolor[HTML]{C8EED4}0.791 &
  {\color[HTML]{808080} 0.01} \\
\multirow{-9}{*}{\begin{tabular}[c]{@{}l@{}}\textbf{\Model{}}\\ with FSAM \dag\end{tabular}} &
  \multirow{-3}{*}{\cellcolor[HTML]{BFBFBF}\begin{tabular}[c]{@{}l@{}}RSP: Thermal\\ BVP: RGB\end{tabular}} &
  \multicolumn{1}{l|}{\cellcolor[HTML]{BFBFBF}\(9\times9\)} &
  \cellcolor[HTML]{D1ECE8}3.055 &
  {\color[HTML]{808080} 0.19} &
  \cellcolor[HTML]{D2ECE9}4.874 &
  {\color[HTML]{808080} 0.33} &
  \cellcolor[HTML]{CDEDDF}21.819 &
  {\color[HTML]{808080} 1.56} &
  \cellcolor[HTML]{D0EBE5}0.379 &
  {\color[HTML]{808080} 0.05} &
  \cellcolor[HTML]{DBE8F7}8.879 &
  {\color[HTML]{808080} 0.43} &
  \cellcolor[HTML]{DBE8F7}0.692 &
  \multicolumn{1}{c|}{{\color[HTML]{808080} 0.01}} &
  \cellcolor[HTML]{DAE9F8}1.786 &
  {\color[HTML]{808080} 0.21} &
  \cellcolor[HTML]{CDEDE0}4.634 &
  {\color[HTML]{808080} 0.50} &
  \cellcolor[HTML]{DAE9F8}2.093 &
  {\color[HTML]{808080} 0.25} &
  \cellcolor[HTML]{DBE8F7}0.940 &
  {\color[HTML]{808080} 0.32} &
  \cellcolor[HTML]{DBE8F7}9.422 &
  {\color[HTML]{808080} 0.46} &
  \cellcolor[HTML]{DAE9F8}0.789 &
  {\color[HTML]{808080} 0.01} \\ \hline
 &
  \cellcolor[HTML]{F2F2F2} &
  \multicolumn{1}{l|}{\cellcolor[HTML]{F2F2F2}\(72\times72\)} &
  \cellcolor[HTML]{F2E4DE}4.566 &
  {\color[HTML]{808080} 0.20} &
  \cellcolor[HTML]{F3E4DE}6.081 &
  {\color[HTML]{808080} 0.36} &
  \cellcolor[HTML]{EFE5E1}27.944 &
  {\color[HTML]{808080} 1.35} &
  \cellcolor[HTML]{F3E3DE}0.110 &
  {\color[HTML]{808080} 0.05} &
  \cellcolor[HTML]{F9E2D7}4.289 &
  {\color[HTML]{808080} 0.36} &
  \cellcolor[HTML]{F7E2D9}0.555 &
  \multicolumn{1}{c|}{{\color[HTML]{808080} 0.01}} &
  \cellcolor[HTML]{C8EFD5}1.706 &
  {\color[HTML]{808080} 0.21} &
  \cellcolor[HTML]{D3EBEB}4.707 &
  {\color[HTML]{808080} 0.55} &
  \cellcolor[HTML]{C7EFD3}1.983 &
  {\color[HTML]{808080} 0.25} &
  \cellcolor[HTML]{D1EBE7}0.943 &
  {\color[HTML]{808080} 0.02} &
  \cellcolor[HTML]{C4EFCD}9.539 &
  {\color[HTML]{808080} 0.45} &
  \cellcolor[HTML]{C1F0C8}\textbf{0.792} &
  {\color[HTML]{808080} 0.01} \\
 &
  \cellcolor[HTML]{F2F2F2} &
  \multicolumn{1}{l|}{\cellcolor[HTML]{F2F2F2}\(36\times36\)} &
  \cellcolor[HTML]{F5E4DB}4.721 &
  {\color[HTML]{808080} 0.20} &
  \cellcolor[HTML]{F6E3DA}6.245 &
  {\color[HTML]{808080} 0.36} &
  \cellcolor[HTML]{F5E4DC}29.338 &
  {\color[HTML]{808080} 1.42} &
  \cellcolor[HTML]{F8E2D9}0.058 &
  {\color[HTML]{808080} 0.05} &
  \cellcolor[HTML]{F9E2D7}4.301 &
  {\color[HTML]{808080} 0.34} &
  \cellcolor[HTML]{F9E2D8}0.547 &
  \multicolumn{1}{c|}{{\color[HTML]{808080} 0.01}} &
  \cellcolor[HTML]{DAE9F8}1.789 &
  {\color[HTML]{808080} 0.22} &
  \cellcolor[HTML]{DAE9F8}4.835 &
  {\color[HTML]{808080} 0.55} &
  \cellcolor[HTML]{D6EAF1}2.068 &
  {\color[HTML]{808080} 0.25} &
  \cellcolor[HTML]{DAE9F8}0.940 &
  {\color[HTML]{808080} 0.02} &
  \cellcolor[HTML]{CAEDD9}9.507 &
  {\color[HTML]{808080} 0.45} &
  \cellcolor[HTML]{C2EFC9}\textbf{0.792} &
  {\color[HTML]{808080} 0.01} \\
 &
  \multirow{-3}{*}{\cellcolor[HTML]{F2F2F2}\begin{tabular}[c]{@{}l@{}}RSP: RGB\\ BVP: RGB\end{tabular}} &
  \multicolumn{1}{l|}{\cellcolor[HTML]{F2F2F2}\(9\times9\)} &
  \cellcolor[HTML]{ECE6E5}4.166 &
  {\color[HTML]{808080} 0.19} &
  \cellcolor[HTML]{EBE6E6}5.744 &
  {\color[HTML]{808080} 0.34} &
  \cellcolor[HTML]{EAE6E7}26.722 &
  {\color[HTML]{808080} 1.44} &
  \cellcolor[HTML]{EEE4E3}0.164 &
  {\color[HTML]{808080} 0.05} &
  \cellcolor[HTML]{F0E4E1}5.643 &
  {\color[HTML]{808080} 0.37} &
  \cellcolor[HTML]{F1E4E0}0.584 &
  \multicolumn{1}{c|}{{\color[HTML]{808080} 0.01}} &
  \cellcolor[HTML]{CBEEDB}1.719 &
  {\color[HTML]{808080} 0.20} &
  \cellcolor[HTML]{C2F0CA}4.485 &
  {\color[HTML]{808080} 0.48} &
  \cellcolor[HTML]{CCEDDD}2.011 &
  {\color[HTML]{808080} 0.24} &
  \cellcolor[HTML]{CEECE0}0.944 &
  {\color[HTML]{808080} 0.02} &
  \cellcolor[HTML]{D3EBEB}9.463 &
  {\color[HTML]{808080} 0.46} &
  \cellcolor[HTML]{D5EAEE}0.790 &
  {\color[HTML]{808080} 0.01} \\ \cline{2-27} 
 &
  \cellcolor[HTML]{D9D9D9} &
  \multicolumn{1}{l|}{\cellcolor[HTML]{D9D9D9}\(72\times72\)} &
  \cellcolor[HTML]{D2ECE9}3.060 &
  {\color[HTML]{808080} 0.19} &
  \cellcolor[HTML]{D9EAF6}4.920 &
  {\color[HTML]{808080} 0.34} &
  \cellcolor[HTML]{D5EBEE}22.189 &
  {\color[HTML]{808080} 1.61} &
  \cellcolor[HTML]{CEECE0}0.382 &
  {\color[HTML]{808080} 0.05} &
  \cellcolor[HTML]{CDECDE}9.047 &
  {\color[HTML]{808080} 0.44} &
  \cellcolor[HTML]{C5EED0}0.703 &
  \multicolumn{1}{c|}{{\color[HTML]{808080} 0.01}} &
  \cellcolor[HTML]{F7E3D9}16.774 &
  {\color[HTML]{808080} 0.95} &
  \cellcolor[HTML]{FBE2D5}25.683 &
  {\color[HTML]{808080} 1.73} &
  \cellcolor[HTML]{FAE3D6}21.369 &
  {\color[HTML]{808080} 1.32} &
  \cellcolor[HTML]{FAE2D6}0.063 &
  {\color[HTML]{808080} 0.05} &
  \cellcolor[HTML]{F7E3DA}-4.072 &
  {\color[HTML]{808080} 0.26} &
  \cellcolor[HTML]{F8E2D8}0.358 &
  {\color[HTML]{808080} 0.01} \\
 &
  \cellcolor[HTML]{D9D9D9} &
  \multicolumn{1}{l|}{\cellcolor[HTML]{D9D9D9}\(36\times36\)} &
  \cellcolor[HTML]{D3EBEB}3.067 &
  {\color[HTML]{808080} 0.19} &
  \cellcolor[HTML]{D8EAF4}4.914 &
  {\color[HTML]{808080} 0.33} &
  \cellcolor[HTML]{D3EBEB}22.103 &
  {\color[HTML]{808080} 1.58} &
  \cellcolor[HTML]{D8E9F4}0.368 &
  {\color[HTML]{808080} 0.05} &
  \cellcolor[HTML]{D2EBE9}9.002 &
  {\color[HTML]{808080} 0.44} &
  \cellcolor[HTML]{C6EED2}0.703 &
  \multicolumn{1}{c|}{{\color[HTML]{808080} 0.01}} &
  \cellcolor[HTML]{F6E3DA}16.468 &
  {\color[HTML]{808080} 0.92} &
  \cellcolor[HTML]{F9E3D7}24.957 &
  {\color[HTML]{808080} 1.67} &
  \cellcolor[HTML]{F9E3D7}20.804 &
  {\color[HTML]{808080} 1.25} &
  \cellcolor[HTML]{FBE2D5}0.044 &
  {\color[HTML]{808080} 0.05} &
  \cellcolor[HTML]{F7E3DA}-4.089 &
  {\color[HTML]{808080} 0.26} &
  \cellcolor[HTML]{F9E2D8}0.356 &
  {\color[HTML]{808080} 0.01} \\
 &
  \multirow{-3}{*}{\cellcolor[HTML]{D9D9D9}\begin{tabular}[c]{@{}l@{}}RSP: Thermal\\ BVP: Thermal\end{tabular}} &
  \multicolumn{1}{l|}{\cellcolor[HTML]{D9D9D9}\(9\times9\)} &
  \cellcolor[HTML]{DAE9F8}3.128 &
  {\color[HTML]{808080} 0.19} &
  \cellcolor[HTML]{DAE9F8}4.927 &
  {\color[HTML]{808080} 0.33} &
  \cellcolor[HTML]{DAE9F8}22.469 &
  {\color[HTML]{808080} 1.58} &
  \cellcolor[HTML]{D6EAF0}0.371 &
  {\color[HTML]{808080} 0.05} &
  \cellcolor[HTML]{D5EAED}8.984 &
  {\color[HTML]{808080} 0.43} &
  \cellcolor[HTML]{D7E9F2}0.695 &
  \multicolumn{1}{c|}{{\color[HTML]{808080} 0.01}} &
  \cellcolor[HTML]{F5E4DC}15.753 &
  {\color[HTML]{808080} 0.92} &
  \cellcolor[HTML]{F9E3D7}24.574 &
  {\color[HTML]{808080} 1.71} &
  \cellcolor[HTML]{F8E3D9}20.033 &
  {\color[HTML]{808080} 1.27} &
  \cellcolor[HTML]{F9E2D7}0.111 &
  {\color[HTML]{808080} 0.05} &
  \cellcolor[HTML]{F6E3DA}-3.977 &
  {\color[HTML]{808080} 0.27} &
  \cellcolor[HTML]{F8E2D8}0.361 &
  {\color[HTML]{808080} 0.01} \\ \cline{2-27} 
 &
  \cellcolor[HTML]{BFBFBF} &
  \multicolumn{1}{l|}{\cellcolor[HTML]{BFBFBF}\(72\times72\)} &
  \cellcolor[HTML]{D4EBED}3.073 &
  {\color[HTML]{808080} 0.19} &
  \cellcolor[HTML]{D7EAF2}4.906 &
  {\color[HTML]{808080} 0.33} &
  \cellcolor[HTML]{D5EBEF}22.215 &
  {\color[HTML]{808080} 1.59} &
  \cellcolor[HTML]{C4EFCE}0.395 &
  {\color[HTML]{808080} 0.04} &
  \cellcolor[HTML]{C1F0C8}\textbf{9.134} &
  {\color[HTML]{808080} 0.45} &
  \cellcolor[HTML]{C2EFCA}\textbf{0.705} &
  \multicolumn{1}{c|}{{\color[HTML]{808080} 0.01}} &
  \cellcolor[HTML]{C1F0C8}\textbf{1.674} &
  {\color[HTML]{808080} 0.20} &
  \cellcolor[HTML]{C2F0CA}4.486 &
  {\color[HTML]{808080} 0.51} &
  \cellcolor[HTML]{C1F0C8}\textbf{1.949} &
  {\color[HTML]{808080} 0.24} &
  \cellcolor[HTML]{C1F0C8}\textbf{0.947} &
  {\color[HTML]{808080} 0.01} &
  \cellcolor[HTML]{C1F0C8}\textbf{9.551} &
  {\color[HTML]{808080} 0.30} &
  \cellcolor[HTML]{C7EED4}0.791 &
  {\color[HTML]{808080} 0.01} \\
 &
  \cellcolor[HTML]{BFBFBF} &
  \multicolumn{1}{l|}{\cellcolor[HTML]{BFBFBF}\(36\times36\)} &
  \cellcolor[HTML]{C6EFD2}2.983 &
  {\color[HTML]{808080} 0.19} &
  \cellcolor[HTML]{CBEEDB}4.823 &
  {\color[HTML]{808080} 0.33} &
  \cellcolor[HTML]{C6EFD2}21.506 &
  {\color[HTML]{808080} 1.55} &
  \cellcolor[HTML]{C1F0C8}\textbf{0.398} &
  {\color[HTML]{808080} 0.04} &
  \cellcolor[HTML]{C2EFC9}\textbf{9.134} &
  {\color[HTML]{808080} 0.44} &
  \cellcolor[HTML]{C1F0C8}\textbf{0.705} &
  \multicolumn{1}{c|}{{\color[HTML]{808080} 0.01}} &
  \cellcolor[HTML]{D8EAF4}1.778 &
  {\color[HTML]{808080} 0.22} &
  \cellcolor[HTML]{DAE9F8}4.794 &
  {\color[HTML]{808080} 0.55} &
  \cellcolor[HTML]{D5EBF0}2.063 &
  {\color[HTML]{808080} 0.25} &
  \cellcolor[HTML]{D6EAF0}0.941 &
  {\color[HTML]{808080} 0.02} &
  \cellcolor[HTML]{CAEDDA}9.507 &
  {\color[HTML]{808080} 0.45} &
  \cellcolor[HTML]{C9EDD7}0.791 &
  {\color[HTML]{808080} 0.01} \\
\multirow{-9}{*}{\begin{tabular}[c]{@{}l@{}}\textbf{\Model{}}\\ with \textbf{\AM{}}\end{tabular}} &
  \multirow{-3}{*}{\cellcolor[HTML]{BFBFBF}\begin{tabular}[c]{@{}l@{}}RSP: Thermal\\ BVP: RGB\end{tabular}} &
  \multicolumn{1}{l|}{\cellcolor[HTML]{BFBFBF}\(9\times9\)} &
  \cellcolor[HTML]{DAE9F8}3.107 &
  {\color[HTML]{808080} 0.19} &
  \cellcolor[HTML]{DAE9F8}4.966 &
  {\color[HTML]{808080} 0.33} &
  \cellcolor[HTML]{DAE9F8}22.417 &
  {\color[HTML]{808080} 1.62} &
  \cellcolor[HTML]{DBE8F7}0.356 &
  {\color[HTML]{808080} 0.05} &
  \cellcolor[HTML]{D5EAED}8.983 &
  {\color[HTML]{808080} 0.43} &
  \cellcolor[HTML]{DAE9F8}0.693 &
  \multicolumn{1}{c|}{{\color[HTML]{808080} 0.01}} &
  \cellcolor[HTML]{D9EAF7}1.785 &
  {\color[HTML]{808080} 0.21} &
  \cellcolor[HTML]{C8EED6}4.571 &
  {\color[HTML]{808080} 0.48} &
  \cellcolor[HTML]{DAE9F8}2.086 &
  {\color[HTML]{808080} 0.24} &
  \cellcolor[HTML]{D7EAF1}0.941 &
  {\color[HTML]{808080} 0.02} &
  \cellcolor[HTML]{DAE9F8}9.428 &
  {\color[HTML]{808080} 0.46} &
  \cellcolor[HTML]{DAE9F7}0.789 &
  {\color[HTML]{808080} 0.01} \\ \hline \hline
\multicolumn{27}{l}{\begin{tabular}[c]{@{}l@{}}Cross Dataset Performance Evaluation\\ Training on SCAMPS Dataset, Testing on BP4D+ Dataset\end{tabular}} \\ \hline \hline
BigSmall \cite{narayanswamy2024bigsmall} &
  \cellcolor[HTML]{F2F2F2}\begin{tabular}[c]{@{}l@{}}Big: RGB \\ Small: RGB\end{tabular} &
  \multicolumn{1}{l|}{\cellcolor[HTML]{F2F2F2}\begin{tabular}[c]{@{}l@{}}Big: \(72\times72\)\\ Small: \(9\times9\)\end{tabular}} &
  \cellcolor[HTML]{FBE2D5}5.243 &
  {\color[HTML]{808080} 0.11} &
  \cellcolor[HTML]{FBE2D5}6.613 &
  {\color[HTML]{808080} 1.61} &
  \cellcolor[HTML]{FBE2D5}34.760 &
  {\color[HTML]{808080} 0.94} &
  \cellcolor[HTML]{FBE2D5}0.005 &
  {\color[HTML]{808080} 0.03} &
  \cellcolor[HTML]{FBE2D5}4.511 &
  {\color[HTML]{808080} 0.18} &
  \cellcolor[HTML]{E2E7F0}0.523 &
  \multicolumn{1}{c|}{{\color[HTML]{808080} 0.00}} &
  \cellcolor[HTML]{FBE2D5}9.472 &
  {\color[HTML]{808080} 0.45} &
  \cellcolor[HTML]{FBE2D5}18.488 &
  {\color[HTML]{808080} 0.87} &
  \cellcolor[HTML]{FBE2D5}11.460 &
  {\color[HTML]{808080} 0.56} &
  \cellcolor[HTML]{FBE2D5}0.435 &
  {\color[HTML]{808080} 0.03} &
  \cellcolor[HTML]{FBE2D5}0.461 &
  {\color[HTML]{808080} 0.23} &
  \cellcolor[HTML]{FBE2D5}0.443 &
  {\color[HTML]{808080} 0.00} \\ \hline
 &
  \cellcolor[HTML]{F2F2F2} &
  \multicolumn{1}{l|}{\cellcolor[HTML]{F2F2F2}\(72\times72\)} &
  \cellcolor[HTML]{C1F0C8}\textbf{4.657} &
  {\color[HTML]{808080} 0.11} &
  \cellcolor[HTML]{C1F0C8}\textbf{6.111} &
  {\color[HTML]{808080} 0.21} &
  \cellcolor[HTML]{C1F0C8}\textbf{31.386} &
  {\color[HTML]{808080} 0.96} &
  \cellcolor[HTML]{C1F0C8}\textbf{0.093} &
  {\color[HTML]{808080} 0.03} &
  \cellcolor[HTML]{D8E9F4}4.821 &
  {\color[HTML]{808080} 0.17} &
  \cellcolor[HTML]{C1F0C8}\textbf{0.526} &
  \multicolumn{1}{c|}{{\color[HTML]{808080} 0.00}} &
  \cellcolor[HTML]{CBEEDB}2.421 &
  {\color[HTML]{808080} 0.17} &
  \cellcolor[HTML]{D4EBEC}6.485 &
  {\color[HTML]{808080} 0.45} &
  \cellcolor[HTML]{D2ECE9}2.983 &
  {\color[HTML]{808080} 0.23} &
  \cellcolor[HTML]{D5EAEE}0.893 &
  {\color[HTML]{808080} 0.01} &
  \cellcolor[HTML]{C4EFCD}7.042 &
  {\color[HTML]{808080} 0.22} &
  \cellcolor[HTML]{C5EFCE}0.687 &
  {\color[HTML]{808080} 0.00} \\
\multirow{-2}{*}{\begin{tabular}[c]{@{}l@{}}\textbf{\Model{}}\\ with FSAM \dag\end{tabular}} &
  \multirow{-2}{*}{\cellcolor[HTML]{F2F2F2}\begin{tabular}[c]{@{}l@{}}RSP: RGB\\ BVP: RGB\end{tabular}} &
  \multicolumn{1}{l|}{\cellcolor[HTML]{F2F2F2}\(9\times9\)} &
  \cellcolor[HTML]{DAE9F8}4.896 &
  {\color[HTML]{808080} 0.01} &
  \cellcolor[HTML]{E3E7EE}6.422 &
  {\color[HTML]{808080} 0.22} &
  \cellcolor[HTML]{E9E6E8}33.369 &
  {\color[HTML]{808080} 1.01} &
  \cellcolor[HTML]{E5E6ED}0.034 &
  {\color[HTML]{808080} 0.03} &
  \cellcolor[HTML]{C1F0C8}\textbf{5.125} &
  {\color[HTML]{808080} 0.18} &
  \cellcolor[HTML]{D2EBE7}0.525 &
  \multicolumn{1}{c|}{{\color[HTML]{808080} 0.00}} &
  \cellcolor[HTML]{DAE9F8}2.566 &
  {\color[HTML]{808080} 0.17} &
  \cellcolor[HTML]{DAE9F8}6.618 &
  {\color[HTML]{808080} 0.45} &
  \cellcolor[HTML]{DAE9F8}3.046 &
  {\color[HTML]{808080} 0.21} &
  \cellcolor[HTML]{DAE9F8}0.889 &
  {\color[HTML]{808080} 0.01} &
  \cellcolor[HTML]{DAE9F8}6.452 &
  {\color[HTML]{808080} 0.22} &
  \cellcolor[HTML]{DAE9F8}0.670 &
  {\color[HTML]{808080} 0.00} \\ \hline
 &
  \cellcolor[HTML]{F2F2F2} &
  \multicolumn{1}{l|}{\cellcolor[HTML]{F2F2F2}\(72\times72\)} &
  \cellcolor[HTML]{CFECE3}4.796 &
  {\color[HTML]{808080} 0.11} &
  \cellcolor[HTML]{D0ECE5}6.252 &
  {\color[HTML]{808080} 0.21} &
  \cellcolor[HTML]{DAE9F8}32.112 &
  {\color[HTML]{808080} 0.96} &
  \cellcolor[HTML]{D0ECE3}0.067 &
  {\color[HTML]{808080} 0.03} &
  \cellcolor[HTML]{DAE9F8}4.786 &
  {\color[HTML]{808080} 0.18} &
  \cellcolor[HTML]{DAE9F8}0.524 &
  \multicolumn{1}{c|}{{\color[HTML]{808080} 0.00}} &
  \cellcolor[HTML]{C1F0C8}\textbf{2.323} &
  {\color[HTML]{808080} 0.16} &
  \cellcolor[HTML]{C1F0C8}\textbf{6.050} &
  {\color[HTML]{808080} 0.39} &
  \cellcolor[HTML]{C1F0C8}\textbf{2.838} &
  {\color[HTML]{808080} 0.21} &
  \cellcolor[HTML]{C1F0C8}\textbf{0.906} &
  {\color[HTML]{808080} 0.01} &
  \cellcolor[HTML]{C1F0C8}\textbf{7.104} &
  {\color[HTML]{808080} 0.22} &
  \cellcolor[HTML]{C1F0C8}\textbf{0.689} &
  {\color[HTML]{808080} 0.00} \\
\multirow{-2}{*}{\begin{tabular}[c]{@{}l@{}}\textbf{\Model{}}\\ with \textbf{\AM{}}\end{tabular}} &
  \multirow{-2}{*}{\cellcolor[HTML]{F2F2F2}\begin{tabular}[c]{@{}l@{}}RSP: RGB\\ BVP: RGB\end{tabular}} &
  \multicolumn{1}{l|}{\cellcolor[HTML]{F2F2F2}\(9\times9\)} &
  \cellcolor[HTML]{DBE9F7}4.909 &
  {\color[HTML]{808080} 0.11} &
  \cellcolor[HTML]{DAE9F8}6.340 &
  {\color[HTML]{808080} 0.21} &
  \cellcolor[HTML]{D3EBEA}31.910 &
  {\color[HTML]{808080} 0.90} &
  \cellcolor[HTML]{DAE9F8}0.047 &
  {\color[HTML]{808080} 0.03} &
  \cellcolor[HTML]{F5E3DC}4.567 &
  {\color[HTML]{808080} 0.18} &
  \cellcolor[HTML]{FBE2D5}0.517 &
  \multicolumn{1}{c|}{{\color[HTML]{808080} 0.00}} &
  \cellcolor[HTML]{DAE9F8}2.647 &
  {\color[HTML]{808080} 0.17} &
  \cellcolor[HTML]{DAE9F8}6.703 &
  {\color[HTML]{808080} 0.42} &
  \cellcolor[HTML]{DAE9F8}3.205 &
  {\color[HTML]{808080} 0.23} &
  \cellcolor[HTML]{DBE8F7}0.885 &
  {\color[HTML]{808080} 0.01} &
  \cellcolor[HTML]{DCE8F6}6.217 &
  {\color[HTML]{808080} 0.21} &
  \cellcolor[HTML]{DCE8F6}0.663 &
  {\color[HTML]{808080} 0.00} \\ \hline \hline
\end{tabular*}
    \footnotesize
    \begin{flushleft}
    \dag FSAM: Factorized Self-Attention Module \cite{joshi2024factorizephys};  TSFM: Proposed Target Signal Constrained Factorization Module;  \\
    Avg: Average; SE: Standard Error; Cell color scale: Within each comparison group (delineated by double horizontal lines), values are highlighted using 3-color scale (green = best, orange = worst, blue = midpoint; Microsoft Excel conditional formatting). \\
    \end{flushleft}
\end{sidewaystable*}

\subsubsection{Within-dataset Performance}\label{results within multi}
Three-fold validation is performed using the BP4D+ dataset \cite{zhang2016multimodal} for within-dataset evaluation, with subject-wise data splits as mentioned in \cite{narayanswamy2024bigsmall} and specified in the repository of rPPG Toolbox \cite{liu2024rppg} source code. In Table \ref{tab:results multi-task}, we report an average performance across three folds for BigSmall, \Model{} trained with FSAM as well as with \AM{} (see detailed fold-wise results in Supplementary Tables \ref{tab:results multi-task fold1}, \ref{tab:results multi-task fold2} and \ref{tab:results multi-task fold3}). 

\textbf{rRSP Estimation:} The performance of all models in the rRSP estimation task is poor when trained with RGB frames, which increases significantly when trained with thermal frames. It is important to note that in BigSmall \cite{narayanswamy2024bigsmall}, only the \textit{Small} branch conducts temporal modeling, and thus, providing thermal frames as input to the \textit{Big} branch does not confer any performance advantages. In contrast, \Model{} has a dedicated branch for the estimation of rRSP signals. 

A comparative analysis of performance between the \Model{} models, when supplied with RGB and thermal frames to its \textit{RSP} branch, indicates that thermal frames provide a rich respiratory signature. \Model{} trained with FSAM \cite{joshi2024factorizephys} and \AM{}, demonstrate comparable performance, both significantly exceeding the SOTA method \cite{narayanswamy2024bigsmall} in the rRSP estimation task. The model \Model{} with FSAM \cite{joshi2024factorizephys} exhibits marginally better performance in the RR error metrics, while \Model{} with \AM{} shows slightly better performance in the RR correlation, SNR and MACC. Smaller spatial resolutions (i.e., \(9\times9\)) consistently exhibit inferior performance compared to higher spatial resolutions (i.e., \(36\times36 \) and \(72\times72\)) of input video frames. 

Furthermore, no significant improvements are detected when using the \(72\times72\) resolution over the \(36\times36\) resolution, suggesting the latter to be adequate for the end-to-end estimation of rRSP signals from thermal video frames.  The low correlation in the estimation of respiratory rate (RR), despite promising MACC, can be attributed to the substantial number of video clips in the BP4D+ dataset \cite{zhang2016multimodal} that are shorter than 20 seconds. This duration insufficiency leads to inadequate frequency resolution within the respiratory waveform frequency range, compromising the accuracy of respiratory rate estimation when using the FFT peak-based method.

\textbf{rPPG Estimation:} Similar to the performance observed in rRSP, studies \Model{} using FSAM \cite{joshi2024factorizephys} and \AM{} demonstrate a significant improvement over the SOTA method \cite{narayanswamy2024bigsmall} in the task of rPPG estimation. In contrast to the results obtained with rRSP, models trained using RGB frames yield promising results. However, when training is conducted with thermal frames, the quality of the rPPG estimation deteriorates markedly, indicating a diminished presence of the rPPG signal within the thermal modality. 

\begin{figure*}[ht]
    \centering
    \includegraphics[width=1.0\textwidth]{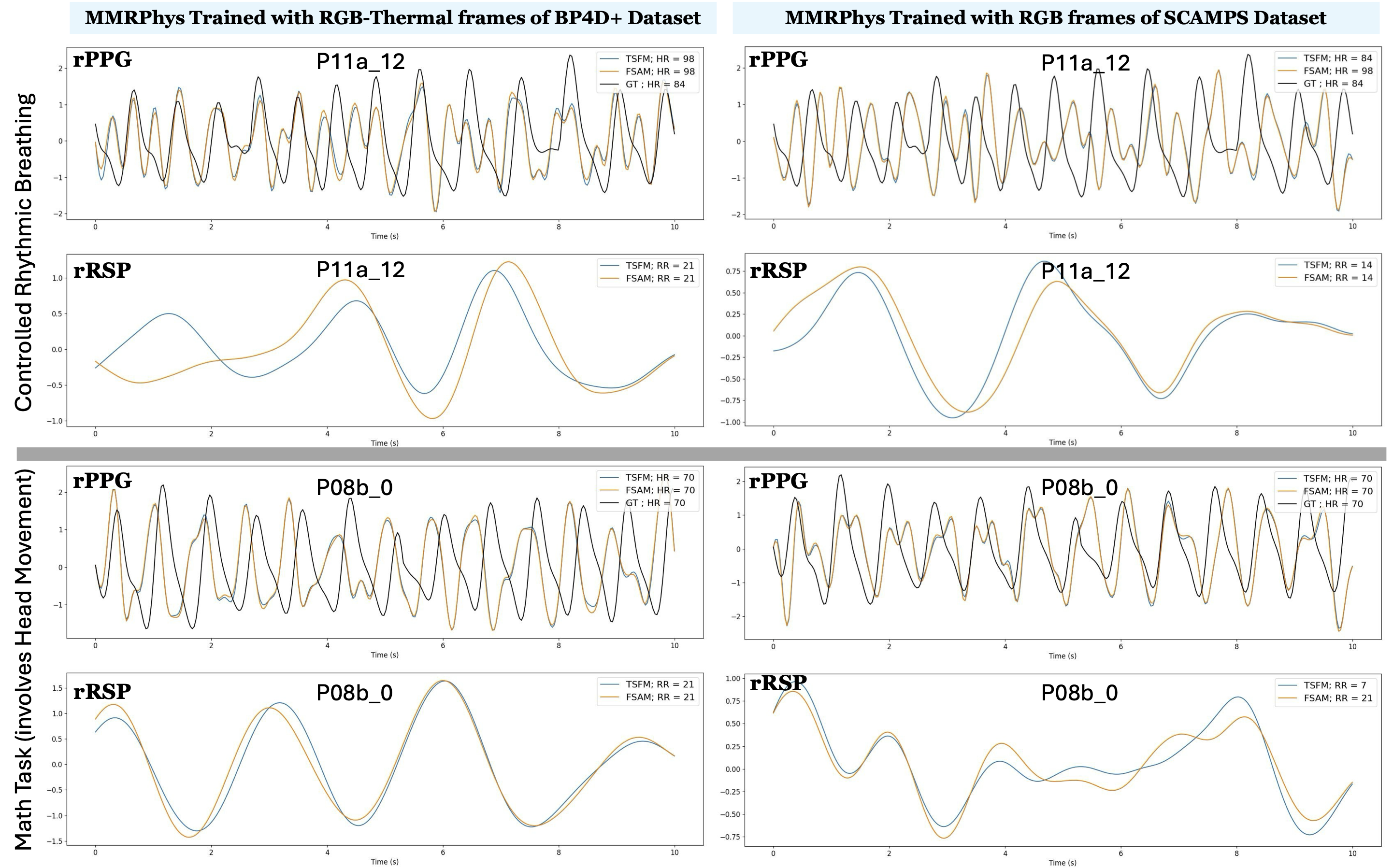}
    \caption{Comparison of rPPG and rRSP signals estimated on the iBVP Dataset \cite{joshi2024ibvp}, for \Model{} trained with \AM{}, and FSAM \cite{joshi2024factorizephys}. The left figures show results from models trained with RGB-Thermal frames from the BP4D+ dataset \cite{zhang2016multimodal, ertugrul2019cross}, and the right figures show outputs from models trained only with RGB frames from the SCAMPS dataset \cite{mcduff2022scamps}.}
    \label{fig:Sample Estimated Signals}
\end{figure*}

\subsubsection{Cross-dataset Generalization}\label{results cross multi}
The assessment of cross-dataset generalization in the context of multi-task remote physiological sensing holds significant relevance for real-world applications; however, it has seldom been investigated in the existing literature, primarily due to the scarcity of datasets possessing ground-truth labels for diverse physiological signals. We report cross-dataset performance for the models trained on SCAMPS dataset \cite{mcduff2022scamps} and tested on BP4D+ dataset \cite{zhang2016multimodal} in Table \ref{tab:results multi-task}. General trends indicate that all models demonstrate reduced performance on the rRSP and rPPG tasks, as evidenced by increased RR and HR errors alongside decreased SNR and MACC metrics, compared to their respective within-dataset performance. In addition, as SCAMPS dataset \cite{mcduff2022scamps} comprises only RGB frames, the performance on rRSP estimation is significantly low for all models. 

The proposed \Model{} consistently shows superior performance compared to the SOTA model \cite{narayanswamy2024bigsmall}, for both rRSP and rPPG estimations, while the results of \Model{} trained with FSAM \cite{joshi2024factorizephys} and with \AM{} remain comparable. For rRSP estimation, the \Model{} trained with FSAM \cite{joshi2024factorizephys} performs slightly better with slightly lower errors and higher correlation. For rPPG estimation \Model{} with \AM{} significantly outperforms FSAM \cite{joshi2024factorizephys} on all metrics. We also observe consistent performance gains when the proposed \Model{} models are trained with \(72\times72\) spatial resolution over \(9\times9\) resolution for the estimation of rRSP and rPPG.

We further made inferences on the multimodal iBVP dataset \cite{joshi2024ibvp}, which included RGB-Thermal video frames, to qualitatively compare the estimated signals, while performing a quantitative multitask assessment of the models trained with the BP4D+ dataset \cite{zhang2016multimodal} was not possible given the lack of additional multimodal datasets containing ground-truth labels for rRSP signals. This task was facilitated by the visualization of these signals, as depicted in Fig. \ref{fig:Sample Estimated Signals} for \Model{} trained with FSAM \cite{joshi2024factorizephys} and with \AM{}. Figure \ref{fig:Sample Estimated Signals} additionally presents estimated rPPG and rRSP signals for models trained exclusively on RGB frames of the SCAMPS dataset \cite{mcduff2022scamps}.

The upper group of rPPG and rRSP signals pertains to a controlled rhythmic breathing condition, marked by negligible head movement \cite{joshi2024ibvp}. In contrast, the lower group of rPPG and rRSP signals is associated with a mathematical task carried out by the participants, characterized by considerable head movement \cite{joshi2024ibvp} and unregulated breathing. Consistent with the cross-dataset findings reported for single-task estimation of rPPG in Table \ref{tab:results cross rPPG averaged}, the multi-task framework exhibits robust rPPG estimation performance across both experimental conditions for \Model{} models trained with both the training datasets. 

In contrast, rRSP estimation demonstrates consistent performance across the experimental conditions only when \Model{} models are trained on the thermal frames of the BP4D+ dataset \cite{zhang2016multimodal}. For \Model{} models trained on the SCAMPS dataset \cite{mcduff2022scamps}, rRSP estimation shows promising performance under controlled rhythmic breathing, which potentially involves rhythmic head movement. However, performance becomes inconsistent in the presence of task-related head movement, as indicated in the plots of the right upper group. From qualitative analysis, it is evident that the proposed \Model{} trained using \AM{} achieves a superior generalization in multitask signal estimation with multimodal input video frames.

\subsection{Learned Spatial-Temporal Features}\label{results feature visualization}
\begin{figure}[ht]
    \centering
    \includegraphics[width=0.95\linewidth]{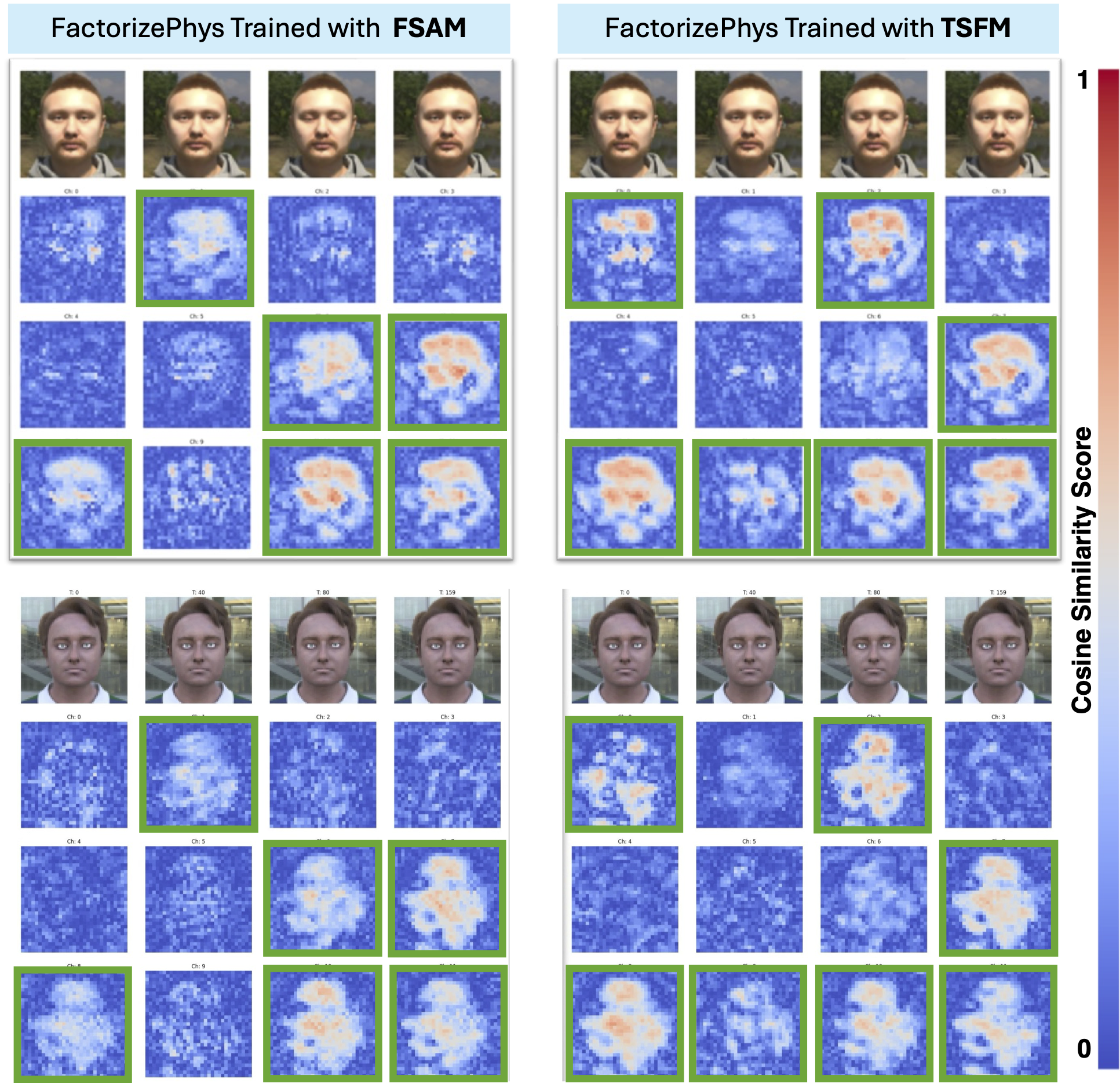}
    \caption{Analysis of spatial-temporal features learned by FactorizePhys \cite{joshi2024factorizephys} with FSAM \cite{joshi2024factorizephys} and the proposed \textbf{\AM{}}. The top row of subplots shows video frames from SCAMPS dataset \cite{mcduff2022scamps} at different time intervals.}
    \label{fig:feature visualization rPPG}
\end{figure}

To visualize the learned spatial-temporal features, we calculate the absolute cosine similarity (CSIM) between the temporal dimension of 4D embeddings, which includes the temporal, spatial, and channel dimensions, and the ground-truth rPPG signal \cite{joshi2024factorizephys}. As depicted in Fig. \ref{fig:feature visualization rPPG}, we performed a comparison of several channels within the fourth layer of the FactorizePhys \cite{joshi2024factorizephys} model, trained with FSAM \cite{joshi2024factorizephys} and our proposed approach \AM{} for rPPG estimation from RGB video frames. 

\begin{figure}[ht]
    \centering
    \includegraphics[width=0.95\linewidth]{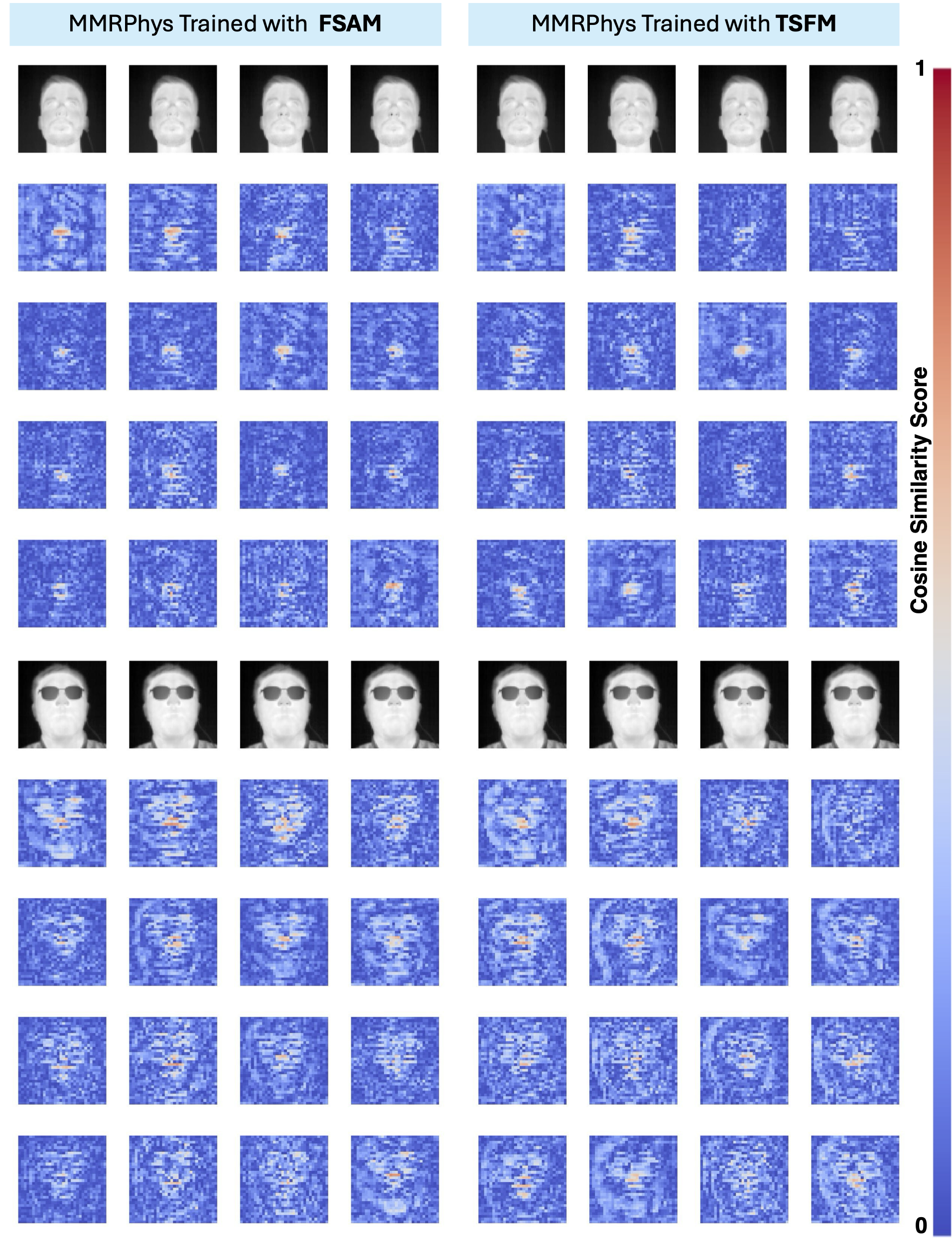}
    \caption{Analysis of spatial-temporal features learned by \Model{} using thermal video frames for rRSP estimation. On the left are features for \Model{} trained with FSAM \cite{joshi2024factorizephys}, and on the right, features trained with \textbf{\AM{}}. The top row in subplots shows video frames from iBVP dataset \cite{joshi2024ibvp}.}
    \label{fig:Learned Features Thermal RSP}
\end{figure}

In Fig. \ref{fig:Learned Features Thermal RSP}, we observe overlapping spatial-temporal features learned by \Model{}, when trained with FSAM \cite{joshi2024factorizephys} and \AM{}, to estimate the rRSP signal from thermal video frames.

The increased number of channels exhibiting higher CSIM scores for the FactorizePhys model \cite{joshi2024factorizephys} when trained with \AM{} substantiates the superior efficacy of our proposed method \AM{} in selecting pertinent spatial-temporal features, thus elucidating its superior performance. The spatial distribution of high CSIM scores for \Model{} trained to estimate the rRSP signal from thermal video frames (see Fig. \ref{fig:Learned Features Thermal RSP}), convincingly indicates nostril and mouth regions, known to have temperature fluctuations related to breathing, in the thermal domain \cite{cho2017robust}.

\subsection{Computational Complexity}\label{results computational complexity}
\begin{figure}[ht]
    \centering
    \includegraphics[width=1\linewidth]{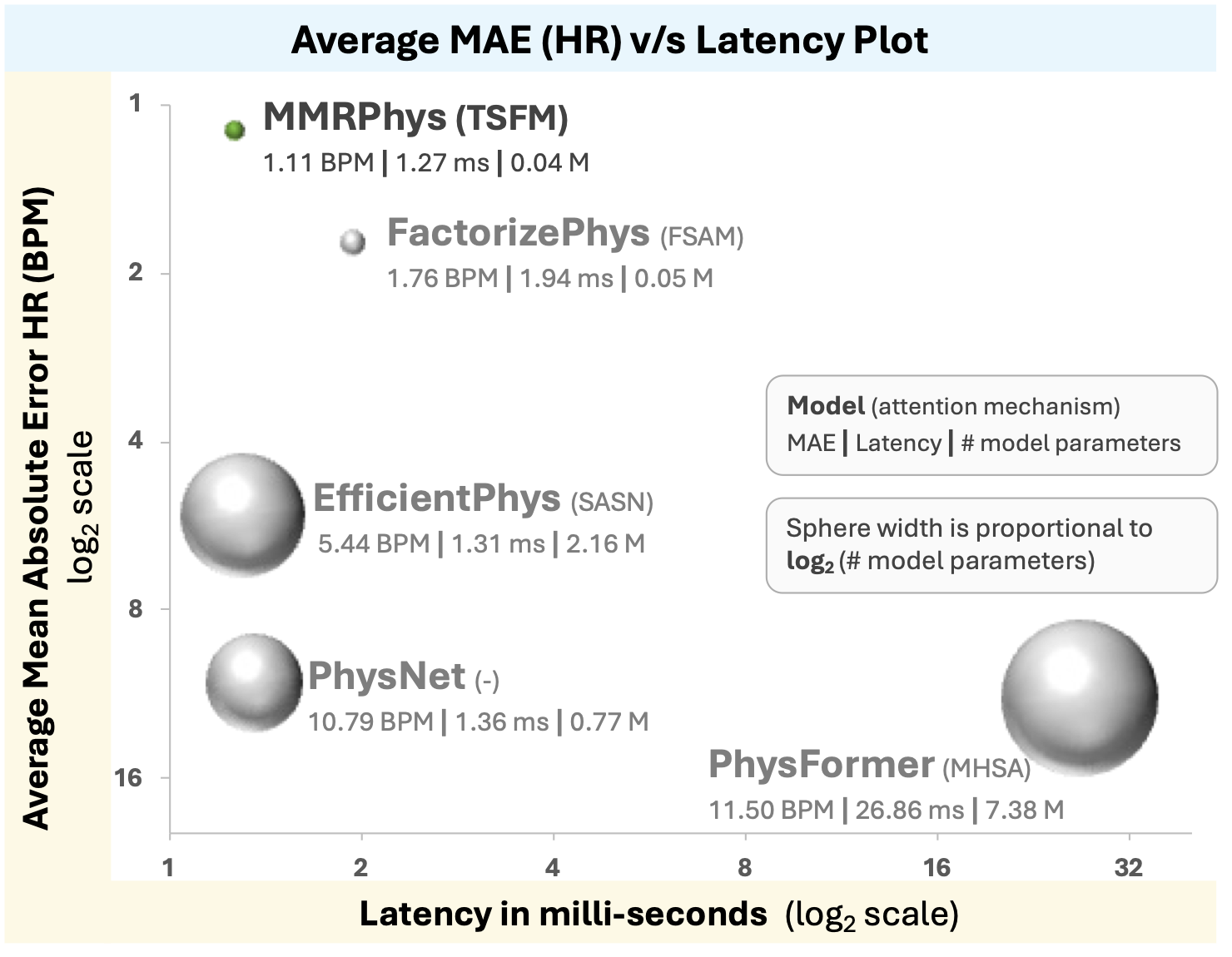}
    \caption{Plot of rPPG Estimation Accuracy (MAE) versus Latency illustrating a comprehensive comparison of cross-dataset performance between SOTA and the proposed methods.}
    \label{fig:summary rPPG performance}
\end{figure}

We assess the computational complexity of SOTA and proposed models by measuring inference latency on a laptop with an NVIDIA GeForce RTX 3070 GPU, 16GB RAM, and an Intel Core i7-10870H processor. We also document each method's trainable parameters. For a thorough analysis of cross-dataset generalization in relation to computational complexity (see Fig. \ref{fig:summary rPPG performance}), single-task rPPG estimation performance is chosen, as this is supported by various benchmarks and datasets.

On the Y-axis of Fig. \ref{fig:summary rPPG performance}, \(log_{2}\) (Mean Absolute Error) for heart rate (HR) metrics is determined by averaging the performance of each model trained on various datasets and evaluated on the corresponding testing datasets, as summarized in Table \ref{tab:results cross rPPG averaged} (see Supplementary Table \ref{tab:results cross all} for details). The size of the sphere corresponding to each model reflects the number of model parameters on the \(log_{2}\) scale. The unscaled values of MAE (in BPM), latency (in milliseconds), and model parameters are annotated alongside each model in Fig. \ref{fig:summary rPPG performance}. The proposed \Model{}, trained with \AM{}, exhibits superior performance with a significantly reduced number of model parameters and demonstrates the lowest latency.

\section{Conclusion}\label{sec conclusions}
This work makes theoretical and practical advances in machine intelligence for remote physiological sensing through two key contributions: target-signal constrained factorization (\AM{}) and the \Model{} architecture. \AM{} significantly enhances cross-dataset generalization by explicitly incorporating physiological signal characteristics as factorization constraints in jointly computing multidimensional attention. As demonstrated in our comprehensive evaluations across five diverse datasets, this constraint-based approach consistently outperforms state-of-the-art methods when faced with domain shifts in ambient conditions, camera specifications, head movements, facial poses, and physiological states. Substantial improvements in HR and RR error metrics (e.g. ~75\% reduction in HR MAE when trained on SCAMPS and tested on BP4D+) validate that \Model{} trained with \AM{} learns more robust and transferable features.

The \Model{} architecture further validates that carefully designed dual-branch 3D-CNN networks can effectively handle multimodal inputs (RGB and thermal) while simultaneously estimating multiple physiological signals (rPPG and rRSP). Our results establish that thermal imaging provides superior performance for respiratory monitoring while RGB excels for extracting rPPG signals, suggesting that optimal physiological monitoring systems can benefit from multimodal approaches. Furthermore, \Model{} demonstrates that parameter-efficient networks can achieve superior cross-dataset generalization while maintaining extremely low inference latency (1.27ms).

The consistent performance at lower spatial resolutions (36×36 compared to 72×72) indicates promising potential for resource-constrained deployments in mobile health applications. This work contributes to advancing robust physiological sensing across domain shifts and offers promising approaches for real-world contactless vital sign monitoring systems in varied environments. Future research should explore extending \AM{} to other physiological signals beyond rPPG and rRSP, and investigate its application in different computer vision domains that require extracting weak signals from high-dimensional data with significant distribution shifts.




\section*{Declarations}
\begin{itemize}
\item Funding: The author JJ was fully supported by the UCL CS PhD Studentship (GDI - Physiological Computing and Artificial Intelligence) which Prof. Cho secured.
\item Ethics approval and consent to participate: Although this work did not require working with human subjects, we have institutional ethics covered for this research, which is approved by the Ethics Committee of the University College London Interaction Center (ID Number: UCLIC/1920/006).
\item Code availability: The source code is available on this GitHub repository \url{https://github.com/PhysiologicAILab/MMRPhys}.
\item Author contribution: Conceptualization, J.J.; methodology, J.J. and Y.C.; software, J.J.; validation, J.J., and Y.C.; formal analysis, J.J.; investigation, J.J. and Y.C.; resources, Y.C.; data curation, J.J.; writing---original draft preparation, J.J.; writing---review and revision, J.J. and Y.C.; visualization, J.J.; overall supervision, Y.C.; project administration, Y.C.; funding acquisition,  Y.C..
\end{itemize}

\appendix
\section{Supplemental Material}
\label{appendix_section}
\subsection{Methods Primer}\label{methods primer}
\subsubsection{Factorized Self-Attention Module}\label{method primer fsam}
The spatial-temporal input video frames, \(\mathcal{I} \in \mathbb{R}^{T \times C \times H \times W}\), in which \(T, C, H, and~W\) represents the total number of frames, channels within a frame (for example, \(C=3\) for RGB and \(C=1\) for thermal imaging), as well as the height and width of pixels in a frame is passed to the feature extractor that produces voxel embeddings \(\varepsilon \in \mathbb{R}^{\tau \times \kappa \times \alpha \times \beta}\), characterized by temporal (\(\tau\)), channel (\(\kappa\)), and spatial (\(\alpha,\beta\)) dimensions.

\(\varepsilon\) is preprocessed through a convolution layer (with \(1 \times 1 \times 1\) kernels), and a ReLU activation to ensure non-negativity of the embeddings. As factorization is performed on a 2-D matrix, the multidimensional \(\varepsilon \in \mathbb{R}^{\tau \times \kappa \times \alpha \times \beta}\) is transformed into \(V \in \mathbb{R}^{M \times N}\) as follows. The temporal features of \(\varepsilon\) are mapped to the vector dimension (\(M\)) in \(V\), while the spatial and channel dimensions are mapped to the feature dimension (\(N\)) of \(V^{st}\). This transformation of \(\varepsilon\) is expressed as:
\begin{equation}
\label{eq 3d mapping}
    \begin{split}
    V \in \mathbb{R}^{M \times N} = \Gamma^{\tau\kappa\alpha\beta \mapsto MN}(\xi_{pre}(\varepsilon \in \mathbb{R}^{\tau \times \kappa \times \alpha \times \beta})) \\
    \ni \tau \mapsto M, \kappa \times \alpha \times \beta \mapsto N
    \end{split}
\end{equation}
where, \(\xi_{pre}\) represents preprocessing operation. The low-rank matrix \(\hat{V}\), which is approximated using NMF, is mapped back into the embedding space, yielding a low-rank voxel embeddings that selectively preserve the spatial and channel features pertinent to the restoration of significant temporal features in \(\varepsilon\). The post-processing steps comprise a convolution layer (with \(1 \times 1 \times 1\) kernels), and a ReLU activation. The resultant \(\hat{\varepsilon}\) is expressed as:
\begin{equation}
\label{eq appx embeddings}
    \hat{\varepsilon} = \xi_{post}(\Gamma^{MN \mapsto \tau\kappa\alpha\beta}(\hat{V} \in \mathbb{R}^{M \times N}))
\end{equation}
where \(\Gamma^{MN \mapsto \tau\kappa\alpha\beta}\) represents mapping operation. The element-wise multiplication of factorized embeddings \(\hat{\varepsilon}\) and \(\varepsilon\) is then performed, which can be interpreted as an excitation step. The product is further instance normalized and together with the residual connection, it is passed to the network head \(\omega\). This is expressed as:
\begin{equation}
\label{eq rppg estimation}
    r^{phys} = \omega(\varepsilon + \mathcal{IN}(\varepsilon \odot \hat{\varepsilon}))
\end{equation}
where \(r^{phys} \in \mathbb{R}^{1 \times T}\) is an estimated physiological signal. 

\subsubsection{Smoothness Constrained NMF}\label{method primer smooth nmf}
As originally proposed in \cite{zdunek2012approximation}, the low-rank matrix \(\hat{A}\) of a given matrix \(A = [a_{1}, a_{2}, ..., a_{N}] \in \mathbb{R}^{M \times N}_{\geq 0}\) , can be expressed with the basic form of the Smooth NMF model as:
\begin{equation}\label{eq snmf}
    \hat{A} = \Phi P Q \ni \Phi P > 0~and~Q>0 
\end{equation}
where \(\hat{A} = [\hat{a_{1}}, \hat{a_{2}}, ..., \hat{a_{N}}] \in \mathbb{R}^{M \times N}_{\geq 0}\) is reconstructed low-rank matrix, \(P = [p1, p2, ..., pL] \in \mathbb{R}^{M \times L}_{\geq 0}\) is nonnegative basis matrix, and \(Q = [q1, q2, ..., qN] \in \mathbb{R}^{L \times N}_{\geq 0}\) is a nonnegative coefficient matrix. \(\Phi\) represents a matrix formed with a set of smooth vectors \([\phi_{1}, \phi_{2}, ..., \phi_{K}] \subset \mathbb{R}^{M}\) that serve as constraints to solve the factorization. \(M\) is the total length of the vector or signal, \(L\) is the rank of factorization, and \(\mathbb{R}^{M \times N}_{\geq 0}\) stands for the set of \(M \times N\) element-wise nonnegative matrices. 

\begin{figure}[ht]
    \centering
    \includegraphics[width=1.0\linewidth]{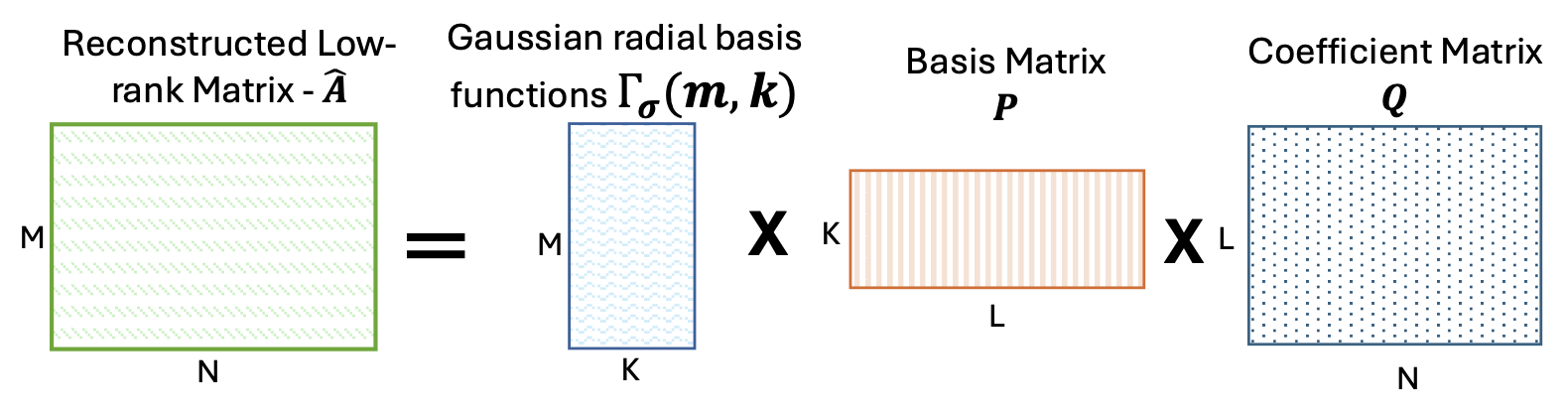}
    \caption{Smooth NMF, with Gaussian Radial Basis Functions as smoothness constraints.
    }
    \label{fig:snmf}
\end{figure}

For optimization, the two parameter matrices \(P\) and \(Q\) are estimated, since \(\Phi\) is known. The smoothness constraint matrix \(\Phi\) comprise Gaussian Radial Basis Functions (GRBF) with standard deviation \(\sigma\) \cite{yokota2015smooth}, as expressed by equation \ref{eq SNMF Phi} and illustrated in Fig. \ref{fig:snmf}.
\begin{equation}\label{eq SNMF Phi}
    \Phi(m, k) \leftarrow \Gamma_{\sigma}(m, k):=\exp{\left[- \frac{(m-k\Delta t)^{2}}{2\sigma^2} \right]}
\end{equation}
where, \(\Delta t:=\) time interval satisfying \(K = \lfloor (M - 1)/ \Delta t \rfloor\ + 1)\), \(K:=\) total number of basis functions.

The flexibility of the representation decreases with large \(\sigma\), making it robust for noisy data. On the other hand, when \(\sigma\) is small, \(\phi_{n}\) forms orthogonal bases, which increases the flexibility of representation with reduced smoothness constraints, although reducing robustness for noisy data \cite{yokota2015smooth}. Hyperparameters of \(\Phi\) including \(\sigma\) and \(\Delta t\) can potentially be adjusted when smooth NMF is applied directly to signals with known temporal characteristics. 

\subsection{Detailed Single-Task rPPG Evaluation}\label{secA1 cross dataset rPPG}
Please see \cref{tab:results cross all}.

\begin{table*}[ht]
\caption{Cross-dataset Performance Evaluation for rPPG Estimation}
\label{tab:results cross all}
\centering
\fontsize{5}{3}\selectfont
\setlength{\tabcolsep}{1pt}
\renewcommand{\arraystretch}{1.8}
\begin{tabular*}{\textwidth}{@{\extracolsep\fill}cllcccccccccccl}
\hline
 &
  \multicolumn{1}{c}{} &
  \multicolumn{1}{c}{} &
  \multicolumn{2}{c}{\textbf{MAE ↓}} &
  \multicolumn{2}{c}{\textbf{RMSE ↓}} &
  \multicolumn{2}{c}{\textbf{MAPE↓}} &
  \multicolumn{2}{c}{\textbf{Corr ↑}} &
  \multicolumn{2}{c}{\textbf{SNR ↑}} &
  \multicolumn{2}{c}{\textbf{MACC ↑}} \\ \cline{4-15} 
\multirow{-2}{*}{\textbf{\begin{tabular}[c]{@{}c@{}}Training\\ Dataset\end{tabular}}} &
  \multicolumn{1}{c}{\multirow{-2}{*}{\textbf{Model}}} &
  \multicolumn{1}{c}{\multirow{-2}{*}{\textbf{\begin{tabular}[c]{@{}c@{}}Attention\\ Module\end{tabular}}}} &
  \textbf{Avg} &
  {\color[HTML]{808080} \textbf{SE}} &
  \textbf{Avg} &
  {\color[HTML]{808080} \textbf{SE}} &
  \textbf{Avg} &
  {\color[HTML]{808080} \textbf{SE}} &
  \textbf{Avg} &
  {\color[HTML]{808080} \textbf{SE}} &
  \textbf{Avg} &
  {\color[HTML]{808080} \textbf{SE}} &
  \textbf{Avg} &
  \multicolumn{1}{c}{{\color[HTML]{808080} \textbf{SE}}} \\ \hline \hline
\multicolumn{15}{l}{\textbf{Performance Evaluation on iBVP Dataset \cite{joshi2024ibvp}}} \\ \hline \hline
 &
  PhysNet &
  - &
  \cellcolor[HTML]{C1F0C8}1.63 &
  {\color[HTML]{808080} 0.33} &
  \cellcolor[HTML]{C1F0CB}3.77 &
  {\color[HTML]{808080} 0.73} &
  \cellcolor[HTML]{C1F0C8}2.17 &
  {\color[HTML]{808080} 0.42} &
  \cellcolor[HTML]{C2EFCC}0.92 &
  {\color[HTML]{808080} 0.04} &
  \cellcolor[HTML]{C5EEDA}6.08 &
  {\color[HTML]{808080} 0.62} &
  \cellcolor[HTML]{C5EEDE}0.55 &
  {\color[HTML]{808080} 0.01} \\ \cline{2-15} 
 &
  PhysFormer &
  MHSA* &
  \cellcolor[HTML]{C9EEF8}2.50 &
  {\color[HTML]{808080} 0.64} &
  \cellcolor[HTML]{CAEDFB}7.09 &
  {\color[HTML]{808080} 1.50} &
  \cellcolor[HTML]{C9EEF9}3.39 &
  {\color[HTML]{808080} 0.82} &
  \cellcolor[HTML]{CAEDF9}0.79 &
  {\color[HTML]{808080} 0.06} &
  \cellcolor[HTML]{C9EDF0}5.21 &
  {\color[HTML]{808080} 0.60} &
  \cellcolor[HTML]{C9EDF3}0.52 &
  {\color[HTML]{808080} 0.01} \\ \cline{2-15} 
 &
  EfficientPhys &
  SASN \dag &
  \cellcolor[HTML]{CBEDFA}3.80 &
  {\color[HTML]{808080} 1.38} &
  \cellcolor[HTML]{D4EBF4}14.82 &
  {\color[HTML]{808080} 3.74} &
  \cellcolor[HTML]{CBEDFA}5.15 &
  {\color[HTML]{808080} 1.87} &
  \cellcolor[HTML]{D7EAF1}0.56 &
  {\color[HTML]{808080} 0.08} &
  \cellcolor[HTML]{D1EBF5}2.93 &
  {\color[HTML]{808080} 0.48} &
  \cellcolor[HTML]{D2EBF4}0.45 &
  {\color[HTML]{808080} 0.01} \\ \cline{2-15} 
 &
  FactorizePhys &
  FSAM \ddag &
  \cellcolor[HTML]{C1F0C9}\textbf{1.66} &
  {\color[HTML]{808080} 0.30} &
  \cellcolor[HTML]{C1F0C8}\textbf{3.55} &
  {\color[HTML]{808080} 0.65} &
  \cellcolor[HTML]{C1F0CD}\textbf{2.31} &
  {\color[HTML]{808080} 0.46} &
  \cellcolor[HTML]{C1F0C8}\textbf{0.93} &
  {\color[HTML]{808080} 0.04} &
  \cellcolor[HTML]{C1F0C8}\textbf{6.78} &
  {\color[HTML]{808080} 0.57} &
  \cellcolor[HTML]{C1F0C8}\textbf{0.58} &
  {\color[HTML]{808080} 0.01} \\ \cline{2-15} 
 &
  FactorizePhys &
  \textbf{TSFM} &
  \cellcolor[HTML]{C4EFDA}1.96 &
  {\color[HTML]{808080} 0.39} &
  \cellcolor[HTML]{C3F0D5}4.52 &
  {\color[HTML]{808080} 0.95} &
  \cellcolor[HTML]{C4EFD9}2.60 &
  {\color[HTML]{808080} 0.51} &
  \cellcolor[HTML]{C5EEDA}0.88 &
  {\color[HTML]{808080} 0.05} &
  \cellcolor[HTML]{C2EFCD}6.60 &
  {\color[HTML]{808080} 0.57} &
  \cellcolor[HTML]{C3EFD0}0.57 &
  {\color[HTML]{808080} 0.01} \\ \cline{2-15} 
\multirow{-8}{*}{PURE} &
  \textbf{MMRPhys} &
  \textbf{TSFM} &
  \cellcolor[HTML]{C3F0D3}1.84 &
  {\color[HTML]{808080} 0.31} &
  \cellcolor[HTML]{C1F0CA}3.71 &
  {\color[HTML]{808080} 0.60} &
  \cellcolor[HTML]{C3F0D4}2.47 &
  {\color[HTML]{808080} 0.43} &
  \cellcolor[HTML]{C2EFCC}0.92 &
  {\color[HTML]{808080} 0.04} &
  \cellcolor[HTML]{C4EFD7}6.21 &
  {\color[HTML]{808080} 0.57} &
  \cellcolor[HTML]{C3EFD2}0.57 &
  {\color[HTML]{808080} 0.01} \\ \hline
 &
  PhysNet &
  - &
  \cellcolor[HTML]{EEE5DF}31.85 &
  {\color[HTML]{808080} 1.89} &
  \cellcolor[HTML]{F1E5DD}37.40 &
  {\color[HTML]{808080} 3.38} &
  \cellcolor[HTML]{EFE5DE}45.62 &
  {\color[HTML]{808080} 2.96} &
  \cellcolor[HTML]{FAE2D5}-0.10 &
  {\color[HTML]{808080} 0.10} &
  \cellcolor[HTML]{F1E4DD}-6.11 &
  {\color[HTML]{808080} 0.22} &
  \cellcolor[HTML]{F9E2D7}0.16 &
  {\color[HTML]{808080} 0.00} \\ \cline{2-15} 
 &
  PhysFormer &
  MHSA* &
  \cellcolor[HTML]{FBE2D5}41.73 &
  {\color[HTML]{808080} 1.31} &
  \cellcolor[HTML]{FAE3D6}43.89 &
  {\color[HTML]{808080} 3.11} &
  \cellcolor[HTML]{FBE2D5}58.56 &
  {\color[HTML]{808080} 2.36} &
  \cellcolor[HTML]{EDE5E0}0.15 &
  {\color[HTML]{808080} 0.10} &
  \cellcolor[HTML]{FBE2D5}-9.13 &
  {\color[HTML]{808080} 0.53} &
  \cellcolor[HTML]{FBE2D5}0.14 &
  {\color[HTML]{808080} 0.00} \\ \cline{2-15} 
 &
  EfficientPhys &
  SASN \dag &
  \cellcolor[HTML]{E7E7E5}26.19 &
  {\color[HTML]{808080} 3.47} &
  \cellcolor[HTML]{FBE2D5}44.55 &
  {\color[HTML]{808080} 6.18} &
  \cellcolor[HTML]{E8E7E4}38.11 &
  {\color[HTML]{808080} 5.21} &
  \cellcolor[HTML]{FBE2D5}-0.12 &
  {\color[HTML]{808080} 0.10} &
  \cellcolor[HTML]{E4E7E7}-2.36 &
  {\color[HTML]{808080} 0.38} &
  \cellcolor[HTML]{E6E6E5}0.30 &
  {\color[HTML]{808080} 0.01} \\ \cline{2-15} 
 &
  FactorizePhys &
  FSAM \ddag &
  \cellcolor[HTML]{CAEDFB}2.71 &
  {\color[HTML]{808080} 0.54} &
  \cellcolor[HTML]{C7EEEE}6.22 &
  {\color[HTML]{808080} 1.38} &
  \cellcolor[HTML]{CAEDFB}3.87 &
  {\color[HTML]{808080} 0.80} &
  \cellcolor[HTML]{C9EDF2}0.81 &
  {\color[HTML]{808080} 0.06} &
  \cellcolor[HTML]{D3EBF4}2.36 &
  {\color[HTML]{808080} 0.47} &
  \cellcolor[HTML]{D5EAF2}0.43 &
  {\color[HTML]{808080} 0.01} \\ \cline{2-15} 
 &
  FactorizePhys &
  \textbf{TSFM} &
  \cellcolor[HTML]{C8EEF1}\textbf{2.37} &
  {\color[HTML]{808080} 0.67} &
  \cellcolor[HTML]{CAEDFB}7.36 &
  {\color[HTML]{808080} 2.00} &
  \cellcolor[HTML]{C8EEF2}\textbf{3.22} &
  {\color[HTML]{808080} 0.90} &
  \cellcolor[HTML]{CBECFA}\textbf{0.78} &
  {\color[HTML]{808080} 0.06} &
  \cellcolor[HTML]{CCECF9}\textbf{4.33} &
  {\color[HTML]{808080} 0.54} &
  \cellcolor[HTML]{CCECF9}\textbf{0.50} &
  {\color[HTML]{808080} 0.01} \\ \cline{2-15} 
\multirow{-8}{*}{SCAMPS} &
  \textbf{MMRPhys} &
  \textbf{TSFM} &
  \cellcolor[HTML]{CAEDFB}2.58 &
  {\color[HTML]{808080} 0.73} &
  \cellcolor[HTML]{CBEDFB}7.98 &
  {\color[HTML]{808080} 2.09} &
  \cellcolor[HTML]{CAEDFB}3.48 &
  {\color[HTML]{808080} 1.00} &
  \cellcolor[HTML]{CCECF9}0.75 &
  {\color[HTML]{808080} 0.06} &
  \cellcolor[HTML]{CFEBF7}3.50 &
  {\color[HTML]{808080} 0.53} &
  \cellcolor[HTML]{CFECF7}0.48 &
  {\color[HTML]{808080} 0.01} \\ \hline
 &
  PhysNet &
  - &
  \cellcolor[HTML]{CAEDFB}3.18 &
  {\color[HTML]{808080} 0.67} &
  \cellcolor[HTML]{CAEDFB}7.65 &
  {\color[HTML]{808080} 1.46} &
  \cellcolor[HTML]{CBEDFB}4.84 &
  {\color[HTML]{808080} 1.14} &
  \cellcolor[HTML]{CFEBF7}0.70 &
  {\color[HTML]{808080} 0.07} &
  \cellcolor[HTML]{C7EEE8}5.54 &
  {\color[HTML]{808080} 0.61} &
  \cellcolor[HTML]{C4EFD7}0.56 &
  {\color[HTML]{808080} 0.01} \\ \cline{2-15} 
 &
  PhysFormer &
  MHSA* &
  \cellcolor[HTML]{D0ECF6}7.86 &
  {\color[HTML]{808080} 1.46} &
  \cellcolor[HTML]{D7EBF1}17.13 &
  {\color[HTML]{808080} 2.69} &
  \cellcolor[HTML]{D1ECF6}11.44 &
  {\color[HTML]{808080} 2.25} &
  \cellcolor[HTML]{E0E8EA}0.38 &
  {\color[HTML]{808080} 0.09} &
  \cellcolor[HTML]{D5EAF2}1.71 &
  {\color[HTML]{808080} 0.56} &
  \cellcolor[HTML]{D5EAF2}0.43 &
  {\color[HTML]{808080} 0.01} \\ \cline{2-15} 
 &
  EfficientPhys &
  SASN \dag &
  \cellcolor[HTML]{CAEDFB}2.74 &
  {\color[HTML]{808080} 0.63} &
  \cellcolor[HTML]{C9EEFA}7.07 &
  {\color[HTML]{808080} 1.81} &
  \cellcolor[HTML]{CAEDFB}4.02 &
  {\color[HTML]{808080} 1.08} &
  \cellcolor[HTML]{CDECF9}0.74 &
  {\color[HTML]{808080} 0.07} &
  \cellcolor[HTML]{CDECF8}4.03 &
  {\color[HTML]{808080} 0.55} &
  \cellcolor[HTML]{CDECF9}0.49 &
  {\color[HTML]{808080} 0.01} \\ \cline{2-15} 
 &
  FactorizePhys &
  FSAM \ddag &
  \cellcolor[HTML]{C2F0CE}\textbf{1.74} &
  {\color[HTML]{808080} 0.39} &
  \cellcolor[HTML]{C3F0D4}4.39 &
  {\color[HTML]{808080} 1.06} &
  \cellcolor[HTML]{C2F0D2}\textbf{2.42} &
  {\color[HTML]{808080} 0.57} &
  \cellcolor[HTML]{C3EFD3}0.90 &
  {\color[HTML]{808080} 0.04} &
  \cellcolor[HTML]{C2EFCD}\textbf{6.59} &
  {\color[HTML]{808080} 0.57} &
  \cellcolor[HTML]{C4EFD7}\textbf{0.56} &
  {\color[HTML]{808080} 0.01} \\ \cline{2-15} 
 &
  FactorizePhys &
  \textbf{TSFM} &
  \cellcolor[HTML]{C2F0D2}1.82 &
  {\color[HTML]{808080} 0.34} &
  \cellcolor[HTML]{C2F0CE}3.99 &
  {\color[HTML]{808080} 0.76} &
  \cellcolor[HTML]{C3F0D4}2.49 &
  {\color[HTML]{808080} 0.48} &
  \cellcolor[HTML]{C2EFCD}\textbf{0.92} &
  {\color[HTML]{808080} 0.04} &
  \cellcolor[HTML]{C4EFD6}6.26 &
  {\color[HTML]{808080} 0.58} &
  \cellcolor[HTML]{C5EEDE}0.55 &
  {\color[HTML]{808080} 0.01} \\ \cline{2-15} 
\multirow{-8}{*}{UBFC-rPPG} &
  \textbf{MMRPhys} &
  \textbf{TSFM} &
  \cellcolor[HTML]{C3F0D7}1.91 &
  {\color[HTML]{808080} 0.36} &
  \cellcolor[HTML]{C2F0D1}\textbf{4.18} &
  {\color[HTML]{808080} 0.77} &
  \cellcolor[HTML]{C4EFD9}2.60 &
  {\color[HTML]{808080} 0.50} &
  \cellcolor[HTML]{C3EFD1}0.91 &
  {\color[HTML]{808080} 0.04} &
  \cellcolor[HTML]{C6EEE2}5.76 &
  {\color[HTML]{808080} 0.58} &
  \cellcolor[HTML]{C7EEE5}0.54 &
  {\color[HTML]{808080} 0.01} \\ \hline \hline
\multicolumn{15}{l}{\textbf{Performance Evaluation on PURE Dataset \cite{stricker2014non}}} \\ \hline \hline
 &
  PhysNet &
  - &
  \cellcolor[HTML]{D3EBF4}7.78 &
  {\color[HTML]{808080} 2.27} &
  \cellcolor[HTML]{D9EAEF}19.12 &
  {\color[HTML]{808080} 3.93} &
  \cellcolor[HTML]{CFECF8}8.94 &
  {\color[HTML]{808080} 2.71} &
  \cellcolor[HTML]{EBE5E1}0.59 &
  {\color[HTML]{808080} 0.11} &
  \cellcolor[HTML]{D1EBF6}9.90 &
  {\color[HTML]{808080} 1.49} &
  \cellcolor[HTML]{CDECF9}0.70 &
  {\color[HTML]{808080} 0.03} \\ \cline{2-15} 
 &
  PhysFormer &
  MHSA* &
  \cellcolor[HTML]{D1ECF6}6.58 &
  {\color[HTML]{808080} 1.98} &
  \cellcolor[HTML]{D4EBF3}16.55 &
  {\color[HTML]{808080} 3.60} &
  \cellcolor[HTML]{CCEDF9}6.93 &
  {\color[HTML]{808080} 1.90} &
  \cellcolor[HTML]{D8E9F0}0.76 &
  {\color[HTML]{808080} 0.09} &
  \cellcolor[HTML]{D1EBF5}9.75 &
  {\color[HTML]{808080} 1.96} &
  \cellcolor[HTML]{CCECFA}0.71 &
  {\color[HTML]{808080} 0.03} \\ \cline{2-15} 
 &
  EfficientPhys &
  SASN \dag &
  \cellcolor[HTML]{C1F0CC}0.56 &
  {\color[HTML]{808080} 0.17} &
  \cellcolor[HTML]{C1F0CA}1.40 &
  {\color[HTML]{808080} 0.33} &
  \cellcolor[HTML]{C1F0CD}0.87 &
  {\color[HTML]{808080} 0.28} &
  \cellcolor[HTML]{C2EFC9}\textbf{0.999} &
  {\color[HTML]{808080} 0.01} &
  \cellcolor[HTML]{CAEDF8}11.96 &
  {\color[HTML]{808080} 0.84} &
  \cellcolor[HTML]{C9EDF3}0.73 &
  {\color[HTML]{808080} 0.02} \\ \cline{2-15} 
 &
  FactorizePhys &
  FSAM \ddag &
  \cellcolor[HTML]{C2F0CD}0.60 &
  {\color[HTML]{808080} 0.21} &
  \cellcolor[HTML]{C1F0CB}1.70 &
  {\color[HTML]{808080} 0.42} &
  \cellcolor[HTML]{C1F0CD}0.87 &
  {\color[HTML]{808080} 0.30} &
  \cellcolor[HTML]{C2EFC9}0.997 &
  {\color[HTML]{808080} 0.01} &
  \cellcolor[HTML]{C3EFD4}15.19 &
  {\color[HTML]{808080} 0.91} &
  \cellcolor[HTML]{C4EFD5}0.77 &
  {\color[HTML]{808080} 0.02} \\ \cline{2-15} 
 &
  FactorizePhys &
  \textbf{TSFM} &
  \cellcolor[HTML]{C1F0C9}\textbf{0.40} &
  {\color[HTML]{808080} 0.15} &
  \cellcolor[HTML]{C1F0C9}\textbf{1.21} &
  {\color[HTML]{808080} 0.32} &
  \cellcolor[HTML]{C1F0CA}\textbf{0.65} &
  {\color[HTML]{808080} 0.26} &
  \cellcolor[HTML]{C2EFC9}\textbf{0.999} &
  {\color[HTML]{808080} 0.01} &
  \cellcolor[HTML]{C1F0C8}\textbf{16.19} &
  {\color[HTML]{808080} 0.96} &
  \cellcolor[HTML]{C4EFD5}0.77 &
  {\color[HTML]{808080} 0.02} \\ \cline{2-15} 
\multirow{-8}{*}{iBVP} &
  \textbf{MMRPhys} &
  \textbf{TSFM} &
  \cellcolor[HTML]{C1F0CA}0.42 &
  {\color[HTML]{808080} 0.16} &
  \cellcolor[HTML]{C1F0C9}1.28 &
  {\color[HTML]{808080} 0.33} &
  \cellcolor[HTML]{C1F0CB}0.66 &
  {\color[HTML]{808080} 0.26} &
  \cellcolor[HTML]{C2EFC9}\textbf{0.999} &
  {\color[HTML]{808080} 0.01} &
  \cellcolor[HTML]{C3EFD0}15.56 &
  {\color[HTML]{808080} 0.96} &
  \cellcolor[HTML]{C1F0C8}\textbf{0.79} &
  {\color[HTML]{808080} 0.02} \\ \hline
 &
  PhysNet &
  - &
  \cellcolor[HTML]{FBE2D5}26.74 &
  {\color[HTML]{808080} 3.17} &
  \cellcolor[HTML]{FBE2D5}36.19 &
  {\color[HTML]{808080} 5.18} &
  \cellcolor[HTML]{FBE2D5}46.73 &
  {\color[HTML]{808080} 5.66} &
  \cellcolor[HTML]{FBE2D5}0.45 &
  {\color[HTML]{808080} 0.12} &
  \cellcolor[HTML]{FBE2D5}-2.21 &
  {\color[HTML]{808080} 0.66} &
  \cellcolor[HTML]{FBE2D5}0.31 &
  {\color[HTML]{808080} 0.02} \\ \cline{2-15} 
 &
  PhysFormer &
  MHSA* &
  \cellcolor[HTML]{E6E7E6}16.64 &
  {\color[HTML]{808080} 2.95} &
  \cellcolor[HTML]{EBE6E2}28.13 &
  {\color[HTML]{808080} 5.00} &
  \cellcolor[HTML]{E8E7E4}30.58 &
  {\color[HTML]{808080} 5.72} &
  \cellcolor[HTML]{F5E3DA}0.51 &
  {\color[HTML]{808080} 0.11} &
  \cellcolor[HTML]{F1E4DD}0.84 &
  {\color[HTML]{808080} 1.00} &
  \cellcolor[HTML]{EEE4DF}0.42 &
  {\color[HTML]{808080} 0.02} \\ \cline{2-15} 
 &
  EfficientPhys &
  SASN \dag &
  \cellcolor[HTML]{D0ECF6}6.21 &
  {\color[HTML]{808080} 2.26} &
  \cellcolor[HTML]{D8EAF0}18.45 &
  {\color[HTML]{808080} 4.54} &
  \cellcolor[HTML]{D2ECF5}12.16 &
  {\color[HTML]{808080} 4.57} &
  \cellcolor[HTML]{DAE9EE}0.74 &
  {\color[HTML]{808080} 0.09} &
  \cellcolor[HTML]{E4E7E7}4.39 &
  {\color[HTML]{808080} 0.78} &
  \cellcolor[HTML]{E4E7E7}0.51 &
  {\color[HTML]{808080} 0.02} \\ \cline{2-15} 
 &
  FactorizePhys &
  FSAM \ddag &
  \cellcolor[HTML]{CFECF8}5.43 &
  {\color[HTML]{808080} 1.93} &
  \cellcolor[HTML]{D3EBF4}15.80 &
  {\color[HTML]{808080} 3.58} &
  \cellcolor[HTML]{D1ECF6}11.10 &
  {\color[HTML]{808080} 4.05} &
  \cellcolor[HTML]{D3EBF4}0.80 &
  {\color[HTML]{808080} 0.08} &
  \cellcolor[HTML]{CBECFA}11.40 &
  {\color[HTML]{808080} 0.76} &
  \cellcolor[HTML]{D0EBF6}0.67 &
  {\color[HTML]{808080} 0.02} \\ \cline{2-15} 
 &
  FactorizePhys &
  \textbf{TSFM} &
  \cellcolor[HTML]{C4EFDA}1.30 &
  {\color[HTML]{808080} 0.93} &
  \cellcolor[HTML]{C6EFE8}7.29 &
  {\color[HTML]{808080} 2.59} &
  \cellcolor[HTML]{C4EFDE}2.25 &
  {\color[HTML]{808080} 1.71} &
  \cellcolor[HTML]{C5EEDC}\textbf{0.95} &
  {\color[HTML]{808080} 0.04} &
  \cellcolor[HTML]{C7EEE7}\textbf{13.51} &
  {\color[HTML]{808080} 0.84} &
  \cellcolor[HTML]{CAEDFB}\textbf{0.72} &
  {\color[HTML]{808080} 0.02} \\ \cline{2-15} 
\multirow{-8}{*}{SCAMPS} &
  \textbf{MMRPhys} &
  \textbf{TSFM} &
  \cellcolor[HTML]{C4EFDA}\textbf{1.27} &
  {\color[HTML]{808080} 0.93} &
  \cellcolor[HTML]{C6EFE8}\textbf{7.28} &
  {\color[HTML]{808080} 2.59} &
  \cellcolor[HTML]{C4EFDE}\textbf{2.24} &
  {\color[HTML]{808080} 1.71} &
  \cellcolor[HTML]{C5EEDC}\textbf{0.95} &
  {\color[HTML]{808080} 0.04} &
  \cellcolor[HTML]{C8EDED}12.97 &
  {\color[HTML]{808080} 0.81} &
  \cellcolor[HTML]{CBECFA}\textbf{0.72} &
  {\color[HTML]{808080} 0.02} \\ \hline
 &
  PhysNet &
  - &
  \cellcolor[HTML]{D9EAF0}10.38 &
  {\color[HTML]{808080} 2.40} &
  \cellcolor[HTML]{DDE9EC}21.14 &
  {\color[HTML]{808080} 3.90} &
  \cellcolor[HTML]{DCE9ED}20.91 &
  {\color[HTML]{808080} 4.97} &
  \cellcolor[HTML]{E3E7E7}0.660 &
  {\color[HTML]{808080} 0.10} &
  \cellcolor[HTML]{CDECF9}11.01 &
  {\color[HTML]{808080} 0.97} &
  \cellcolor[HTML]{CAEDFA}0.72 &
  {\color[HTML]{808080} 0.02} \\ \cline{2-15} 
 &
  PhysFormer &
  MHSA* &
  \cellcolor[HTML]{D6EBF2}8.90 &
  {\color[HTML]{808080} 2.15} &
  \cellcolor[HTML]{D9EAF0}18.77 &
  {\color[HTML]{808080} 3.67} &
  \cellcolor[HTML]{D9EAF0}17.68 &
  {\color[HTML]{808080} 4.52} &
  \cellcolor[HTML]{DEE8EC}0.710 &
  {\color[HTML]{808080} 0.09} &
  \cellcolor[HTML]{D5EAF2}8.73 &
  {\color[HTML]{808080} 1.02} &
  \cellcolor[HTML]{D2EBF5}0.66 &
  {\color[HTML]{808080} 0.02} \\ \cline{2-15} 
 &
  EfficientPhys &
  SASN \dag &
  \cellcolor[HTML]{CDEDF9}4.71 &
  {\color[HTML]{808080} 1.79} &
  \cellcolor[HTML]{D1ECF6}14.52 &
  {\color[HTML]{808080} 3.65} &
  \cellcolor[HTML]{CDEDF9}7.63 &
  {\color[HTML]{808080} 2.97} &
  \cellcolor[HTML]{D3EBF4}0.800 &
  {\color[HTML]{808080} 0.08} &
  \cellcolor[HTML]{D5EAF3}8.77 &
  {\color[HTML]{808080} 1.00} &
  \cellcolor[HTML]{D2EBF5}0.66 &
  {\color[HTML]{808080} 0.02} \\ \cline{2-15} 
 &
  FactorizePhys &
  FSAM \ddag &
  \cellcolor[HTML]{C1F0CB}0.48 &
  {\color[HTML]{808080} 0.17} &
  \cellcolor[HTML]{C1F0CA}1.39 &
  {\color[HTML]{808080} 0.35} &
  \cellcolor[HTML]{C1F0CB}0.72 &
  {\color[HTML]{808080} 0.28} &
  \cellcolor[HTML]{C2EFC9}0.998 &
  {\color[HTML]{808080} 0.01} &
  \cellcolor[HTML]{C6EEDF}14.16 &
  {\color[HTML]{808080} 0.83} &
  \cellcolor[HTML]{C2EFCE}\textbf{0.78} &
  {\color[HTML]{808080} 0.02} \\ \cline{2-15} 
 &
  FactorizePhys &
  \textbf{TSFM} &
  \cellcolor[HTML]{C1F0C8}\textbf{0.30} &
  {\color[HTML]{808080} 0.12} &
  \cellcolor[HTML]{C1F0C8}\textbf{1.00} &
  {\color[HTML]{808080} 0.29} &
  \cellcolor[HTML]{C1F0C8}\textbf{0.41} &
  {\color[HTML]{808080} 0.18} &
  \cellcolor[HTML]{C1F0C8}\textbf{0.999} &
  {\color[HTML]{808080} 0.01} &
  \cellcolor[HTML]{C4EFD5}\textbf{15.06} &
  {\color[HTML]{808080} 0.90} &
  \cellcolor[HTML]{C2EFCE}\textbf{0.78} &
  {\color[HTML]{808080} 0.02} \\ \cline{2-15} 
\multirow{-8}{*}{UBFC-rPPG} &
  \textbf{MMRPhys} &
  \textbf{TSFM} &
  \cellcolor[HTML]{C1F0C8}0.34 &
  {\color[HTML]{808080} 0.13} &
  \cellcolor[HTML]{C1F0C8}1.02 &
  {\color[HTML]{808080} 0.29} &
  \cellcolor[HTML]{C1F0C9}0.50 &
  {\color[HTML]{808080} 0.19} &
  \cellcolor[HTML]{C1F0C8}\textbf{0.999} &
  {\color[HTML]{808080} 0.01} &
  \cellcolor[HTML]{C6EEE2}13.90 &
  {\color[HTML]{808080} 0.88} &
  \cellcolor[HTML]{C3EFD3}0.77 &
  {\color[HTML]{808080} 0.02} \\ \hline \hline
\multicolumn{15}{l}{\textbf{Performance Evaluation on UBFC-rPPG Dataset   \cite{bobbia2019unsupervised}}} \\ \hline \hline
 &
  PhysNet &
  - &
  \cellcolor[HTML]{D3EBF4}3.09 &
  {\color[HTML]{808080} 1.79} &
  \cellcolor[HTML]{E1E8E9}10.72 &
  {\color[HTML]{808080} 4.24} &
  \cellcolor[HTML]{CFECF7}2.83 &
  {\color[HTML]{808080} 1.44} &
  \cellcolor[HTML]{D9E9EF}0.810 &
  {\color[HTML]{808080} 0.10} &
  \cellcolor[HTML]{D3EBF4}7.13 &
  {\color[HTML]{808080} 1.53} &
  \cellcolor[HTML]{D0EBF6}0.81 &
  {\color[HTML]{808080} 0.02} \\ \cline{2-15} 
 &
  PhysFormer &
  MHSA* &
  \cellcolor[HTML]{F4E4DB}9.88 &
  {\color[HTML]{808080} 2.95} &
  \cellcolor[HTML]{FBE2D5}19.59 &
  {\color[HTML]{808080} 5.35} &
  \cellcolor[HTML]{E7E7E5}8.72 &
  {\color[HTML]{808080} 2.42} &
  \cellcolor[HTML]{F7E3D8}0.440 &
  {\color[HTML]{808080} 0.16} &
  \cellcolor[HTML]{EBE5E1}2.80 &
  {\color[HTML]{808080} 2.21} &
  \cellcolor[HTML]{DEE8EB}0.70 &
  {\color[HTML]{808080} 0.03} \\ \cline{2-15} 
 &
  EfficientPhys &
  SASN \dag &
  \cellcolor[HTML]{C9EEF9}1.14 &
  {\color[HTML]{808080} 0.45} &
  \cellcolor[HTML]{CAEDFB}2.85 &
  {\color[HTML]{808080} 0.88} &
  \cellcolor[HTML]{CAEDFB}1.42 &
  {\color[HTML]{808080} 0.58} &
  \cellcolor[HTML]{CBECFA}0.987 &
  {\color[HTML]{808080} 0.03} &
  \cellcolor[HTML]{C9EDF3}8.71 &
  {\color[HTML]{808080} 1.23} &
  \cellcolor[HTML]{CCECF9}0.84 &
  {\color[HTML]{808080} 0.01} \\ \cline{2-15} 
 &
  FactorizePhys &
  FSAM \ddag &
  \cellcolor[HTML]{C8EEF3}1.04 &
  {\color[HTML]{808080} 0.38} &
  \cellcolor[HTML]{C8EEF2}2.40 &
  {\color[HTML]{808080} 0.69} &
  \cellcolor[HTML]{C8EEF3}1.23 &
  {\color[HTML]{808080} 0.48} &
  \cellcolor[HTML]{C9EDF2}0.990 &
  {\color[HTML]{808080} 0.03} &
  \cellcolor[HTML]{C8EDEC}8.84 &
  {\color[HTML]{808080} 1.31} &
  \cellcolor[HTML]{C8EDED}0.86 &
  {\color[HTML]{808080} 0.01} \\ \cline{2-15} 
 &
  FactorizePhys &
  \textbf{TSFM} &
  \cellcolor[HTML]{CAEDFB}1.28 &
  {\color[HTML]{808080} 0.48} &
  \cellcolor[HTML]{CBEDFA}3.06 &
  {\color[HTML]{808080} 0.89} &
  \cellcolor[HTML]{CAEDFB}1.49 &
  {\color[HTML]{808080} 0.59} &
  \cellcolor[HTML]{CBECFA}0.984 &
  {\color[HTML]{808080} 0.03} &
  \cellcolor[HTML]{C7EEE8}8.92 &
  {\color[HTML]{808080} 1.28} &
  \cellcolor[HTML]{C7EEE6}\textbf{0.87} &
  {\color[HTML]{808080} 0.01} \\ \cline{2-15} 
\multirow{-8}{*}{iBVP} &
  \textbf{MMRPhys} &
  \textbf{TSFM} &
  \cellcolor[HTML]{C1F0C9}\textbf{0.43} &
  {\color[HTML]{808080} 0.23} &
  \cellcolor[HTML]{C1F0C8}\textbf{1.37} &
  {\color[HTML]{808080} 0.51} &
  \cellcolor[HTML]{C1F0C9}\textbf{0.41} &
  {\color[HTML]{808080} 0.20} &
  \cellcolor[HTML]{C1F0C8}\textbf{0.997} &
  {\color[HTML]{808080} 0.01} &
  \cellcolor[HTML]{C5EEDB}\textbf{9.15} &
  {\color[HTML]{808080} 1.20} &
  \cellcolor[HTML]{C6EEE5}\textbf{0.87} &
  {\color[HTML]{808080} 0.01} \\ \hline
 &
  PhysNet &
  - &
  \cellcolor[HTML]{CAEDFB}1.23 &
  {\color[HTML]{808080} 0.41} &
  \cellcolor[HTML]{CAEDFB}2.65 &
  {\color[HTML]{808080} 0.70} &
  \cellcolor[HTML]{CAEDFB}1.42 &
  {\color[HTML]{808080} 0.50} &
  \cellcolor[HTML]{CBECFA}0.988 &
  {\color[HTML]{808080} 0.03} &
  \cellcolor[HTML]{CCECFA}8.34 &
  {\color[HTML]{808080} 1.22} &
  \cellcolor[HTML]{CAEDFB}0.85 &
  {\color[HTML]{808080} 0.01} \\ \cline{2-15} 
 &
  PhysFormer &
  MHSA* &
  \cellcolor[HTML]{C8EEF1}1.01 &
  {\color[HTML]{808080} 0.38} &
  \cellcolor[HTML]{C8EEF2}2.40 &
  {\color[HTML]{808080} 0.69} &
  \cellcolor[HTML]{C8EEF2}1.21 &
  {\color[HTML]{808080} 0.48} &
  \cellcolor[HTML]{C9EDF2}0.990 &
  {\color[HTML]{808080} 0.03} &
  \cellcolor[HTML]{CBECFA}8.42 &
  {\color[HTML]{808080} 1.24} &
  \cellcolor[HTML]{CAEDFB}0.85 &
  {\color[HTML]{808080} 0.01} \\ \cline{2-15} 
 &
  EfficientPhys &
  SASN \dag &
  \cellcolor[HTML]{CBEDFB}1.41 &
  {\color[HTML]{808080} 0.49} &
  \cellcolor[HTML]{CBEDFA}3.16 &
  {\color[HTML]{808080} 0.93} &
  \cellcolor[HTML]{CBEDFB}1.68 &
  {\color[HTML]{808080} 0.64} &
  \cellcolor[HTML]{CBECFA}0.982 &
  {\color[HTML]{808080} 0.03} &
  \cellcolor[HTML]{D4EAF3}6.87 &
  {\color[HTML]{808080} 1.15} &
  \cellcolor[HTML]{D2EBF4}0.79 &
  {\color[HTML]{808080} 0.02} \\ \cline{2-15} 
 &
  FactorizePhys &
  FSAM \ddag &
  \cellcolor[HTML]{C8EEF3}1.04 &
  {\color[HTML]{808080} 0.38} &
  \cellcolor[HTML]{C8EEF4}2.44 &
  {\color[HTML]{808080} 0.69} &
  \cellcolor[HTML]{C8EEF3}1.23 &
  {\color[HTML]{808080} 0.48} &
  \cellcolor[HTML]{CAEDF8}0.989 &
  {\color[HTML]{808080} 0.03} &
  \cellcolor[HTML]{C7EEEA}8.88 &
  {\color[HTML]{808080} 1.30} &
  \cellcolor[HTML]{C5EEDF}0.87 &
  {\color[HTML]{808080} 0.01} \\ \cline{2-15} 
 &
  FactorizePhys &
  \textbf{TSFM} &
  \cellcolor[HTML]{C5EFE1}0.77 &
  {\color[HTML]{808080} 0.33} &
  \cellcolor[HTML]{C5EFE4}2.05 &
  {\color[HTML]{808080} 0.64} &
  \cellcolor[HTML]{C5EFE0}0.86 &
  {\color[HTML]{808080} 0.39} &
  \cellcolor[HTML]{C6EEE0}0.993 &
  {\color[HTML]{808080} 0.02} &
  \cellcolor[HTML]{C7EEE5}8.97 &
  {\color[HTML]{808080} 1.29} &
  \cellcolor[HTML]{C4EFD6}0.88 &
  {\color[HTML]{808080} 0.01} \\ \cline{2-15} 
\multirow{-8}{*}{PURE} &
  \textbf{MMRPhys} &
  \textbf{TSFM} &
  \cellcolor[HTML]{C1F0C8}\textbf{0.40} &
  {\color[HTML]{808080} 0.23} &
  \cellcolor[HTML]{C1F0C8}\textbf{1.36} &
  {\color[HTML]{808080} 0.51} &
  \cellcolor[HTML]{C1F0C8}\textbf{0.38} &
  {\color[HTML]{808080} 0.20} &
  \cellcolor[HTML]{C1F0C8}\textbf{0.997} &
  {\color[HTML]{808080} 0.01} &
  \cellcolor[HTML]{C1F0C8}\textbf{9.49} &
  {\color[HTML]{808080} 1.20} &
  \cellcolor[HTML]{C1F0C8}\textbf{0.89} &
  {\color[HTML]{808080} 0.01} \\ \hline
 &
  PhysNet &
  - &
  \cellcolor[HTML]{FBE2D5}11.24 &
  {\color[HTML]{808080} 2.63} &
  \cellcolor[HTML]{F8E3D7}18.81 &
  {\color[HTML]{808080} 4.71} &
  \cellcolor[HTML]{FBE2D5}13.55 &
  {\color[HTML]{808080} 3.81} &
  \cellcolor[HTML]{FBE2D5}0.380 &
  {\color[HTML]{808080} 0.17} &
  \cellcolor[HTML]{FBE2D5}-0.09 &
  {\color[HTML]{808080} 1.02} &
  \cellcolor[HTML]{FBE2D5}0.48 &
  {\color[HTML]{808080} 0.03} \\ \cline{2-15} 
 &
  PhysFormer &
  MHSA* &
  \cellcolor[HTML]{EDE6E0}8.42 &
  {\color[HTML]{808080} 2.72} &
  \cellcolor[HTML]{F5E4DA}17.73 &
  {\color[HTML]{808080} 5.09} &
  \cellcolor[HTML]{F1E5DD}11.27 &
  {\color[HTML]{808080} 4.24} &
  \cellcolor[HTML]{F3E3DB}0.490 &
  {\color[HTML]{808080} 0.16} &
  \cellcolor[HTML]{EEE5DF}2.29 &
  {\color[HTML]{808080} 1.33} &
  \cellcolor[HTML]{EAE5E2}0.61 &
  {\color[HTML]{808080} 0.03} \\ \cline{2-15} 
 &
  EfficientPhys &
  SASN \dag &
  \cellcolor[HTML]{CEECF8}2.18 &
  {\color[HTML]{808080} 0.75} &
  \cellcolor[HTML]{D0ECF7}4.82 &
  {\color[HTML]{808080} 1.43} &
  \cellcolor[HTML]{CDEDF8}2.35 &
  {\color[HTML]{808080} 0.76} &
  \cellcolor[HTML]{CDECF9}0.960 &
  {\color[HTML]{808080} 0.05} &
  \cellcolor[HTML]{E2E7E8}4.40 &
  {\color[HTML]{808080} 1.03} &
  \cellcolor[HTML]{E2E7E8}0.67 &
  {\color[HTML]{808080} 0.01} \\ \cline{2-15} 
 &
  FactorizePhys &
  FSAM \ddag &
  \cellcolor[HTML]{CAEDFB}1.17 &
  {\color[HTML]{808080} 0.40} &
  \cellcolor[HTML]{C9EEF9}2.56 &
  {\color[HTML]{808080} 0.70} &
  \cellcolor[HTML]{C9EEF9}1.35 &
  {\color[HTML]{808080} 0.49} &
  \cellcolor[HTML]{CAEDF8}0.989 &
  {\color[HTML]{808080} 0.03} &
  \cellcolor[HTML]{CBECFA}8.41 &
  {\color[HTML]{808080} 1.19} &
  \cellcolor[HTML]{CEECF7}0.82 &
  {\color[HTML]{808080} 0.01} \\ \cline{2-15} 
 &
  FactorizePhys &
  \textbf{TSFM} &
  \cellcolor[HTML]{C7EEED}0.96 &
  {\color[HTML]{808080} 0.38} &
  \cellcolor[HTML]{C9EEF5}2.48 &
  {\color[HTML]{808080} 0.69} &
  \cellcolor[HTML]{C7EEEE}1.14 &
  {\color[HTML]{808080} 0.48} &
  \cellcolor[HTML]{C9EDF2}0.990 &
  {\color[HTML]{808080} 0.03} &
  \cellcolor[HTML]{C7EEE8}8.91 &
  {\color[HTML]{808080} 1.25} &
  \cellcolor[HTML]{CAEDF8}\textbf{0.85} &
  {\color[HTML]{808080} 0.01} \\ \cline{2-15} 
\multirow{-8}{*}{SCAMPS} &
  \textbf{MMRPhys} &
  \textbf{TSFM} &
  \cellcolor[HTML]{C5EFE1}\textbf{0.77} &
  {\color[HTML]{808080} \textbf{0.32}} &
  \cellcolor[HTML]{C5EFE2}\textbf{2.00} &
  {\color[HTML]{808080} 0.60} &
  \cellcolor[HTML]{C5EFDE}\textbf{0.83} &
  {\color[HTML]{808080} 0.36} &
  \cellcolor[HTML]{C6EEE0}\textbf{0.993} &
  {\color[HTML]{808080} 0.02} &
  \cellcolor[HTML]{C6EEE2}\textbf{9.03} &
  {\color[HTML]{808080} 1.16} &
  \cellcolor[HTML]{CAEDF8}\textbf{0.85} &
  {\color[HTML]{808080} 0.01} \\ \hline \hline
\end{tabular*}
    \footnotesize
    \begin{flushleft}
    \ddag FSAM: Factorized Self-Attention Module \cite{joshi2024factorizephys}; \dag SASN: Self-Attention Shifted Network \cite{liu2023efficientphys}; \\
    * MHSA: Temporal Difference Multi-Head Self-Attention \cite{yu2022physformer};  TSFM: Proposed Attention Module; SE: Standard Error; \\
    Cell color scale: Within each comparison group (delineated by double horizontal lines), values are highlighted using \\ 
    3-color scale (green = best, orange = worst, blue = midpoint; Microsoft Excel conditional formatting).\\
    \end{flushleft}
\end{table*}

\subsection{Detailed Multi-task rRSP and rPPG Evaluation}\label{appendix multi task evaluation}

\begin{sidewaystable*}[ht]
\caption{BP4D+ Fold1: multi-task Performance Evaluation for Simultaneous rRSP and rPPG Estimation}
\label{tab:results multi-task fold1}
\centering
\fontsize{6}{4}\selectfont
\setlength{\tabcolsep}{0.5pt}
\renewcommand{\arraystretch}{2.5}
\begin{tabular*}{\columnwidth}{@{\extracolsep\fill}lllcccccccccccccccccccccccc}
\hline
\multicolumn{1}{c}{} &
  \multicolumn{1}{c}{} &
  \multicolumn{1}{c}{} &
  \multicolumn{12}{c}{\textbf{rRSP, RR}} &
  \multicolumn{12}{c}{\textbf{rPPG, HR}} \\ \cline{4-27} 
\multicolumn{1}{c}{} &
  \multicolumn{1}{c}{} &
  \multicolumn{1}{c}{} &
  \multicolumn{2}{c}{\textbf{MAE ↓}} &
  \multicolumn{2}{c}{\textbf{RMSE ↓}} &
  \multicolumn{2}{c}{\textbf{MAPE ↓}} &
  \multicolumn{2}{c}{\textbf{Corr ↑}} &
  \multicolumn{2}{c}{\textbf{SNR ↑}} &
  \multicolumn{2}{c}{\textbf{MACC ↑}} &
  \multicolumn{2}{c}{\textbf{MAE ↓}} &
  \multicolumn{2}{c}{\textbf{RMSE ↓}} &
  \multicolumn{2}{c}{\textbf{MAPE ↓}} &
  \multicolumn{2}{c}{\textbf{Corr ↑}} &
  \multicolumn{2}{c}{\textbf{SNR ↑}} &
  \multicolumn{2}{c}{\textbf{MACC ↑}} \\ \cline{4-27} 
\multicolumn{1}{c}{\multirow{-3}{*}{\textbf{Model}}} &
  \multicolumn{1}{c}{\multirow{-3}{*}{\textbf{\begin{tabular}[c]{@{}c@{}}Image\\ Modality\end{tabular}}}} &
  \multicolumn{1}{c}{\multirow{-3}{*}{\textbf{\begin{tabular}[c]{@{}c@{}}Spatial\\ Resolution\end{tabular}}}} &
  \textbf{Avg} &
  {\color[HTML]{808080} \textbf{SE}} &
  \textbf{Avg} &
  {\color[HTML]{808080} \textbf{SE}} &
  \textbf{Avg} &
  {\color[HTML]{808080} \textbf{SE}} &
  \textbf{Avg} &
  {\color[HTML]{808080} \textbf{SE}} &
  \textbf{Avg} &
  {\color[HTML]{808080} \textbf{SE}} &
  \textbf{Avg} &
  {\color[HTML]{808080} \textbf{SE}} &
  \textbf{Avg} &
  {\color[HTML]{808080} \textbf{SE}} &
  \textbf{Avg} &
  {\color[HTML]{808080} \textbf{SE}} &
  \textbf{Avg} &
  {\color[HTML]{808080} \textbf{SE}} &
  \textbf{Avg} &
  {\color[HTML]{808080} \textbf{SE}} &
  \textbf{Avg} &
  {\color[HTML]{808080} \textbf{SE}} &
  \textbf{Avg} &
  {\color[HTML]{808080} \textbf{SE}} \\ \hline \hline
 &
  \cellcolor[HTML]{F2F2F2}\begin{tabular}[c]{@{}l@{}}Big: RGB \\ Small: RGB\end{tabular} &
  \cellcolor[HTML]{F2F2F2}\begin{tabular}[c]{@{}l@{}}Big:   \(72 \times 72\)\\ Small: \(9 \times 9\)\end{tabular} &
  \cellcolor[HTML]{F8E3D8}4.937 &
  {\color[HTML]{808080} 0.20} &
  \cellcolor[HTML]{F9E3D7}6.361 &
  {\color[HTML]{808080} 2.59} &
  \cellcolor[HTML]{F6E3DA}30.720 &
  {\color[HTML]{808080} 1.40} &
  \cellcolor[HTML]{F8E2D9}0.072 &
  {\color[HTML]{808080} 0.05} &
  \cellcolor[HTML]{FAE2D6}4.399 &
  {\color[HTML]{808080} 0.41} &
  \cellcolor[HTML]{FBE2D5}0.529 &
  {\color[HTML]{808080} 0.01} &
  \cellcolor[HTML]{DBE9F7}2.842 &
  {\color[HTML]{808080} 0.28} &
  \cellcolor[HTML]{DBE9F7}6.274 &
  {\color[HTML]{808080} 0.71} &
  \cellcolor[HTML]{DBE9F7}3.332 &
  {\color[HTML]{808080} 0.31} &
  \cellcolor[HTML]{DCE8F6}0.873 &
  {\color[HTML]{808080} 0.02} &
  \cellcolor[HTML]{E0E7F2}6.081 &
  {\color[HTML]{808080} 0.37} &
  \cellcolor[HTML]{E3E7EE}0.649 &
  {\color[HTML]{808080} 0.01} \\ \cline{2-27} 
 &
  \cellcolor[HTML]{D9D9D9}\begin{tabular}[c]{@{}l@{}}Big: Thermal \\ Small: Thermal\end{tabular} &
  \cellcolor[HTML]{D9D9D9}\begin{tabular}[c]{@{}l@{}}Big:   \(72 \times 72\)\\ Small: \(9 \times 9\)\end{tabular} &
  \cellcolor[HTML]{E9E6E8}4.166 &
  {\color[HTML]{808080} 0.21} &
  \cellcolor[HTML]{EFE5E2}6.019 &
  {\color[HTML]{808080} 2.91} &
  \cellcolor[HTML]{F4E4DC}30.317 &
  {\color[HTML]{808080} 1.83} &
  \cellcolor[HTML]{E4E6ED}0.243 &
  {\color[HTML]{808080} 0.05} &
  \cellcolor[HTML]{CBEDDB}8.735 &
  {\color[HTML]{808080} 0.46} &
  \cellcolor[HTML]{E1E7F1}0.646 &
  {\color[HTML]{808080} 0.01} &
  \cellcolor[HTML]{FBE2D5}18.298 &
  {\color[HTML]{808080} 0.81} &
  \cellcolor[HTML]{F8E3D8}24.599 &
  {\color[HTML]{808080} 1.50} &
  \cellcolor[HTML]{FAE3D6}21.251 &
  {\color[HTML]{808080} 0.89} &
  \cellcolor[HTML]{F6E3DB}0.111 &
  {\color[HTML]{808080} 0.05} &
  \cellcolor[HTML]{FBE2D5}-6.557 &
  {\color[HTML]{808080} 0.27} &
  \cellcolor[HTML]{FBE2D5}0.310 &
  {\color[HTML]{808080} 0.01} \\ \cline{2-27} 
\multirow{-3}{*}{BigSmall} &
  \cellcolor[HTML]{BFBFBF}\begin{tabular}[c]{@{}l@{}}Big: Thermal \\ Small: RGB\end{tabular} &
  \cellcolor[HTML]{BFBFBF}\begin{tabular}[c]{@{}l@{}}Big:   \(72 \times 72\)\\ Small: \(9 \times 9\)\end{tabular} &
  \cellcolor[HTML]{FBE2D5}5.074 &
  {\color[HTML]{808080} 0.19} &
  \cellcolor[HTML]{FAE3D6}6.409 &
  {\color[HTML]{808080} 2.52} &
  \cellcolor[HTML]{F9E3D7}31.287 &
  {\color[HTML]{808080} 1.33} &
  \cellcolor[HTML]{FBE2D5}0.041 &
  {\color[HTML]{808080} 0.05} &
  \cellcolor[HTML]{FBE2D5}4.306 &
  {\color[HTML]{808080} 0.41} &
  \cellcolor[HTML]{FBE2D5}0.532 &
  {\color[HTML]{808080} 0.01} &
  \cellcolor[HTML]{DBE9F7}2.999 &
  {\color[HTML]{808080} 0.32} &
  \cellcolor[HTML]{DCE9F6}7.069 &
  {\color[HTML]{808080} 0.80} &
  \cellcolor[HTML]{DCE9F6}3.653 &
  {\color[HTML]{808080} 0.41} &
  \cellcolor[HTML]{DDE8F5}0.840 &
  {\color[HTML]{808080} 0.03} &
  \cellcolor[HTML]{E0E7F2}6.102 &
  {\color[HTML]{808080} 0.37} &
  \cellcolor[HTML]{E3E7EE}0.647 &
  {\color[HTML]{808080} 0.01} \\ \hline
 &
  \cellcolor[HTML]{F2F2F2} &
  \cellcolor[HTML]{F2F2F2}\(72   \times 72\) &
  \cellcolor[HTML]{F1E5E0}4.576 &
  {\color[HTML]{808080} 0.20} &
  \cellcolor[HTML]{F1E5E0}6.097 &
  {\color[HTML]{808080} 0.35} &
  \cellcolor[HTML]{EDE5E4}28.722 &
  {\color[HTML]{808080} 1.46} &
  \cellcolor[HTML]{F7E2D9}0.075 &
  {\color[HTML]{808080} 0.05} &
  \cellcolor[HTML]{FBE2D5}4.271 &
  {\color[HTML]{808080} 0.36} &
  \cellcolor[HTML]{F4E3DC}0.562 &
  {\color[HTML]{808080} 0.01} &
  \cellcolor[HTML]{C6EFD3}1.816 &
  {\color[HTML]{808080} 0.21} &
  \cellcolor[HTML]{C7EFD5}4.705 &
  {\color[HTML]{808080} 0.52} &
  \cellcolor[HTML]{C6EFD2}2.053 &
  {\color[HTML]{808080} 0.22} &
  \cellcolor[HTML]{C8EED5}0.928 &
  {\color[HTML]{808080} 0.02} &
  \cellcolor[HTML]{C7EED4}8.666 &
  {\color[HTML]{808080} 0.44} &
  \cellcolor[HTML]{C7EED2}\textbf{0.774} &
  {\color[HTML]{808080} 0.01} \\ \cline{3-27} 
 &
  \cellcolor[HTML]{F2F2F2} &
  \cellcolor[HTML]{F2F2F2}\(36 \times 36\) &
  \cellcolor[HTML]{F4E4DC}4.765 &
  {\color[HTML]{808080} 0.21} &
  \cellcolor[HTML]{FAE3D6}6.390 &
  {\color[HTML]{808080} 0.38} &
  \cellcolor[HTML]{F4E4DD}30.209 &
  {\color[HTML]{808080} 1.57} &
  \cellcolor[HTML]{FBE2D5}0.037 &
  {\color[HTML]{808080} 0.05} &
  \cellcolor[HTML]{FBE2D5}4.257 &
  {\color[HTML]{808080} 0.35} &
  \cellcolor[HTML]{F3E3DD}0.565 &
  {\color[HTML]{808080} 0.01} &
  \cellcolor[HTML]{C9EED8}1.865 &
  {\color[HTML]{808080} 0.21} &
  \cellcolor[HTML]{C8EFD5}4.710 &
  {\color[HTML]{808080} 0.52} &
  \cellcolor[HTML]{C9EED7}2.114 &
  {\color[HTML]{808080} 0.22} &
  \cellcolor[HTML]{C7EED4}0.928 &
  {\color[HTML]{808080} 0.02} &
  \cellcolor[HTML]{CBEDDB}8.638 &
  {\color[HTML]{808080} 0.45} &
  \cellcolor[HTML]{C3EFCC}\textbf{0.774} &
  {\color[HTML]{808080} 0.01} \\ \cline{3-27} 
 &
  \multirow{-3}{*}{\cellcolor[HTML]{F2F2F2}\begin{tabular}[c]{@{}l@{}}RSP: RGB\\ BVP: RGB\end{tabular}} &
  \cellcolor[HTML]{F2F2F2}\(9 \times 9\) &
  \cellcolor[HTML]{EEE5E3}4.418 &
  {\color[HTML]{808080} 0.21} &
  \cellcolor[HTML]{F2E4DF}6.121 &
  {\color[HTML]{808080} 0.38} &
  \cellcolor[HTML]{F1E5E0}29.534 &
  {\color[HTML]{808080} 1.67} &
  \cellcolor[HTML]{F4E3DC}0.102 &
  {\color[HTML]{808080} 0.05} &
  \cellcolor[HTML]{EEE4E3}6.031 &
  {\color[HTML]{808080} 0.40} &
  \cellcolor[HTML]{EFE4E2}0.586 &
  {\color[HTML]{808080} 0.01} &
  \cellcolor[HTML]{D3EBEA}2.031 &
  {\color[HTML]{808080} 0.23} &
  \cellcolor[HTML]{D5EBEE}5.136 &
  {\color[HTML]{808080} 0.55} &
  \cellcolor[HTML]{D2ECE9}2.301 &
  {\color[HTML]{808080} 0.24} &
  \cellcolor[HTML]{D5EAEE}0.914 &
  {\color[HTML]{808080} 0.02} &
  \cellcolor[HTML]{D6EAF0}8.563 &
  {\color[HTML]{808080} 0.45} &
  \cellcolor[HTML]{DBE8F7}0.770 &
  {\color[HTML]{808080} 0.01} \\ \cline{2-27} 
 &
  \cellcolor[HTML]{D9D9D9} &
  \cellcolor[HTML]{D9D9D9}\(72   \times 72\) &
  \cellcolor[HTML]{C2F0CB}3.141 &
  {\color[HTML]{808080} 0.20} &
  \cellcolor[HTML]{CBEEDC}5.062 &
  {\color[HTML]{808080} 0.35} &
  \cellcolor[HTML]{C6EFD1}22.977 &
  {\color[HTML]{808080} 1.67} &
  \cellcolor[HTML]{D1EBE6}0.365 &
  {\color[HTML]{808080} 0.05} &
  \cellcolor[HTML]{D5EAEE}8.648 &
  {\color[HTML]{808080} 0.45} &
  \cellcolor[HTML]{CEECE0}0.680 &
  {\color[HTML]{808080} 0.01} &
  \cellcolor[HTML]{F8E3D8}17.087 &
  {\color[HTML]{808080} 0.90} &
  \cellcolor[HTML]{F9E3D7}24.957 &
  {\color[HTML]{808080} 1.63} &
  \cellcolor[HTML]{FAE3D6}21.222 &
  {\color[HTML]{808080} 1.19} &
  \cellcolor[HTML]{FAE2D6}-0.024 &
  {\color[HTML]{808080} 0.05} &
  \cellcolor[HTML]{F7E2D9}-4.665 &
  {\color[HTML]{808080} 0.25} &
  \cellcolor[HTML]{F9E2D7}0.349 &
  {\color[HTML]{808080} 0.01} \\ \cline{3-27} 
 &
  \cellcolor[HTML]{D9D9D9} &
  \cellcolor[HTML]{D9D9D9}\(36 \times 36\) &
  \cellcolor[HTML]{C2F0CB}3.141 &
  {\color[HTML]{808080} 0.19} &
  \cellcolor[HTML]{CAEED9}5.040 &
  {\color[HTML]{808080} 0.35} &
  \cellcolor[HTML]{C7EFD4}23.083 &
  {\color[HTML]{808080} 1.66} &
  \cellcolor[HTML]{D1EBE6}0.365 &
  {\color[HTML]{808080} 0.05} &
  \cellcolor[HTML]{DAE9F8}8.598 &
  {\color[HTML]{808080} 0.45} &
  \cellcolor[HTML]{D0EBE5}0.679 &
  {\color[HTML]{808080} 0.01} &
  \cellcolor[HTML]{F5E4DB}15.839 &
  {\color[HTML]{808080} 0.90} &
  \cellcolor[HTML]{F7E3D9}24.194 &
  {\color[HTML]{808080} 1.66} &
  \cellcolor[HTML]{F7E3DA}19.399 &
  {\color[HTML]{808080} 1.17} &
  \cellcolor[HTML]{FAE2D6}-0.021 &
  {\color[HTML]{808080} 0.05} &
  \cellcolor[HTML]{F7E2D9}-4.526 &
  {\color[HTML]{808080} 0.26} &
  \cellcolor[HTML]{F9E2D8}0.349 &
  {\color[HTML]{808080} 0.01} \\ \cline{3-27} 
 &
  \multirow{-3}{*}{\cellcolor[HTML]{D9D9D9}\begin{tabular}[c]{@{}l@{}}RSP: Thermal\\ BVP: Thermal\end{tabular}} &
  \cellcolor[HTML]{D9D9D9}\(9 \times 9\) &
  \cellcolor[HTML]{C1F0C8}3.124 &
  {\color[HTML]{808080} 0.18} &
  \cellcolor[HTML]{C1F0C8}\textbf{4.880} &
  {\color[HTML]{808080} 0.32} &
  \cellcolor[HTML]{C1F0C8}\textbf{22.514} &
  {\color[HTML]{808080} 1.58} &
  \cellcolor[HTML]{C1F0C8}\textbf{0.423} &
  {\color[HTML]{808080} 0.04} &
  \cellcolor[HTML]{D3EBEA}8.666 &
  {\color[HTML]{808080} 0.42} &
  \cellcolor[HTML]{DBE8F7}0.675 &
  {\color[HTML]{808080} 0.01} &
  \cellcolor[HTML]{F5E4DB}15.820 &
  {\color[HTML]{808080} 0.89} &
  \cellcolor[HTML]{F7E3D9}24.008 &
  {\color[HTML]{808080} 1.67} &
  \cellcolor[HTML]{F7E3D9}19.488 &
  {\color[HTML]{808080} 1.17} &
  \cellcolor[HTML]{F8E2D8}0.037 &
  {\color[HTML]{808080} 0.05} &
  \cellcolor[HTML]{F7E2D9}-4.518 &
  {\color[HTML]{808080} 0.27} &
  \cellcolor[HTML]{F8E2D8}0.356 &
  {\color[HTML]{808080} 0.01} \\ \cline{2-27} 
 &
  \cellcolor[HTML]{BFBFBF} &
  \cellcolor[HTML]{BFBFBF}\(72   \times 72\) &
  \cellcolor[HTML]{D4EBEE}3.322 &
  {\color[HTML]{808080} 0.19} &
  \cellcolor[HTML]{CFEDE3}5.126 &
  {\color[HTML]{808080} 0.34} &
  \cellcolor[HTML]{D6EBF0}24.402 &
  {\color[HTML]{808080} 1.66} &
  \cellcolor[HTML]{D8E9F3}0.340 &
  {\color[HTML]{808080} 0.05} &
  \cellcolor[HTML]{C1F0C8}\textbf{8.824} &
  {\color[HTML]{808080} 0.46} &
  \cellcolor[HTML]{D1EBE5}0.679 &
  {\color[HTML]{808080} 0.01} &
  \cellcolor[HTML]{C2F0CA}1.738 &
  {\color[HTML]{808080} 0.21} &
  \cellcolor[HTML]{C2F0CB}4.532 &
  {\color[HTML]{808080} 0.51} &
  \cellcolor[HTML]{C2F0CB}1.976 &
  {\color[HTML]{808080} 0.22} &
  \cellcolor[HTML]{C3EFCC}0.933 &
  {\color[HTML]{808080} 0.02} &
  \cellcolor[HTML]{C2EFC9}8.705 &
  {\color[HTML]{808080} 0.44} &
  \cellcolor[HTML]{C6EED1}\textbf{0.774} &
  {\color[HTML]{808080} 0.01} \\ \cline{3-27} 
 &
  \cellcolor[HTML]{BFBFBF} &
  \cellcolor[HTML]{BFBFBF}\(36 \times 36\) &
  \cellcolor[HTML]{CBEEDC}3.227 &
  {\color[HTML]{808080} 0.19} &
  \cellcolor[HTML]{CBEEDB}5.056 &
  {\color[HTML]{808080} 0.34} &
  \cellcolor[HTML]{C9EED8}23.274 &
  {\color[HTML]{808080} 1.63} &
  \cellcolor[HTML]{D4EAED}0.352 &
  {\color[HTML]{808080} 0.05} &
  \cellcolor[HTML]{D1EBE6}8.683 &
  {\color[HTML]{808080} 0.45} &
  \cellcolor[HTML]{CFECE2}0.679 &
  {\color[HTML]{808080} 0.01} &
  \cellcolor[HTML]{C5EFD0}1.790 &
  {\color[HTML]{808080} 0.21} &
  \cellcolor[HTML]{C4F0CD}4.579 &
  {\color[HTML]{808080} 0.52} &
  \cellcolor[HTML]{C4F0CE}2.016 &
  {\color[HTML]{808080} 0.22} &
  \cellcolor[HTML]{C4EFCD}0.932 &
  {\color[HTML]{808080} 0.02} &
  \cellcolor[HTML]{C6EED1}8.676 &
  {\color[HTML]{808080} 0.44} &
  \cellcolor[HTML]{C8EED4}0.773 &
  {\color[HTML]{808080} 0.01} \\ \cline{3-27} 
\multirow{-9}{*}{\begin{tabular}[c]{@{}l@{}}MMRPhys \\ with FSAM \dag\end{tabular}} &
  \multirow{-3}{*}{\cellcolor[HTML]{BFBFBF}\begin{tabular}[c]{@{}l@{}}RSP: Thermal\\ BVP: RGB\end{tabular}} &
  \cellcolor[HTML]{BFBFBF}\(9 \times 9\) &
  \cellcolor[HTML]{C8EED7}3.201 &
  {\color[HTML]{808080} 0.20} &
  \cellcolor[HTML]{CDEDE0}5.099 &
  {\color[HTML]{808080} 0.34} &
  \cellcolor[HTML]{CBEEDC}23.459 &
  {\color[HTML]{808080} 1.72} &
  \cellcolor[HTML]{D1EBE7}0.364 &
  {\color[HTML]{808080} 0.05} &
  \cellcolor[HTML]{DBE8F7}8.589 &
  {\color[HTML]{808080} 0.42} &
  \cellcolor[HTML]{DAE9F8}0.675 &
  {\color[HTML]{808080} 0.01} &
  \cellcolor[HTML]{DAE9F8}2.157 &
  {\color[HTML]{808080} 0.24} &
  \cellcolor[HTML]{DAE9F8}5.372 &
  {\color[HTML]{808080} 0.56} &
  \cellcolor[HTML]{DAE9F8}2.481 &
  {\color[HTML]{808080} 0.27} &
  \cellcolor[HTML]{DBE8F7}0.904 &
  {\color[HTML]{808080} 0.02} &
  \cellcolor[HTML]{DBE8F7}8.506 &
  {\color[HTML]{808080} 0.45} &
  \cellcolor[HTML]{DAE9F8}0.770 &
  {\color[HTML]{808080} 0.01} \\ \hline
 &
  \cellcolor[HTML]{F2F2F2} &
  \cellcolor[HTML]{F2F2F2}\(72 \times   72\) &
  \cellcolor[HTML]{EFE5E2}4.477 &
  {\color[HTML]{808080} 0.21} &
  \cellcolor[HTML]{F2E4DF}6.114 &
  {\color[HTML]{808080} 0.37} &
  \cellcolor[HTML]{EBE6E6}28.464 &
  {\color[HTML]{808080} 1.52} &
  \cellcolor[HTML]{EFE4E2}0.150 &
  {\color[HTML]{808080} 0.05} &
  \cellcolor[HTML]{FBE2D5}4.328 &
  {\color[HTML]{808080} 0.37} &
  \cellcolor[HTML]{F5E3DB}0.556 &
  {\color[HTML]{808080} 0.01} &
  \cellcolor[HTML]{C4EFCE}1.779 &
  {\color[HTML]{808080} 0.21} &
  \cellcolor[HTML]{C6EFD3}4.668 &
  {\color[HTML]{808080} 0.53} &
  \cellcolor[HTML]{C3F0CD}2.000 &
  {\color[HTML]{808080} 0.22} &
  \cellcolor[HTML]{C7EED2}0.929 &
  {\color[HTML]{808080} 0.02} &
  \cellcolor[HTML]{C3EFCB}8.696 &
  {\color[HTML]{808080} 0.44} &
  \cellcolor[HTML]{C1F0C8}\textbf{0.774} &
  {\color[HTML]{808080} 0.01} \\ \cline{3-27} 
 &
  \cellcolor[HTML]{F2F2F2} &
  \cellcolor[HTML]{F2F2F2}\(36 \times 36\) &
  \cellcolor[HTML]{F7E3D9}4.881 &
  {\color[HTML]{808080} 0.20} &
  \cellcolor[HTML]{FBE2D5}6.410 &
  {\color[HTML]{808080} 0.37} &
  \cellcolor[HTML]{FBE2D5}31.594 &
  {\color[HTML]{808080} 1.57} &
  \cellcolor[HTML]{FBE2D5}0.044 &
  {\color[HTML]{808080} 0.05} &
  \cellcolor[HTML]{FBE2D5}4.351 &
  {\color[HTML]{808080} 0.34} &
  \cellcolor[HTML]{F7E3DA}0.551 &
  {\color[HTML]{808080} 0.01} &
  \cellcolor[HTML]{C9EED9}1.872 &
  {\color[HTML]{808080} 0.21} &
  \cellcolor[HTML]{C7EFD4}4.695 &
  {\color[HTML]{808080} 0.52} &
  \cellcolor[HTML]{C9EED8}2.119 &
  {\color[HTML]{808080} 0.22} &
  \cellcolor[HTML]{C7EED4}0.928 &
  {\color[HTML]{808080} 0.02} &
  \cellcolor[HTML]{C8EED5}8.661 &
  {\color[HTML]{808080} 0.45} &
  \cellcolor[HTML]{C3EFCB}\textbf{0.774} &
  {\color[HTML]{808080} 0.01} \\ \cline{3-27} 
 &
  \multirow{-3}{*}{\cellcolor[HTML]{F2F2F2}\begin{tabular}[c]{@{}l@{}}RSP: RGB\\ BVP: RGB\end{tabular}} &
  \cellcolor[HTML]{F2F2F2}\(9 \times 9\) &
  \cellcolor[HTML]{EBE6E6}4.272 &
  {\color[HTML]{808080} 0.20} &
  \cellcolor[HTML]{EBE6E5}5.910 &
  {\color[HTML]{808080} 0.36} &
  \cellcolor[HTML]{EDE5E4}28.736 &
  {\color[HTML]{808080} 1.62} &
  \cellcolor[HTML]{F5E3DC}0.097 &
  {\color[HTML]{808080} 0.05} &
  \cellcolor[HTML]{EFE4E1}5.859 &
  {\color[HTML]{808080} 0.40} &
  \cellcolor[HTML]{EFE4E2}0.586 &
  {\color[HTML]{808080} 0.01} &
  \cellcolor[HTML]{D4EBEE}2.062 &
  {\color[HTML]{808080} 0.23} &
  \cellcolor[HTML]{D6EBF0}5.170 &
  {\color[HTML]{808080} 0.55} &
  \cellcolor[HTML]{D3EBEC}2.334 &
  {\color[HTML]{808080} 0.24} &
  \cellcolor[HTML]{D6EAF1}0.912 &
  {\color[HTML]{808080} 0.02} &
  \cellcolor[HTML]{D9E9F6}8.543 &
  {\color[HTML]{808080} 0.45} &
  \cellcolor[HTML]{DAE9F8}0.770 &
  {\color[HTML]{808080} 0.01} \\ \cline{2-27} 
 &
  \cellcolor[HTML]{D9D9D9} &
  \cellcolor[HTML]{D9D9D9}\(72   \times 72\) &
  \cellcolor[HTML]{D8EAF4}3.357 &
  {\color[HTML]{808080} 0.20} &
  \cellcolor[HTML]{DAE9F8}5.327 &
  {\color[HTML]{808080} 0.37} &
  \cellcolor[HTML]{D8EAF4}24.589 &
  {\color[HTML]{808080} 1.78} &
  \cellcolor[HTML]{DAE9F8}0.330 &
  {\color[HTML]{808080} 0.05} &
  \cellcolor[HTML]{CCEDDC}8.734 &
  {\color[HTML]{808080} 0.44} &
  \cellcolor[HTML]{CBEDDB}0.681 &
  {\color[HTML]{808080} 0.01} &
  \cellcolor[HTML]{F6E3DA}16.129 &
  {\color[HTML]{808080} 0.89} &
  \cellcolor[HTML]{F7E3D9}24.149 &
  {\color[HTML]{808080} 1.62} &
  \cellcolor[HTML]{F7E3D9}19.870 &
  {\color[HTML]{808080} 1.16} &
  \cellcolor[HTML]{F8E2D8}0.051 &
  {\color[HTML]{808080} 0.05} &
  \cellcolor[HTML]{F7E2D9}-4.570 &
  {\color[HTML]{808080} 0.26} &
  \cellcolor[HTML]{F8E2D8}0.353 &
  {\color[HTML]{808080} 0.01} \\ \cline{3-27} 
 &
  \cellcolor[HTML]{D9D9D9} &
  \cellcolor[HTML]{D9D9D9}\(36 \times 36\) &
  \cellcolor[HTML]{D3EBEB}3.310 &
  {\color[HTML]{808080} 0.20} &
  \cellcolor[HTML]{D6EBF0}5.251 &
  {\color[HTML]{808080} 0.36} &
  \cellcolor[HTML]{D2EBEA}24.126 &
  {\color[HTML]{808080} 1.70} &
  \cellcolor[HTML]{D8E9F4}0.339 &
  {\color[HTML]{808080} 0.05} &
  \cellcolor[HTML]{DBE8F7}8.507 &
  {\color[HTML]{808080} 0.44} &
  \cellcolor[HTML]{D6EAEF}0.676 &
  {\color[HTML]{808080} 0.01} &
  \cellcolor[HTML]{F9E3D7}17.470 &
  {\color[HTML]{808080} 0.96} &
  \cellcolor[HTML]{FBE2D5}26.145 &
  {\color[HTML]{808080} 1.73} &
  \cellcolor[HTML]{FBE2D5}21.673 &
  {\color[HTML]{808080} 1.27} &
  \cellcolor[HTML]{FBE2D5}-0.059 &
  {\color[HTML]{808080} 0.05} &
  \cellcolor[HTML]{F7E2D9}-4.521 &
  {\color[HTML]{808080} 0.25} &
  \cellcolor[HTML]{F9E2D7}0.348 &
  {\color[HTML]{808080} 0.01} \\ \cline{3-27} 
 &
  \multirow{-3}{*}{\cellcolor[HTML]{D9D9D9}\begin{tabular}[c]{@{}l@{}}RSP: Thermal\\ BVP: Thermal\end{tabular}} &
  \cellcolor[HTML]{D9D9D9}\(9 \times 9\) &
  \cellcolor[HTML]{C1F0C8}\textbf{3.122} &
  {\color[HTML]{808080} 0.19} &
  \cellcolor[HTML]{C2F0C9}4.898 &
  {\color[HTML]{808080} 0.33} &
  \cellcolor[HTML]{C4F0CE}22.806 &
  {\color[HTML]{808080} 1.64} &
  \cellcolor[HTML]{C3EFCC}0.417 &
  {\color[HTML]{808080} 0.04} &
  \cellcolor[HTML]{C7EED3}8.775 &
  {\color[HTML]{808080} 0.43} &
  \cellcolor[HTML]{CBEDDA}0.681 &
  {\color[HTML]{808080} 0.01} &
  \cellcolor[HTML]{F6E3DA}15.999 &
  {\color[HTML]{808080} 0.94} &
  \cellcolor[HTML]{F8E3D8}24.866 &
  {\color[HTML]{808080} 1.75} &
  \cellcolor[HTML]{F7E3D9}19.754 &
  {\color[HTML]{808080} 1.23} &
  \cellcolor[HTML]{F9E2D8}0.028 &
  {\color[HTML]{808080} 0.05} &
  \cellcolor[HTML]{F7E2D9}-4.474 &
  {\color[HTML]{808080} 0.26} &
  \cellcolor[HTML]{F8E2D8}0.354 &
  {\color[HTML]{808080} 0.01} \\ \cline{2-27} 
 &
  \cellcolor[HTML]{BFBFBF} &
  \cellcolor[HTML]{BFBFBF}\(72   \times 72\) &
  \cellcolor[HTML]{DAE9F8}3.375 &
  {\color[HTML]{808080} 0.20} &
  \cellcolor[HTML]{D8EAF5}5.295 &
  {\color[HTML]{808080} 0.36} &
  \cellcolor[HTML]{DAE9F8}24.759 &
  {\color[HTML]{808080} 1.74} &
  \cellcolor[HTML]{D9E9F6}0.334 &
  {\color[HTML]{808080} 0.05} &
  \cellcolor[HTML]{D3EBEB}8.663 &
  {\color[HTML]{808080} 0.44} &
  \cellcolor[HTML]{C1F0C8}\textbf{0.684} &
  {\color[HTML]{808080} 0.01} &
  \cellcolor[HTML]{C1F0C8}\textbf{1.716} &
  {\color[HTML]{808080} 0.20} &
  \cellcolor[HTML]{C1F0C8}\textbf{4.481} &
  {\color[HTML]{808080} 0.52} &
  \cellcolor[HTML]{C1F0C8}\textbf{1.942} &
  {\color[HTML]{808080} 0.21} &
  \cellcolor[HTML]{C1F0C8}\textbf{0.935} &
  {\color[HTML]{808080} 0.02} &
  \cellcolor[HTML]{C1F0C8}\textbf{8.706} &
  {\color[HTML]{808080} 0.44} &
  \cellcolor[HTML]{C7EED2}\textbf{0.774} &
  {\color[HTML]{808080} 0.01} \\ \cline{3-27} 
 &
  \cellcolor[HTML]{BFBFBF} &
  \cellcolor[HTML]{BFBFBF}\(36 \times 36\) &
  \cellcolor[HTML]{D0ECE5}3.275 &
  {\color[HTML]{808080} 0.20} &
  \cellcolor[HTML]{D2ECE9}5.180 &
  {\color[HTML]{808080} 0.35} &
  \cellcolor[HTML]{D1ECE7}23.967 &
  {\color[HTML]{808080} 1.68} &
  \cellcolor[HTML]{D3EBEA}0.358 &
  {\color[HTML]{808080} 0.05} &
  \cellcolor[HTML]{D6EAF0}8.639 &
  {\color[HTML]{808080} 0.44} &
  \cellcolor[HTML]{CCEDDD}0.680 &
  {\color[HTML]{808080} 0.01} &
  \cellcolor[HTML]{C6EFD2}1.811 &
  {\color[HTML]{808080} 0.21} &
  \cellcolor[HTML]{C3F0CD}4.577 &
  {\color[HTML]{808080} 0.52} &
  \cellcolor[HTML]{C6EFD2}2.056 &
  {\color[HTML]{808080} 0.22} &
  \cellcolor[HTML]{C4EFCD}0.932 &
  {\color[HTML]{808080} 0.02} &
  \cellcolor[HTML]{C5EFCF}8.681 &
  {\color[HTML]{808080} 0.44} &
  \cellcolor[HTML]{C8EED6}0.773 &
  {\color[HTML]{808080} 0.01} \\ \cline{3-27} 
\multirow{-9}{*}{\begin{tabular}[c]{@{}l@{}}MMRPhys\\ with \AM{}\end{tabular}} &
  \multirow{-3}{*}{\cellcolor[HTML]{BFBFBF}\begin{tabular}[c]{@{}l@{}}RSP: Thermal\\ BVP: RGB\end{tabular}} &
  \cellcolor[HTML]{BFBFBF}\(9 \times 9\) &
  \cellcolor[HTML]{DAE9F8}3.384 &
  {\color[HTML]{808080} 0.20} &
  \cellcolor[HTML]{DAE9F8}5.316 &
  {\color[HTML]{808080} 0.35} &
  \cellcolor[HTML]{DAE9F8}24.846 &
  {\color[HTML]{808080} 1.78} &
  \cellcolor[HTML]{DDE8F5}0.311 &
  {\color[HTML]{808080} 0.05} &
  \cellcolor[HTML]{CEECE0}8.714 &
  {\color[HTML]{808080} 0.43} &
  \cellcolor[HTML]{D0EBE5}0.679 &
  {\color[HTML]{808080} 0.01} &
  \cellcolor[HTML]{DAE9F8}2.153 &
  {\color[HTML]{808080} 0.24} &
  \cellcolor[HTML]{DAE9F8}5.291 &
  {\color[HTML]{808080} 0.55} &
  \cellcolor[HTML]{DAE9F8}2.458 &
  {\color[HTML]{808080} 0.25} &
  \cellcolor[HTML]{DAE9F8}0.908 &
  {\color[HTML]{808080} 0.02} &
  \cellcolor[HTML]{DAE9F8}8.532 &
  {\color[HTML]{808080} 0.45} &
  \cellcolor[HTML]{D9E9F6}0.770 &
  {\color[HTML]{808080} 0.01} \\ \hline \hline
\end{tabular*}
    \footnotesize
    \begin{flushleft}
    \dag FSAM: Factorized Self-Attention Module \cite{joshi2024factorizephys};  TSFM: Proposed Target Signal Constrained Factorization Module;  \\
    Avg: Average; SE: Standard Error; Cell color scale: Within each comparison group (delineated by double horizontal lines), values are highlighted using 3-color scale (green = best, orange = worst, blue = midpoint; Microsoft Excel conditional formatting).\\
    \end{flushleft}
\end{sidewaystable*}

\begin{sidewaystable*}[ht]
\caption{BP4D+ Fold2: multi-task Performance Evaluation for Simultaneous rRSP and rPPG Estimation}
\label{tab:results multi-task fold2}
\centering
\fontsize{6}{4}\selectfont
\setlength{\tabcolsep}{0.5pt}
\renewcommand{\arraystretch}{2.5}
\begin{tabular*}{\textwidth}{@{\extracolsep\fill}lllcccccccccccccccccccccccc}
\hline
\multicolumn{1}{c}{} &
  \multicolumn{1}{c}{} &
  \multicolumn{1}{c}{} &
  \multicolumn{12}{c}{\textbf{rRSP, RR}} &
  \multicolumn{12}{c}{\textbf{rPPG, HR}} \\ \cline{4-27} 
\multicolumn{1}{c}{} &
  \multicolumn{1}{c}{} &
  \multicolumn{1}{c}{} &
  \multicolumn{2}{c}{\textbf{MAE ↓}} &
  \multicolumn{2}{c}{\textbf{RMSE ↓}} &
  \multicolumn{2}{c}{\textbf{MAPE ↓}} &
  \multicolumn{2}{c}{\textbf{Corr ↑}} &
  \multicolumn{2}{c}{\textbf{SNR ↑}} &
  \multicolumn{2}{c}{\textbf{MACC ↑}} &
  \multicolumn{2}{c}{\textbf{MAE ↓}} &
  \multicolumn{2}{c}{\textbf{RMSE ↓}} &
  \multicolumn{2}{c}{\textbf{MAPE ↓}} &
  \multicolumn{2}{c}{\textbf{Corr ↑}} &
  \multicolumn{2}{c}{\textbf{SNR ↑}} &
  \multicolumn{2}{c}{\textbf{MACC ↑}} \\ \cline{4-27} 
\multicolumn{1}{c}{\multirow{-3}{*}{\textbf{Model}}} &
  \multicolumn{1}{c}{\multirow{-3}{*}{\textbf{\begin{tabular}[c]{@{}c@{}}Image\\ Modality\end{tabular}}}} &
  \multicolumn{1}{c}{\multirow{-3}{*}{\textbf{\begin{tabular}[c]{@{}c@{}}Spatial\\ Resolution\end{tabular}}}} &
  \textbf{Avg} &
  {\color[HTML]{808080} \textbf{SE}} &
  \textbf{Avg} &
  {\color[HTML]{808080} \textbf{SE}} &
  \textbf{Avg} &
  {\color[HTML]{808080} \textbf{SE}} &
  \textbf{Avg} &
  {\color[HTML]{808080} \textbf{SE}} &
  \textbf{Avg} &
  {\color[HTML]{808080} \textbf{SE}} &
  \textbf{Avg} &
  {\color[HTML]{808080} \textbf{SE}} &
  \textbf{Avg} &
  {\color[HTML]{808080} \textbf{SE}} &
  \textbf{Avg} &
  {\color[HTML]{808080} \textbf{SE}} &
  \textbf{Avg} &
  {\color[HTML]{808080} \textbf{SE}} &
  \textbf{Avg} &
  {\color[HTML]{808080} \textbf{SE}} &
  \textbf{Avg} &
  {\color[HTML]{808080} \textbf{SE}} &
  \textbf{Avg} &
  {\color[HTML]{808080} \textbf{SE}} \\ \hline \hline
 &
  \cellcolor[HTML]{F2F2F2}\begin{tabular}[c]{@{}l@{}}Big: RGB \\ Small: RGB\end{tabular} &
  \cellcolor[HTML]{F2F2F2}\begin{tabular}[c]{@{}l@{}}Big:   \(72 \times 72\)\\ Small: \(9 \times 9\)\end{tabular} &
  \cellcolor[HTML]{FBE2D5}4.994 &
  {\color[HTML]{808080} 0.20} &
  \cellcolor[HTML]{FBE2D5}6.444 &
  {\color[HTML]{808080} 2.89} &
  \cellcolor[HTML]{FBE2D5}30.253 &
  {\color[HTML]{808080} 1.30} &
  \cellcolor[HTML]{FBE2D5}-0.009 &
  {\color[HTML]{808080} 0.05} &
  \cellcolor[HTML]{FBE2D5}4.023 &
  {\color[HTML]{808080} 0.36} &
  \cellcolor[HTML]{FBE2D5}0.529 &
  {\color[HTML]{808080} 0.01} &
  \cellcolor[HTML]{DBE9F7}2.462 &
  {\color[HTML]{808080} 0.23} &
  \cellcolor[HTML]{D3EBEB}5.382 &
  {\color[HTML]{808080} 0.55} &
  \cellcolor[HTML]{DBE9F7}3.043 &
  {\color[HTML]{808080} 0.31} &
  \cellcolor[HTML]{DAE9F7}0.937 &
  {\color[HTML]{808080} 0.02} &
  \cellcolor[HTML]{E0E7F2}7.034 &
  {\color[HTML]{808080} 0.37} &
  \cellcolor[HTML]{E3E7EF}0.685 &
  {\color[HTML]{808080} 0.01} \\ \cline{2-27} 
 &
  \cellcolor[HTML]{D9D9D9}\begin{tabular}[c]{@{}l@{}}Big: Thermal \\ Small: Thermal\end{tabular} &
  \cellcolor[HTML]{D9D9D9}\begin{tabular}[c]{@{}l@{}}Big:   \(72 \times 72\)\\ Small: \(9 \times 9\)\end{tabular} &
  \cellcolor[HTML]{E1E8F0}3.465 &
  {\color[HTML]{808080} 0.19} &
  \cellcolor[HTML]{E3E7EE}5.327 &
  {\color[HTML]{808080} 2.44} &
  \cellcolor[HTML]{E9E6E8}25.420 &
  {\color[HTML]{808080} 1.69} &
  \cellcolor[HTML]{DBE8F7}0.352 &
  {\color[HTML]{808080} 0.05} &
  \cellcolor[HTML]{DBE8F7}8.789 &
  {\color[HTML]{808080} 0.42} &
  \cellcolor[HTML]{E0E7F2}0.684 &
  {\color[HTML]{808080} 0.01} &
  \cellcolor[HTML]{FBE2D5}18.563 &
  {\color[HTML]{808080} 0.83} &
  \cellcolor[HTML]{F6E3DA}25.394 &
  {\color[HTML]{808080} 1.49} &
  \cellcolor[HTML]{F8E3D9}21.759 &
  {\color[HTML]{808080} 0.95} &
  \cellcolor[HTML]{F6E3DB}0.184 &
  {\color[HTML]{808080} 0.05} &
  \cellcolor[HTML]{FBE2D5}-6.153 &
  {\color[HTML]{808080} 0.28} &
  \cellcolor[HTML]{FBE2D5}0.318 &
  {\color[HTML]{808080} 0.01} \\ \cline{2-27} 
\multirow{-3}{*}{BigSmall} &
  \cellcolor[HTML]{BFBFBF}\begin{tabular}[c]{@{}l@{}}Big: Thermal \\ Small: RGB\end{tabular} &
  \cellcolor[HTML]{BFBFBF}\begin{tabular}[c]{@{}l@{}}Big:   \(72 \times 72\)\\ Small: \(9 \times 9\)\end{tabular} &
  \cellcolor[HTML]{F8E3D8}4.845 &
  {\color[HTML]{808080} 0.19} &
  \cellcolor[HTML]{F7E3D9}6.266 &
  {\color[HTML]{808080} 2.59} &
  \cellcolor[HTML]{F7E3D9}29.309 &
  {\color[HTML]{808080} 1.25} &
  \cellcolor[HTML]{F8E2D8}0.027 &
  {\color[HTML]{808080} 0.05} &
  \cellcolor[HTML]{FBE2D5}3.907 &
  {\color[HTML]{808080} 0.38} &
  \cellcolor[HTML]{FBE2D5}0.531 &
  {\color[HTML]{808080} 0.01} &
  \cellcolor[HTML]{DAE9F8}2.369 &
  {\color[HTML]{808080} 0.22} &
  \cellcolor[HTML]{C1F0C8}\textbf{5.195} &
  {\color[HTML]{808080} 0.54} &
  \cellcolor[HTML]{DAE9F7}2.948 &
  {\color[HTML]{808080} 0.31} &
  \cellcolor[HTML]{C1F0C8}\textbf{0.941} &
  {\color[HTML]{808080} 0.02} &
  \cellcolor[HTML]{E0E7F2}6.982 &
  {\color[HTML]{808080} 0.37} &
  \cellcolor[HTML]{E3E7EF}0.684 &
  {\color[HTML]{808080} 0.01} \\ \hline
 &
  \cellcolor[HTML]{F2F2F2} &
  \cellcolor[HTML]{F2F2F2}\(72   \times 72\) &
  \cellcolor[HTML]{F3E4DE}4.530 &
  {\color[HTML]{808080} 0.20} &
  \cellcolor[HTML]{F3E4DE}6.083 &
  {\color[HTML]{808080} 0.36} &
  \cellcolor[HTML]{F2E4DE}27.999 &
  {\color[HTML]{808080} 1.41} &
  \cellcolor[HTML]{F8E2D9}0.034 &
  {\color[HTML]{808080} 0.05} &
  \cellcolor[HTML]{F7E3DA}4.629 &
  {\color[HTML]{808080} 0.35} &
  \cellcolor[HTML]{F4E3DC}0.569 &
  {\color[HTML]{808080} 0.01} &
  \cellcolor[HTML]{C6EFD1}1.846 &
  {\color[HTML]{808080} 0.24} &
  \cellcolor[HTML]{D4EBED}5.391 &
  {\color[HTML]{808080} 0.63} &
  \cellcolor[HTML]{C4F0CE}2.219 &
  {\color[HTML]{808080} 0.31} &
  \cellcolor[HTML]{D1EBE7}0.938 &
  {\color[HTML]{808080} 0.02} &
  \cellcolor[HTML]{CAEDD9}9.635 &
  {\color[HTML]{808080} 0.47} &
  \cellcolor[HTML]{DBE8F7}\textbf{0.803} &
  {\color[HTML]{808080} 0.01} \\ \cline{3-27} 
 &
  \cellcolor[HTML]{F2F2F2} &
  \cellcolor[HTML]{F2F2F2}\(36 \times 36\) &
  \cellcolor[HTML]{EEE5E2}4.266 &
  {\color[HTML]{808080} 0.19} &
  \cellcolor[HTML]{EEE5E2}5.864 &
  {\color[HTML]{808080} 0.35} &
  \cellcolor[HTML]{ECE6E5}26.252 &
  {\color[HTML]{808080} 1.33} &
  \cellcolor[HTML]{F0E4E0}0.117 &
  {\color[HTML]{808080} 0.05} &
  \cellcolor[HTML]{F5E3DB}4.826 &
  {\color[HTML]{808080} 0.32} &
  \cellcolor[HTML]{F5E3DC}0.567 &
  {\color[HTML]{808080} 0.01} &
  \cellcolor[HTML]{DAE9F8}1.943 &
  {\color[HTML]{808080} 0.25} &
  \cellcolor[HTML]{DAE9F8}5.522 &
  {\color[HTML]{808080} 0.64} &
  \cellcolor[HTML]{DAE9F8}2.322 &
  {\color[HTML]{808080} 0.32} &
  \cellcolor[HTML]{DBE8F7}0.935 &
  {\color[HTML]{808080} 0.02} &
  \cellcolor[HTML]{CDECDE}9.629 &
  {\color[HTML]{808080} 0.47} &
  \cellcolor[HTML]{CEECE0}\textbf{0.804} &
  {\color[HTML]{808080} 0.01} \\ \cline{3-27} 
 &
  \multirow{-3}{*}{\cellcolor[HTML]{F2F2F2}\begin{tabular}[c]{@{}l@{}}RSP: RGB\\ BVP: RGB\end{tabular}} &
  \cellcolor[HTML]{F2F2F2}\(9 \times 9\) &
  \cellcolor[HTML]{E9E6E8}3.918 &
  {\color[HTML]{808080} 0.18} &
  \cellcolor[HTML]{E7E7EB}5.489 &
  {\color[HTML]{808080} 0.34} &
  \cellcolor[HTML]{E5E7EC}24.470 &
  {\color[HTML]{808080} 1.29} &
  \cellcolor[HTML]{EAE5E7}0.186 &
  {\color[HTML]{808080} 0.05} &
  \cellcolor[HTML]{EFE4E2}5.762 &
  {\color[HTML]{808080} 0.35} &
  \cellcolor[HTML]{F1E4E0}0.588 &
  {\color[HTML]{808080} 0.01} &
  \cellcolor[HTML]{C1F0C8}\textbf{1.821} &
  {\color[HTML]{808080} 0.24} &
  \cellcolor[HTML]{CCEDDE}5.315 &
  {\color[HTML]{808080} 0.63} &
  \cellcolor[HTML]{C1F0C8}\textbf{2.204} &
  {\color[HTML]{808080} 0.31} &
  \cellcolor[HTML]{C7EED3}0.940 &
  {\color[HTML]{808080} 0.02} &
  \cellcolor[HTML]{D6EAF0}9.608 &
  {\color[HTML]{808080} 0.47} &
  \cellcolor[HTML]{C7EED2}\textbf{0.805} &
  {\color[HTML]{808080} 0.01} \\ \cline{2-27} 
 &
  \cellcolor[HTML]{D9D9D9} &
  \cellcolor[HTML]{D9D9D9}\(72   \times 72\) &
  \cellcolor[HTML]{DAE9F8}3.001 &
  {\color[HTML]{808080} 0.18} &
  \cellcolor[HTML]{DAE9F8}4.866 &
  {\color[HTML]{808080} 0.33} &
  \cellcolor[HTML]{DAE9F8}21.299 &
  {\color[HTML]{808080} 1.56} &
  \cellcolor[HTML]{DAE9F8}0.362 &
  {\color[HTML]{808080} 0.04} &
  \cellcolor[HTML]{D9E9F5}8.817 &
  {\color[HTML]{808080} 0.42} &
  \cellcolor[HTML]{D0ECE3}0.723 &
  {\color[HTML]{808080} 0.01} &
  \cellcolor[HTML]{F9E3D7}17.884 &
  {\color[HTML]{808080} 0.98} &
  \cellcolor[HTML]{F9E3D7}27.113 &
  {\color[HTML]{808080} 1.77} &
  \cellcolor[HTML]{FAE3D7}22.993 &
  {\color[HTML]{808080} 1.37} &
  \cellcolor[HTML]{FBE2D5}0.035 &
  {\color[HTML]{808080} 0.05} &
  \cellcolor[HTML]{F7E2D9}-3.997 &
  {\color[HTML]{808080} 0.26} &
  \cellcolor[HTML]{F9E2D8}0.362 &
  {\color[HTML]{808080} 0.01} \\ \cline{3-27} 
 &
  \cellcolor[HTML]{D9D9D9} &
  \cellcolor[HTML]{D9D9D9}\(36 \times 36\) &
  \cellcolor[HTML]{C8EED6}2.822 &
  {\color[HTML]{808080} 0.17} &
  \cellcolor[HTML]{C1F0C8}\textbf{4.564} &
  {\color[HTML]{808080} 0.30} &
  \cellcolor[HTML]{C7EFD4}19.873 &
  {\color[HTML]{808080} 1.42} &
  \cellcolor[HTML]{C8EED6}0.409 &
  {\color[HTML]{808080} 0.04} &
  \cellcolor[HTML]{DAE9F8}8.800 &
  {\color[HTML]{808080} 0.43} &
  \cellcolor[HTML]{CAEDDA}0.726 &
  {\color[HTML]{808080} 0.01} &
  \cellcolor[HTML]{F8E3D8}17.288 &
  {\color[HTML]{808080} 0.95} &
  \cellcolor[HTML]{F8E3D9}26.203 &
  {\color[HTML]{808080} 1.72} &
  \cellcolor[HTML]{F9E3D8}22.369 &
  {\color[HTML]{808080} 1.35} &
  \cellcolor[HTML]{FAE2D6}0.064 &
  {\color[HTML]{808080} 0.05} &
  \cellcolor[HTML]{F7E2D9}-4.045 &
  {\color[HTML]{808080} 0.27} &
  \cellcolor[HTML]{F9E2D8}0.362 &
  {\color[HTML]{808080} 0.01} \\ \cline{3-27} 
 &
  \multirow{-3}{*}{\cellcolor[HTML]{D9D9D9}\begin{tabular}[c]{@{}l@{}}RSP: Thermal\\ BVP: Thermal\end{tabular}} &
  \cellcolor[HTML]{D9D9D9}\(9 \times 9\) &
  \cellcolor[HTML]{CCEDDE}2.865 &
  {\color[HTML]{808080} 0.18} &
  \cellcolor[HTML]{C9EED8}\textbf{4.668} &
  {\color[HTML]{808080} 0.31} &
  \cellcolor[HTML]{CEEDE2}\textbf{20.432} &
  {\color[HTML]{808080} 1.48} &
  \cellcolor[HTML]{D2EBE8}\textbf{0.385} &
  {\color[HTML]{808080} 0.04} &
  \cellcolor[HTML]{CFECE2}8.924 &
  {\color[HTML]{808080} 0.43} &
  \cellcolor[HTML]{D4EAEB}0.719 &
  {\color[HTML]{808080} 0.01} &
  \cellcolor[HTML]{F6E3DA}16.494 &
  {\color[HTML]{808080} 0.95} &
  \cellcolor[HTML]{F7E3D9}25.684 &
  {\color[HTML]{808080} 1.72} &
  \cellcolor[HTML]{F7E3D9}21.382 &
  {\color[HTML]{808080} 1.33} &
  \cellcolor[HTML]{F8E2D9}0.124 &
  {\color[HTML]{808080} 0.05} &
  \cellcolor[HTML]{F7E3DA}-3.869 &
  {\color[HTML]{808080} 0.27} &
  \cellcolor[HTML]{F8E2D8}0.363 &
  {\color[HTML]{808080} 0.01} \\ \cline{2-27} 
 &
  \cellcolor[HTML]{BFBFBF} &
  \cellcolor[HTML]{BFBFBF}\(72   \times 72\) &
  \cellcolor[HTML]{D7EAF2}2.974 &
  {\color[HTML]{808080} 0.18} &
  \cellcolor[HTML]{D6EAF1}4.828 &
  {\color[HTML]{808080} 0.33} &
  \cellcolor[HTML]{D8EAF4}21.176 &
  {\color[HTML]{808080} 1.57} &
  \cellcolor[HTML]{D9E9F5}0.367 &
  {\color[HTML]{808080} 0.04} &
  \cellcolor[HTML]{D1EBE6}\textbf{8.899} &
  {\color[HTML]{808080} 0.42} &
  \cellcolor[HTML]{CDECDF}0.724 &
  {\color[HTML]{808080} 0.01} &
  \cellcolor[HTML]{C9EED8}1.863 &
  {\color[HTML]{808080} 0.24} &
  \cellcolor[HTML]{CFECE4}5.343 &
  {\color[HTML]{808080} 0.63} &
  \cellcolor[HTML]{CDEDDF}2.261 &
  {\color[HTML]{808080} 0.31} &
  \cellcolor[HTML]{CBEDDB}0.939 &
  {\color[HTML]{808080} 0.02} &
  \cellcolor[HTML]{C1F0C8}\textbf{9.655} &
  {\color[HTML]{808080} 0.48} &
  \cellcolor[HTML]{D5EAEE}\textbf{0.804} &
  {\color[HTML]{808080} 0.01} \\ \cline{3-27} 
 &
  \cellcolor[HTML]{BFBFBF} &
  \cellcolor[HTML]{BFBFBF}\(36 \times 36\) &
  \cellcolor[HTML]{D4EBED}2.945 &
  {\color[HTML]{808080} 0.18} &
  \cellcolor[HTML]{CFEDE2}4.734 &
  {\color[HTML]{808080} 0.31} &
  \cellcolor[HTML]{D1ECE8}20.662 &
  {\color[HTML]{808080} 1.45} &
  \cellcolor[HTML]{DAE9F8}0.362 &
  {\color[HTML]{808080} 0.04} &
  \cellcolor[HTML]{D1EBE7}8.895 &
  {\color[HTML]{808080} 0.42} &
  \cellcolor[HTML]{CBEDDB}0.726 &
  {\color[HTML]{808080} 0.01} &
  \cellcolor[HTML]{D0ECE5}1.895 &
  {\color[HTML]{808080} 0.24} &
  \cellcolor[HTML]{D5EBEE}5.398 &
  {\color[HTML]{808080} 0.63} &
  \cellcolor[HTML]{CEEDE1}2.266 &
  {\color[HTML]{808080} 0.31} &
  \cellcolor[HTML]{CFECE2}0.938 &
  {\color[HTML]{808080} 0.02} &
  \cellcolor[HTML]{C9EDD8}9.637 &
  {\color[HTML]{808080} 0.47} &
  \cellcolor[HTML]{CEECE0}0.804 &
  {\color[HTML]{808080} 0.01} \\ \cline{3-27} 
\multirow{-9}{*}{\begin{tabular}[c]{@{}l@{}}MMRPhys \\ with FSAM \dag\end{tabular}} &
  \multirow{-3}{*}{\cellcolor[HTML]{BFBFBF}\begin{tabular}[c]{@{}l@{}}RSP: Thermal\\ BVP: RGB\end{tabular}} &
  \cellcolor[HTML]{BFBFBF}\(9 \times 9\) &
  \cellcolor[HTML]{C7EFD4}2.810 &
  {\color[HTML]{808080} 0.17} &
  \cellcolor[HTML]{C2F0CA}4.580 &
  {\color[HTML]{808080} 0.30} &
  \cellcolor[HTML]{C6EFD1}19.765 &
  {\color[HTML]{808080} 1.40} &
  \cellcolor[HTML]{C1F0C8}\textbf{0.427} &
  {\color[HTML]{808080} 0.04} &
  \cellcolor[HTML]{D5EAEF}8.854 &
  {\color[HTML]{808080} 0.42} &
  \cellcolor[HTML]{D5EAED}0.719 &
  {\color[HTML]{808080} 0.01} &
  \cellcolor[HTML]{CDEDE0}1.883 &
  {\color[HTML]{808080} 0.24} &
  \cellcolor[HTML]{D4EBED}5.392 &
  {\color[HTML]{808080} 0.63} &
  \cellcolor[HTML]{CCEDDD}2.257 &
  {\color[HTML]{808080} 0.31} &
  \cellcolor[HTML]{D1EBE6}0.938 &
  {\color[HTML]{808080} 0.94} &
  \cellcolor[HTML]{D2EBE9}9.616 &
  {\color[HTML]{808080} 0.47} &
  \cellcolor[HTML]{C1F0C8}\textbf{0.805} &
  {\color[HTML]{808080} 0.01} \\ \hline
 &
  \cellcolor[HTML]{F2F2F2} &
  \cellcolor[HTML]{F2F2F2}\(72 \times   72\) &
  \cellcolor[HTML]{F0E5E1}4.330 &
  {\color[HTML]{808080} 0.19} &
  \cellcolor[HTML]{EEE5E3}5.835 &
  {\color[HTML]{808080} 0.35} &
  \cellcolor[HTML]{EBE6E6}25.988 &
  {\color[HTML]{808080} 1.22} &
  \cellcolor[HTML]{F0E4E0}0.115 &
  {\color[HTML]{808080} 0.05} &
  \cellcolor[HTML]{F7E2D9}4.524 &
  {\color[HTML]{808080} 0.35} &
  \cellcolor[HTML]{F6E3DB}0.561 &
  {\color[HTML]{808080} 0.01} &
  \cellcolor[HTML]{C4EFCF}1.840 &
  {\color[HTML]{808080} 0.24} &
  \cellcolor[HTML]{D0ECE5}5.348 &
  {\color[HTML]{808080} 0.63} &
  \cellcolor[HTML]{C4EFCF}2.222 &
  {\color[HTML]{808080} 0.31} &
  \cellcolor[HTML]{CCEDDC}0.939 &
  {\color[HTML]{808080} 0.02} &
  \cellcolor[HTML]{CEECE1}9.626 &
  {\color[HTML]{808080} 0.47} &
  \cellcolor[HTML]{D5EAEE}\textbf{0.804} &
  {\color[HTML]{808080} 0.01} \\ \cline{3-27} 
 &
  \cellcolor[HTML]{F2F2F2} &
  \cellcolor[HTML]{F2F2F2}\(36 \times 36\) &
  \cellcolor[HTML]{F4E4DC}4.605 &
  {\color[HTML]{808080} 0.20} &
  \cellcolor[HTML]{F5E4DC}6.158 &
  {\color[HTML]{808080} 0.36} &
  \cellcolor[HTML]{F3E4DE}28.171 &
  {\color[HTML]{808080} 1.37} &
  \cellcolor[HTML]{F7E2D9}0.041 &
  {\color[HTML]{808080} 0.05} &
  \cellcolor[HTML]{FAE2D6}4.090 &
  {\color[HTML]{808080} 0.33} &
  \cellcolor[HTML]{FAE2D6}0.538 &
  {\color[HTML]{808080} 0.01} &
  \cellcolor[HTML]{CCEDDE}1.879 &
  {\color[HTML]{808080} 0.24} &
  \cellcolor[HTML]{D5EBEF}5.402 &
  {\color[HTML]{808080} 0.63} &
  \cellcolor[HTML]{CAEEDA}2.249 &
  {\color[HTML]{808080} 0.31} &
  \cellcolor[HTML]{D2EBE7}0.938 &
  {\color[HTML]{808080} 0.02} &
  \cellcolor[HTML]{D6EAEF}9.609 &
  {\color[HTML]{808080} 0.47} &
  \cellcolor[HTML]{CDECDE}\textbf{0.804} &
  {\color[HTML]{808080} 0.01} \\ \cline{3-27} 
 &
  \multirow{-3}{*}{\cellcolor[HTML]{F2F2F2}\begin{tabular}[c]{@{}l@{}}RSP: RGB\\ BVP: RGB\end{tabular}} &
  \cellcolor[HTML]{F2F2F2}\(9 \times 9\) &
  \cellcolor[HTML]{EEE5E2}4.269 &
  {\color[HTML]{808080} 0.20} &
  \cellcolor[HTML]{EFE5E2}5.902 &
  {\color[HTML]{808080} 0.35} &
  \cellcolor[HTML]{EEE5E3}26.790 &
  {\color[HTML]{808080} 1.41} &
  \cellcolor[HTML]{F0E4E1}0.122 &
  {\color[HTML]{808080} 0.05} &
  \cellcolor[HTML]{EFE4E2}5.758 &
  {\color[HTML]{808080} 0.34} &
  \cellcolor[HTML]{F1E4E0}0.589 &
  {\color[HTML]{808080} 0.01} &
  \cellcolor[HTML]{C6EFD2}1.847 &
  {\color[HTML]{808080} 0.24} &
  \cellcolor[HTML]{D0ECE5}5.349 &
  {\color[HTML]{808080} 0.63} &
  \cellcolor[HTML]{C8EED6}2.240 &
  {\color[HTML]{808080} 0.31} &
  \cellcolor[HTML]{CCEDDD}0.939 &
  {\color[HTML]{808080} 0.02} &
  \cellcolor[HTML]{DBE8F7}9.594 &
  {\color[HTML]{808080} 0.47} &
  \cellcolor[HTML]{C7EED3}\textbf{0.805} &
  {\color[HTML]{808080} 0.01} \\ \cline{2-27} 
 &
  \cellcolor[HTML]{D9D9D9} &
  \cellcolor[HTML]{D9D9D9}\(72   \times 72\) &
  \cellcolor[HTML]{D5EBEF}2.955 &
  {\color[HTML]{808080} 0.18} &
  \cellcolor[HTML]{D0ECE5}4.753 &
  {\color[HTML]{808080} 0.32} &
  \cellcolor[HTML]{D9EAF6}21.238 &
  {\color[HTML]{808080} 1.52} &
  \cellcolor[HTML]{CAEDD9}0.404 &
  {\color[HTML]{808080} 0.04} &
  \cellcolor[HTML]{D3EBEA}8.877 &
  {\color[HTML]{808080} 0.42} &
  \cellcolor[HTML]{CAEDD8}0.727 &
  {\color[HTML]{808080} 0.01} &
  \cellcolor[HTML]{FAE3D6}18.417 &
  {\color[HTML]{808080} 1.02} &
  \cellcolor[HTML]{FBE2D5}28.169 &
  {\color[HTML]{808080} 1.84} &
  \cellcolor[HTML]{FBE2D5}23.620 &
  {\color[HTML]{808080} 1.45} &
  \cellcolor[HTML]{FBE2D5}0.018 &
  {\color[HTML]{808080} 0.05} &
  \cellcolor[HTML]{F7E2D9}-3.971 &
  {\color[HTML]{808080} 0.26} &
  \cellcolor[HTML]{F9E2D8}0.361 &
  {\color[HTML]{808080} 0.01} \\ \cline{3-27} 
 &
  \cellcolor[HTML]{D9D9D9} &
  \cellcolor[HTML]{D9D9D9}\(36 \times 36\) &
  \cellcolor[HTML]{CFECE3}2.891 &
  {\color[HTML]{808080} 0.18} &
  \cellcolor[HTML]{CCEDDE}4.703 &
  {\color[HTML]{808080} 0.31} &
  \cellcolor[HTML]{CEEDE1}20.409 &
  {\color[HTML]{808080} 1.47} &
  \cellcolor[HTML]{D3EAEB}0.380 &
  {\color[HTML]{808080} 0.04} &
  \cellcolor[HTML]{DBE8F7}8.798 &
  {\color[HTML]{808080} 0.42} &
  \cellcolor[HTML]{C8EED4}0.728 &
  {\color[HTML]{808080} 0.01} &
  \cellcolor[HTML]{FAE3D6}18.314 &
  {\color[HTML]{808080} 0.97} &
  \cellcolor[HTML]{F9E3D7}27.227 &
  {\color[HTML]{808080} 1.74} &
  \cellcolor[HTML]{FAE3D6}23.347 &
  {\color[HTML]{808080} 1.34} &
  \cellcolor[HTML]{FBE2D5}0.027 &
  {\color[HTML]{808080} 0.05} &
  \cellcolor[HTML]{F7E2D9}-4.165 &
  {\color[HTML]{808080} 0.27} &
  \cellcolor[HTML]{F9E2D7}0.359 &
  {\color[HTML]{808080} 0.01} \\ \cline{3-27} 
 &
  \multirow{-3}{*}{\cellcolor[HTML]{D9D9D9}\begin{tabular}[c]{@{}l@{}}RSP: Thermal\\ BVP: Thermal\end{tabular}} &
  \cellcolor[HTML]{D9D9D9}\(9 \times 9\) &
  \cellcolor[HTML]{DBE9F7}\textbf{3.066} &
  {\color[HTML]{808080} 0.18} &
  \cellcolor[HTML]{DAE9F8}4.897 &
  {\color[HTML]{808080} 0.32} &
  \cellcolor[HTML]{DDE9F5}22.156 &
  {\color[HTML]{808080} 1.56} &
  \cellcolor[HTML]{DCE8F6}0.348 &
  {\color[HTML]{808080} 0.05} &
  \cellcolor[HTML]{CFECE2}8.924 &
  {\color[HTML]{808080} 0.42} &
  \cellcolor[HTML]{DBE8F7}0.712 &
  {\color[HTML]{808080} 0.01} &
  \cellcolor[HTML]{F6E3DA}16.545 &
  {\color[HTML]{808080} 0.93} &
  \cellcolor[HTML]{F6E3DA}25.384 &
  {\color[HTML]{808080} 1.69} &
  \cellcolor[HTML]{F7E3D9}21.295 &
  {\color[HTML]{808080} 1.29} &
  \cellcolor[HTML]{F7E2D9}0.147 &
  {\color[HTML]{808080} 0.05} &
  \cellcolor[HTML]{F7E2D9}-3.906 &
  {\color[HTML]{808080} 0.27} &
  \cellcolor[HTML]{F8E2D8}0.365 &
  {\color[HTML]{808080} 0.01} \\ \cline{2-27} 
 &
  \cellcolor[HTML]{BFBFBF} &
  \cellcolor[HTML]{BFBFBF}\(72   \times 72\) &
  \cellcolor[HTML]{D4EBED}2.945 &
  {\color[HTML]{808080} 0.18} &
  \cellcolor[HTML]{CEEDE1}4.723 &
  {\color[HTML]{808080} 0.31} &
  \cellcolor[HTML]{D5EBF0}20.985 &
  {\color[HTML]{808080} 1.50} &
  \cellcolor[HTML]{C8EED4}0.411 &
  {\color[HTML]{808080} 0.04} &
  \cellcolor[HTML]{D1EBE5}8.903 &
  {\color[HTML]{808080} 0.42} &
  \cellcolor[HTML]{CAEDD9}\textbf{0.726} &
  {\color[HTML]{808080} 0.01} &
  \cellcolor[HTML]{DAE9F8}\textbf{1.942} &
  {\color[HTML]{808080} 0.24} &
  \cellcolor[HTML]{DAE9F8}\textbf{5.446} &
  {\color[HTML]{808080} 0.63} &
  \cellcolor[HTML]{DAE9F8}\textbf{2.350} &
  {\color[HTML]{808080} 0.32} &
  \cellcolor[HTML]{DAE9F8}\textbf{0.937} &
  {\color[HTML]{808080} 0.02} &
  \cellcolor[HTML]{C9EDD7}\textbf{9.637} &
  {\color[HTML]{808080} 0.02} &
  \cellcolor[HTML]{DAE9F8}\textbf{0.804} &
  {\color[HTML]{808080} 0.01} \\ \cline{3-27} 
 &
  \cellcolor[HTML]{BFBFBF} &
  \cellcolor[HTML]{BFBFBF}\(36 \times 36\) &
  \cellcolor[HTML]{C1F0C8}\textbf{2.741} &
  {\color[HTML]{808080} 0.18} &
  \cellcolor[HTML]{C3F0CD}4.596 &
  {\color[HTML]{808080} 0.31} &
  \cellcolor[HTML]{C1F0C8}\textbf{19.371} &
  {\color[HTML]{808080} 1.45} &
  \cellcolor[HTML]{C6EED1}0.415 &
  {\color[HTML]{808080} 0.04} &
  \cellcolor[HTML]{C1F0C8}\textbf{9.061} &
  {\color[HTML]{808080} 0.42} &
  \cellcolor[HTML]{C1F0C8}\textbf{0.733} &
  {\color[HTML]{808080} 0.01} &
  \cellcolor[HTML]{D8EAF5}1.937 &
  {\color[HTML]{808080} 0.25} &
  \cellcolor[HTML]{DAE9F8}5.470 &
  {\color[HTML]{808080} 0.63} &
  \cellcolor[HTML]{D9EAF6}2.318 &
  {\color[HTML]{808080} 0.32} &
  \cellcolor[HTML]{DBE8F7}0.936 &
  {\color[HTML]{808080} 0.02} &
  \cellcolor[HTML]{DAE9F8}9.598 &
  {\color[HTML]{808080} 0.47} &
  \cellcolor[HTML]{CFECE2}0.804 &
  {\color[HTML]{808080} 0.01} \\ \cline{3-27} 
\multirow{-9}{*}{\begin{tabular}[c]{@{}l@{}}MMRPhys\\ with \AM{}\end{tabular}} &
  \multirow{-3}{*}{\cellcolor[HTML]{BFBFBF}\begin{tabular}[c]{@{}l@{}}RSP: Thermal\\ BVP: RGB\end{tabular}} &
  \cellcolor[HTML]{BFBFBF}\(9 \times 9\) &
  \cellcolor[HTML]{C6EFD2}2.799 &
  {\color[HTML]{808080} 0.18} &
  \cellcolor[HTML]{C6EFD2}4.631 &
  {\color[HTML]{808080} 0.32} &
  \cellcolor[HTML]{C9EED7}20.009 &
  {\color[HTML]{808080} 1.48} &
  \cellcolor[HTML]{C7EED3}0.412 &
  {\color[HTML]{808080} 0.04} &
  \cellcolor[HTML]{D0ECE4}8.912 &
  {\color[HTML]{808080} 0.41} &
  \cellcolor[HTML]{DAE9F8}0.714 &
  {\color[HTML]{808080} 0.01} &
  \cellcolor[HTML]{D9EAF7}1.941 &
  {\color[HTML]{808080} 0.25} &
  \cellcolor[HTML]{DAE9F8}5.500 &
  {\color[HTML]{808080} 0.63} &
  \cellcolor[HTML]{D8EAF5}2.316 &
  {\color[HTML]{808080} 0.32} &
  \cellcolor[HTML]{DBE8F7}0.935 &
  {\color[HTML]{808080} 0.02} &
  \cellcolor[HTML]{D9E9F6}9.601 &
  {\color[HTML]{808080} 0.47} &
  \cellcolor[HTML]{C6EED0}\textbf{0.805} &
  {\color[HTML]{808080} 0.01} \\ \hline \hline
\end{tabular*}
    \footnotesize
    \begin{flushleft}
    \dag FSAM: Factorized Self-Attention Module \cite{joshi2024factorizephys};  TSFM: Proposed Target Signal Constrained Factorization Module;  \\
    Avg: Average; SE: Standard Error; Cell color scale: Within each comparison group (delineated by double horizontal lines), values are highlighted using 3-color scale (green = best, orange = worst, blue = midpoint; Microsoft Excel conditional formatting). \\
    \end{flushleft}
\end{sidewaystable*}

\begin{sidewaystable*}[ht]
\caption{BP4D+ Fold3: multi-task Performance Evaluation for Simultaneous rRSP and rPPG Estimation}
\label{tab:results multi-task fold3}
\centering
\fontsize{6}{4}\selectfont
\setlength{\tabcolsep}{0.5pt}
\renewcommand{\arraystretch}{2.5}
\begin{tabular*}{\textheight}{@{\extracolsep\fill}lllcccccccccccccccccccccccc}
\hline
\multicolumn{1}{c}{} &
  \multicolumn{1}{c}{} &
  \multicolumn{1}{c}{} &
  \multicolumn{12}{c}{\textbf{rRSP, RR}} &
  \multicolumn{12}{c}{\textbf{rPPG, HR}} \\ \cline{4-27} 
\multicolumn{1}{c}{} &
  \multicolumn{1}{c}{} &
  \multicolumn{1}{c}{} &
  \multicolumn{2}{c}{\textbf{MAE ↓}} &
  \multicolumn{2}{c}{\textbf{RMSE ↓}} &
  \multicolumn{2}{c}{\textbf{MAPE ↓}} &
  \multicolumn{2}{c}{\textbf{Corr ↑}} &
  \multicolumn{2}{c}{\textbf{SNR ↑}} &
  \multicolumn{2}{c}{\textbf{MACC ↑}} &
  \multicolumn{2}{c}{\textbf{MAE ↓}} &
  \multicolumn{2}{c}{\textbf{RMSE ↓}} &
  \multicolumn{2}{c}{\textbf{MAPE ↓}} &
  \multicolumn{2}{c}{\textbf{Corr ↑}} &
  \multicolumn{2}{c}{\textbf{SNR ↑}} &
  \multicolumn{2}{c}{\textbf{MACC ↑}} \\ \cline{4-27} 
\multicolumn{1}{c}{\multirow{-3}{*}{\textbf{Model}}} &
  \multicolumn{1}{c}{\multirow{-3}{*}{\textbf{\begin{tabular}[c]{@{}c@{}}Image\\ Modality\end{tabular}}}} &
  \multicolumn{1}{c}{\multirow{-3}{*}{\textbf{\begin{tabular}[c]{@{}c@{}}Spatial\\ Resolution\end{tabular}}}} &
  \textbf{Avg} &
  {\color[HTML]{808080} \textbf{SE}} &
  \textbf{Avg} &
  {\color[HTML]{808080} \textbf{SE}} &
  \textbf{Avg} &
  {\color[HTML]{808080} \textbf{SE}} &
  \textbf{Avg} &
  {\color[HTML]{808080} \textbf{SE}} &
  \textbf{Avg} &
  {\color[HTML]{808080} \textbf{SE}} &
  \textbf{Avg} &
  {\color[HTML]{808080} \textbf{SE}} &
  \textbf{Avg} &
  {\color[HTML]{808080} \textbf{SE}} &
  \textbf{Avg} &
  {\color[HTML]{808080} \textbf{SE}} &
  \textbf{Avg} &
  {\color[HTML]{808080} \textbf{SE}} &
  \textbf{Avg} &
  {\color[HTML]{808080} \textbf{SE}} &
  \textbf{Avg} &
  {\color[HTML]{808080} \textbf{SE}} &
  \textbf{Avg} &
  {\color[HTML]{808080} \textbf{SE}} \\ \hline \hline
 &
  \cellcolor[HTML]{F2F2F2}\begin{tabular}[c]{@{}l@{}}Big: RGB \\ Small: RGB\end{tabular} &
  \cellcolor[HTML]{F2F2F2}\begin{tabular}[c]{@{}l@{}}Big:   \(72 \times 72\)\\ Small: \(9 \times 9\)\end{tabular} &
  \cellcolor[HTML]{F6E3DA}4.962 &
  {\color[HTML]{808080} 0.20} &
  \cellcolor[HTML]{F5E4DB}6.361 &
  {\color[HTML]{808080} 2.56} &
  \cellcolor[HTML]{F6E4DB}30.407 &
  {\color[HTML]{808080} 1.41} &
  \cellcolor[HTML]{F3E3DE}0.080 &
  {\color[HTML]{808080} 0.05} &
  \cellcolor[HTML]{FBE2D5}3.699 &
  {\color[HTML]{808080} 0.37} &
  \cellcolor[HTML]{FBE2D5}0.541 &
  {\color[HTML]{808080} 0.01} &
  \cellcolor[HTML]{DDE9F5}3.280 &
  {\color[HTML]{808080} 0.37} &
  \cellcolor[HTML]{DFE8F2}8.092 &
  {\color[HTML]{808080} 0.92} &
  \cellcolor[HTML]{DDE9F5}4.107 &
  {\color[HTML]{808080} 0.48} &
  \cellcolor[HTML]{DFE8F3}0.841 &
  {\color[HTML]{808080} 0.03} &
  \cellcolor[HTML]{E1E7F0}6.686 &
  {\color[HTML]{808080} 0.38} &
  \cellcolor[HTML]{E3E7EF}0.670 &
  {\color[HTML]{808080} 0.01} \\ \cline{2-27} 
 &
  \cellcolor[HTML]{D9D9D9}\begin{tabular}[c]{@{}l@{}}Big: Thermal \\ Small: Thermal\end{tabular} &
  \cellcolor[HTML]{D9D9D9}\begin{tabular}[c]{@{}l@{}}Big:   \(72 \times 72\)\\ Small: \(9 \times 9\)\end{tabular} &
  \cellcolor[HTML]{E1E8F1}3.605 &
  {\color[HTML]{808080} 0.21} &
  \cellcolor[HTML]{E4E7ED}5.506 &
  {\color[HTML]{808080} 2.66} &
  \cellcolor[HTML]{E9E6E8}26.920 &
  {\color[HTML]{808080} 1.86} &
  \cellcolor[HTML]{DFE7F2}0.292 &
  {\color[HTML]{808080} 0.05} &
  \cellcolor[HTML]{DCE8F6}8.931 &
  {\color[HTML]{808080} 0.44} &
  \cellcolor[HTML]{E1E7F1}0.657 &
  {\color[HTML]{808080} 0.01} &
  \cellcolor[HTML]{FBE2D5}19.077 &
  {\color[HTML]{808080} 0.86} &
  \cellcolor[HTML]{FBE2D5}25.750 &
  {\color[HTML]{808080} 1.56} &
  \cellcolor[HTML]{FBE2D5}22.380 &
  {\color[HTML]{808080} 0.98} &
  \cellcolor[HTML]{FAE2D6}0.144 &
  {\color[HTML]{808080} 0.05} &
  \cellcolor[HTML]{FBE2D5}-6.462 &
  {\color[HTML]{808080} 0.31} &
  \cellcolor[HTML]{FBE2D5}0.314 &
  {\color[HTML]{808080} 0.01} \\ \cline{2-27} 
\multirow{-3}{*}{BigSmall} &
  \cellcolor[HTML]{BFBFBF}\begin{tabular}[c]{@{}l@{}}Big: Thermal \\ Small: RGB\end{tabular} &
  \cellcolor[HTML]{BFBFBF}\begin{tabular}[c]{@{}l@{}}Big:   \(72 \times 72\)\\ Small: \(9 \times 9\)\end{tabular} &
  \cellcolor[HTML]{FBE2D5}5.212 &
  {\color[HTML]{808080} 0.20} &
  \cellcolor[HTML]{FBE2D5}6.622 &
  {\color[HTML]{808080} 2.81} &
  \cellcolor[HTML]{FBE2D5}31.774 &
  {\color[HTML]{808080} 1.44} &
  \cellcolor[HTML]{FBE2D5}-0.013 &
  {\color[HTML]{808080} 0.05} &
  \cellcolor[HTML]{FBE2D5}3.551 &
  {\color[HTML]{808080} 0.38} &
  \cellcolor[HTML]{FBE2D5}0.542 &
  {\color[HTML]{808080} 0.01} &
  \cellcolor[HTML]{DCE9F6}2.823 &
  {\color[HTML]{808080} 0.34} &
  \cellcolor[HTML]{DEE8F3}\textbf{7.477} &
  {\color[HTML]{808080} 0.92} &
  \cellcolor[HTML]{DCE9F6}3.529 &
  {\color[HTML]{808080} 0.45} &
  \cellcolor[HTML]{DEE8F4}\textbf{0.863} &
  {\color[HTML]{808080} 0.03} &
  \cellcolor[HTML]{E1E7F1}6.839 &
  {\color[HTML]{808080} 0.38} &
  \cellcolor[HTML]{E3E7EF}0.673 &
  {\color[HTML]{808080} 0.01} \\ \hline
 &
  \cellcolor[HTML]{F2F2F2} &
  \cellcolor[HTML]{F2F2F2}\(72   \times 72\) &
  \cellcolor[HTML]{EDE5E4}4.355 &
  {\color[HTML]{808080} 0.19} &
  \cellcolor[HTML]{ECE6E5}5.866 &
  {\color[HTML]{808080} 0.35} &
  \cellcolor[HTML]{E8E6E9}26.609 &
  {\color[HTML]{808080} 1.33} &
  \cellcolor[HTML]{EBE5E6}0.168 &
  {\color[HTML]{808080} 0.05} &
  \cellcolor[HTML]{F3E3DD}4.937 &
  {\color[HTML]{808080} 0.38} &
  \cellcolor[HTML]{F6E3DA}0.565 &
  {\color[HTML]{808080} 0.01} &
  \cellcolor[HTML]{D3EBEC}1.501 &
  {\color[HTML]{808080} 0.19} &
  \cellcolor[HTML]{D6EBF0}4.120 &
  {\color[HTML]{808080} 0.48} &
  \cellcolor[HTML]{D4EBEC}1.731 &
  {\color[HTML]{808080} 0.22} &
  \cellcolor[HTML]{D6EAF0}0.959 &
  {\color[HTML]{808080} 0.01} &
  \cellcolor[HTML]{C2EFCA}10.303 &
  {\color[HTML]{808080} 0.45} &
  \cellcolor[HTML]{C2EFCA}\textbf{0.798} &
  {\color[HTML]{808080} 0.01} \\ \cline{3-27} 
 &
  \cellcolor[HTML]{F2F2F2} &
  \cellcolor[HTML]{F2F2F2}\(36 \times 36\) &
  \cellcolor[HTML]{EFE5E1}4.515 &
  {\color[HTML]{808080} 0.20} &
  \cellcolor[HTML]{EFE5E2}6.018 &
  {\color[HTML]{808080} 0.36} &
  \cellcolor[HTML]{EDE5E4}27.848 &
  {\color[HTML]{808080} 1.42} &
  \cellcolor[HTML]{F1E4DF}0.099 &
  {\color[HTML]{808080} 0.05} &
  \cellcolor[HTML]{F2E3DE}5.100 &
  {\color[HTML]{808080} 0.37} &
  \cellcolor[HTML]{F5E3DB}0.570 &
  {\color[HTML]{808080} 0.01} &
  \cellcolor[HTML]{D4EBED}1.509 &
  {\color[HTML]{808080} 0.19} &
  \cellcolor[HTML]{D6EAF1}4.136 &
  {\color[HTML]{808080} 0.49} &
  \cellcolor[HTML]{D3EBEC}1.729 &
  {\color[HTML]{808080} 0.21} &
  \cellcolor[HTML]{D6EAF1}0.959 &
  {\color[HTML]{808080} 0.01} &
  \cellcolor[HTML]{C7EED4}10.272 &
  {\color[HTML]{808080} 0.45} &
  \cellcolor[HTML]{C6EED1}\textbf{0.797} &
  {\color[HTML]{808080} 0.01} \\ \cline{3-27} 
 &
  \multirow{-3}{*}{\cellcolor[HTML]{F2F2F2}\begin{tabular}[c]{@{}l@{}}RSP: RGB\\ BVP: RGB\end{tabular}} &
  \cellcolor[HTML]{F2F2F2}\(9 \times 9\) &
  \cellcolor[HTML]{E8E6E9}4.063 &
  {\color[HTML]{808080} 0.19} &
  \cellcolor[HTML]{E7E7EA}5.625 &
  {\color[HTML]{808080} 0.34} &
  \cellcolor[HTML]{E4E7ED}25.355 &
  {\color[HTML]{808080} 1.37} &
  \cellcolor[HTML]{E9E5E8}0.186 &
  {\color[HTML]{808080} 0.05} &
  \cellcolor[HTML]{EFE4E1}5.617 &
  {\color[HTML]{808080} 0.38} &
  \cellcolor[HTML]{F0E4E1}0.593 &
  {\color[HTML]{808080} 0.01} &
  \cellcolor[HTML]{C3F0CC}\textbf{1.279} &
  {\color[HTML]{808080} 0.13} &
  \cellcolor[HTML]{C1F0C9}2.962 &
  {\color[HTML]{808080} 0.27} &
  \cellcolor[HTML]{C3F0CD}\textbf{1.500} &
  {\color[HTML]{808080} 0.15} &
  \cellcolor[HTML]{C2EFCA}0.979 &
  {\color[HTML]{808080} 0.01} &
  \cellcolor[HTML]{CDECDE}10.239 &
  {\color[HTML]{808080} 0.45} &
  \cellcolor[HTML]{D9E9F6}\textbf{0.793} &
  {\color[HTML]{808080} 0.01} \\ \cline{2-27} 
 &
  \cellcolor[HTML]{D9D9D9} &
  \cellcolor[HTML]{D9D9D9}\(72   \times 72\) &
  \cellcolor[HTML]{C4F0CE}2.840 &
  {\color[HTML]{808080} 0.18} &
  \cellcolor[HTML]{C5EFD1}4.592 &
  {\color[HTML]{808080} 0.32} &
  \cellcolor[HTML]{C1F0C8}20.223 &
  {\color[HTML]{808080} 1.47} &
  \cellcolor[HTML]{C6EED1}0.430 &
  {\color[HTML]{808080} 0.04} &
  \cellcolor[HTML]{CBEDDB}9.602 &
  {\color[HTML]{808080} 0.45} &
  \cellcolor[HTML]{C4EFCD}0.702 &
  {\color[HTML]{808080} 0.01} &
  \cellcolor[HTML]{F4E4DD}15.504 &
  {\color[HTML]{808080} 0.91} &
  \cellcolor[HTML]{F8E3D8}23.990 &
  {\color[HTML]{808080} 1.69} &
  \cellcolor[HTML]{F7E3DA}20.019 &
  {\color[HTML]{808080} 1.28} &
  \cellcolor[HTML]{FBE2D5}0.095 &
  {\color[HTML]{808080} 0.05} &
  \cellcolor[HTML]{F6E3DA}-3.840 &
  {\color[HTML]{808080} 0.26} &
  \cellcolor[HTML]{F9E2D7}0.355 &
  {\color[HTML]{808080} 0.01} \\ \cline{3-27} 
 &
  \cellcolor[HTML]{D9D9D9} &
  \cellcolor[HTML]{D9D9D9}\(36 \times 36\) &
  \cellcolor[HTML]{C7EFD4}2.889 &
  {\color[HTML]{808080} 0.18} &
  \cellcolor[HTML]{C9EED7}\textbf{4.648} &
  {\color[HTML]{808080} 0.32} &
  \cellcolor[HTML]{C7EFD4}20.777 &
  {\color[HTML]{808080} 1.52} &
  \cellcolor[HTML]{C9EDD7}0.417 &
  {\color[HTML]{808080} 0.05} &
  \cellcolor[HTML]{C4EFCD}9.773 &
  {\color[HTML]{808080} 0.48} &
  \cellcolor[HTML]{C5EFCF}0.701 &
  {\color[HTML]{808080} 0.01} &
  \cellcolor[HTML]{F0E5E0}13.697 &
  {\color[HTML]{808080} 0.84} &
  \cellcolor[HTML]{F4E4DC}21.784 &
  {\color[HTML]{808080} 1.58} &
  \cellcolor[HTML]{F3E4DE}17.632 &
  {\color[HTML]{808080} 1.18} &
  \cellcolor[HTML]{F7E2D9}0.217 &
  {\color[HTML]{808080} 0.05} &
  \cellcolor[HTML]{F6E3DA}-3.688 &
  {\color[HTML]{808080} 0.27} &
  \cellcolor[HTML]{F8E2D8}0.360 &
  {\color[HTML]{808080} 0.01} \\ \cline{3-27} 
 &
  \multirow{-3}{*}{\cellcolor[HTML]{D9D9D9}\begin{tabular}[c]{@{}l@{}}RSP: Thermal\\ BVP: Thermal\end{tabular}} &
  \cellcolor[HTML]{D9D9D9}\(9 \times 9\) &
  \cellcolor[HTML]{D9EAF7}3.148 &
  {\color[HTML]{808080} 0.19} &
  \cellcolor[HTML]{D8EAF5}\textbf{4.930} &
  {\color[HTML]{808080} 0.33} &
  \cellcolor[HTML]{D7EAF2}\textbf{22.133} &
  {\color[HTML]{808080} 1.55} &
  \cellcolor[HTML]{D8E9F3}\textbf{0.358} &
  {\color[HTML]{808080} 0.05} &
  \cellcolor[HTML]{DAE9F8}9.219 &
  {\color[HTML]{808080} 0.45} &
  \cellcolor[HTML]{D9E9F5}0.687 &
  {\color[HTML]{808080} 0.01} &
  \cellcolor[HTML]{F3E4DD}15.352 &
  {\color[HTML]{808080} 0.92} &
  \cellcolor[HTML]{F8E3D8}24.068 &
  {\color[HTML]{808080} 1.69} &
  \cellcolor[HTML]{F6E3DA}19.699 &
  {\color[HTML]{808080} 1.27} &
  \cellcolor[HTML]{FBE2D5}0.115 &
  {\color[HTML]{808080} 0.05} &
  \cellcolor[HTML]{F6E3DA}-3.646 &
  {\color[HTML]{808080} 0.27} &
  \cellcolor[HTML]{F8E2D8}0.362 &
  {\color[HTML]{808080} 0.01} \\ \cline{2-27} 
 &
  \cellcolor[HTML]{BFBFBF} &
  \cellcolor[HTML]{BFBFBF}\(72   \times 72\) &
  \cellcolor[HTML]{CAEEDA}2.933 &
  {\color[HTML]{808080} 0.18} &
  \cellcolor[HTML]{CBEEDB}4.686 &
  {\color[HTML]{808080} 0.32} &
  \cellcolor[HTML]{C8EED6}20.856 &
  {\color[HTML]{808080} 1.50} &
  \cellcolor[HTML]{D0EBE5}0.389 &
  {\color[HTML]{808080} 0.05} &
  \cellcolor[HTML]{CEECE0}\textbf{9.536} &
  {\color[HTML]{808080} 0.45} &
  \cellcolor[HTML]{C2EFC9}\textbf{0.703} &
  {\color[HTML]{808080} 0.01} &
  \cellcolor[HTML]{D0ECE6}1.459 &
  {\color[HTML]{808080} 0.18} &
  \cellcolor[HTML]{D1ECE8}3.880 &
  {\color[HTML]{808080} 0.45} &
  \cellcolor[HTML]{D2ECE8}1.703 &
  {\color[HTML]{808080} 0.21} &
  \cellcolor[HTML]{D1EBE7}0.964 &
  {\color[HTML]{808080} 0.01} &
  \cellcolor[HTML]{C4EFCE}\textbf{10.292} &
  {\color[HTML]{808080} 0.45} &
  \cellcolor[HTML]{C3EFCC}\textbf{0.798} &
  {\color[HTML]{808080} 0.01} \\ \cline{3-27} 
 &
  \cellcolor[HTML]{BFBFBF} &
  \cellcolor[HTML]{BFBFBF}\(36 \times 36\) &
  \cellcolor[HTML]{C1F0C8}\textbf{2.790} &
  {\color[HTML]{808080} 0.17} &
  \cellcolor[HTML]{C1F0C8}\textbf{4.504} &
  {\color[HTML]{808080} 0.31} &
  \cellcolor[HTML]{C1F0C8}\textbf{20.185} &
  {\color[HTML]{808080} 1.49} &
  \cellcolor[HTML]{C1F0C8}\textbf{0.450} &
  {\color[HTML]{808080} 0.04} &
  \cellcolor[HTML]{C3EFCC}\textbf{9.788} &
  {\color[HTML]{808080} 0.46} &
  \cellcolor[HTML]{C4EFCD}0.702 &
  {\color[HTML]{808080} 0.01} &
  \cellcolor[HTML]{D3EBEB}1.496 &
  {\color[HTML]{808080} 0.19} &
  \cellcolor[HTML]{D6EAF2}4.162 &
  {\color[HTML]{808080} 0.49} &
  \cellcolor[HTML]{D2ECE9}1.711 &
  {\color[HTML]{808080} 0.22} &
  \cellcolor[HTML]{D7EAF2}0.959 &
  {\color[HTML]{808080} 0.01} &
  \cellcolor[HTML]{CBEDDA}10.250 &
  {\color[HTML]{808080} 0.44} &
  \cellcolor[HTML]{C9EDD7}0.796 &
  {\color[HTML]{808080} 0.01} \\ \cline{3-27} 
\multirow{-9}{*}{\begin{tabular}[c]{@{}l@{}}MMRPhys \\ with FSAM \dag\end{tabular}} &
  \multirow{-3}{*}{\cellcolor[HTML]{BFBFBF}\begin{tabular}[c]{@{}l@{}}RSP: Thermal\\ BVP: RGB\end{tabular}} &
  \cellcolor[HTML]{BFBFBF}\(9 \times 9\) &
  \cellcolor[HTML]{DAE9F8}3.154 &
  {\color[HTML]{808080} 0.19} &
  \cellcolor[HTML]{D9EAF7}4.943 &
  {\color[HTML]{808080} 0.33} &
  \cellcolor[HTML]{D8EAF4}22.232 &
  {\color[HTML]{808080} 1.55} &
  \cellcolor[HTML]{DAE9F8}\textbf{0.345} &
  {\color[HTML]{808080} 0.05} &
  \cellcolor[HTML]{DBE8F7}9.194 &
  {\color[HTML]{808080} 0.46} &
  \cellcolor[HTML]{DBE8F7}0.684 &
  {\color[HTML]{808080} 0.01} &
  \cellcolor[HTML]{C6EFD1}1.319 &
  {\color[HTML]{808080} 0.14} &
  \cellcolor[HTML]{C4EFCF}3.138 &
  {\color[HTML]{808080} 0.31} &
  \cellcolor[HTML]{C6EFD3}1.542 &
  {\color[HTML]{808080} 0.16} &
  \cellcolor[HTML]{C5EFCE}0.977 &
  {\color[HTML]{808080} 0.01} &
  \cellcolor[HTML]{DBE8F7}10.144 &
  {\color[HTML]{808080} 0.45} &
  \cellcolor[HTML]{DBE8F7}\textbf{0.793} &
  {\color[HTML]{808080} 0.01} \\ \hline
 &
  \cellcolor[HTML]{F2F2F2} &
  \cellcolor[HTML]{F2F2F2}\(72 \times   72\) &
  \cellcolor[HTML]{F5E4DB}4.892 &
  {\color[HTML]{808080} 0.20} &
  \cellcolor[HTML]{F4E4DC}6.294 &
  {\color[HTML]{808080} 0.35} &
  \cellcolor[HTML]{F2E4DE}29.379 &
  {\color[HTML]{808080} 1.32} &
  \cellcolor[HTML]{F4E3DC}0.066 &
  {\color[HTML]{808080} 0.05} &
  \cellcolor[HTML]{F9E2D7}4.014 &
  {\color[HTML]{808080} 0.37} &
  \cellcolor[HTML]{FAE2D6}0.547 &
  {\color[HTML]{808080} 0.01} &
  \cellcolor[HTML]{D3EBEB}1.499 &
  {\color[HTML]{808080} 0.19} &
  \cellcolor[HTML]{D5EBF0}4.105 &
  {\color[HTML]{808080} 0.48} &
  \cellcolor[HTML]{D3EBEC}1.728 &
  {\color[HTML]{808080} 0.21} &
  \cellcolor[HTML]{D6EAF0}0.960 &
  {\color[HTML]{808080} 0.01} &
  \cellcolor[HTML]{C4EFCD}10.295 &
  {\color[HTML]{808080} 0.45} &
  \cellcolor[HTML]{C1F0C8}\textbf{0.798} &
  {\color[HTML]{808080} 0.01} \\ \cline{3-27} 
 &
  \cellcolor[HTML]{F2F2F2} &
  \cellcolor[HTML]{F2F2F2}\(36 \times 36\) &
  \cellcolor[HTML]{F2E4DF}4.677 &
  {\color[HTML]{808080} 0.20} &
  \cellcolor[HTML]{F2E4DF}6.168 &
  {\color[HTML]{808080} 0.35} &
  \cellcolor[HTML]{EEE5E3}28.249 &
  {\color[HTML]{808080} 1.32} &
  \cellcolor[HTML]{F2E4DF}0.090 &
  {\color[HTML]{808080} 0.05} &
  \cellcolor[HTML]{F6E3DA}4.462 &
  {\color[HTML]{808080} 0.36} &
  \cellcolor[HTML]{F9E2D7}0.552 &
  {\color[HTML]{808080} 0.01} &
  \cellcolor[HTML]{DAE9F8}1.617 &
  {\color[HTML]{808080} 0.20} &
  \cellcolor[HTML]{DAE9F8}4.407 &
  {\color[HTML]{808080} 0.50} &
  \cellcolor[HTML]{DAE9F8}1.838 &
  {\color[HTML]{808080} 0.22} &
  \cellcolor[HTML]{DBE8F7}0.953 &
  {\color[HTML]{808080} 0.01} &
  \cellcolor[HTML]{CAEDDA}10.253 &
  {\color[HTML]{808080} 0.45} &
  \cellcolor[HTML]{C3EFCC}\textbf{0.798} &
  {\color[HTML]{808080} 0.01} \\ \cline{3-27} 
 &
  \multirow{-3}{*}{\cellcolor[HTML]{F2F2F2}\begin{tabular}[c]{@{}l@{}}RSP: RGB\\ BVP: RGB\end{tabular}} &
  \cellcolor[HTML]{F2F2F2}\(9 \times 9\) &
  \cellcolor[HTML]{E6E7EB}3.957 &
  {\color[HTML]{808080} 0.18} &
  \cellcolor[HTML]{E3E8EF}5.419 &
  {\color[HTML]{808080} 0.32} &
  \cellcolor[HTML]{E1E8F0}24.641 &
  {\color[HTML]{808080} 1.27} &
  \cellcolor[HTML]{E1E7F0}0.272 &
  {\color[HTML]{808080} 0.05} &
  \cellcolor[HTML]{F1E4DF}5.314 &
  {\color[HTML]{808080} 0.37} &
  \cellcolor[HTML]{F4E3DD}0.576 &
  {\color[HTML]{808080} 0.01} &
  \cellcolor[HTML]{C1F0C8}\textbf{1.248} &
  {\color[HTML]{808080} 0.13} &
  \cellcolor[HTML]{C1F0C8}2.936 &
  {\color[HTML]{808080} 0.27} &
  \cellcolor[HTML]{C1F0C8}\textbf{1.458} &
  {\color[HTML]{808080} 0.15} &
  \cellcolor[HTML]{C2EFC9}\textbf{0.980} &
  {\color[HTML]{808080} 0.01} &
  \cellcolor[HTML]{CAEDD9}10.254 &
  {\color[HTML]{808080} 0.45} &
  \cellcolor[HTML]{D1EBE7}\textbf{0.795} &
  {\color[HTML]{808080} 0.01} \\ \cline{2-27} 
 &
  \cellcolor[HTML]{D9D9D9} &
  \cellcolor[HTML]{D9D9D9}\(72   \times 72\) &
  \cellcolor[HTML]{C6EFD1}2.866 &
  {\color[HTML]{808080} 0.18} &
  \cellcolor[HTML]{CAEEDB}4.681 &
  {\color[HTML]{808080} 0.32} &
  \cellcolor[HTML]{C7EFD4}20.741 &
  {\color[HTML]{808080} 1.53} &
  \cellcolor[HTML]{CBEDDA}0.411 &
  {\color[HTML]{808080} 0.05} &
  \cellcolor[HTML]{CEECE0}9.529 &
  {\color[HTML]{808080} 0.45} &
  \cellcolor[HTML]{C3EFCB}0.702 &
  {\color[HTML]{808080} 0.01} &
  \cellcolor[HTML]{F4E4DC}15.776 &
  {\color[HTML]{808080} 0.94} &
  \cellcolor[HTML]{F9E3D7}24.730 &
  {\color[HTML]{808080} 1.72} &
  \cellcolor[HTML]{F8E3D8}20.618 &
  {\color[HTML]{808080} 1.35} &
  \cellcolor[HTML]{FBE2D6}0.121 &
  {\color[HTML]{808080} 0.05} &
  \cellcolor[HTML]{F6E3DA}-3.676 &
  {\color[HTML]{808080} 0.26} &
  \cellcolor[HTML]{F8E2D8}0.358 &
  {\color[HTML]{808080} 0.01} \\ \cline{3-27} 
 &
  \cellcolor[HTML]{D9D9D9} &
  \cellcolor[HTML]{D9D9D9}\(36 \times 36\) &
  \cellcolor[HTML]{CFECE3}3.000 &
  {\color[HTML]{808080} 0.18} &
  \cellcolor[HTML]{D0ECE6}4.789 &
  {\color[HTML]{808080} 0.33} &
  \cellcolor[HTML]{D2EBEA}21.773 &
  {\color[HTML]{808080} 1.57} &
  \cellcolor[HTML]{D1EBE6}0.385 &
  {\color[HTML]{808080} 0.05} &
  \cellcolor[HTML]{C7EED3}9.699 &
  {\color[HTML]{808080} 0.46} &
  \cellcolor[HTML]{C1F0C8}\textbf{0.703} &
  {\color[HTML]{808080} 0.01} &
  \cellcolor[HTML]{F0E5E0}13.620 &
  {\color[HTML]{808080} 0.82} &
  \cellcolor[HTML]{F4E4DC}21.498 &
  {\color[HTML]{808080} 1.53} &
  \cellcolor[HTML]{F2E4DE}17.393 &
  {\color[HTML]{808080} 1.14} &
  \cellcolor[HTML]{F9E2D7}0.165 &
  {\color[HTML]{808080} 0.05} &
  \cellcolor[HTML]{F6E3DB}-3.580 &
  {\color[HTML]{808080} 0.26} &
  \cellcolor[HTML]{F8E2D8}0.361 &
  {\color[HTML]{808080} 0.01} \\ \cline{3-27} 
 &
  \multirow{-3}{*}{\cellcolor[HTML]{D9D9D9}\begin{tabular}[c]{@{}l@{}}RSP: Thermal\\ BVP: Thermal\end{tabular}} &
  \cellcolor[HTML]{D9D9D9}\(9 \times 9\) &
  \cellcolor[HTML]{DAE9F8}\textbf{3.198} &
  {\color[HTML]{808080} 0.19} &
  \cellcolor[HTML]{DAE9F8}4.985 &
  {\color[HTML]{808080} 0.33} &
  \cellcolor[HTML]{DAE9F8}22.447 &
  {\color[HTML]{808080} 1.54} &
  \cellcolor[HTML]{DAE9F7}0.348 &
  {\color[HTML]{808080} 0.05} &
  \cellcolor[HTML]{D9E9F6}9.253 &
  {\color[HTML]{808080} 0.45} &
  \cellcolor[HTML]{D4EAEB}0.691 &
  {\color[HTML]{808080} 0.01} &
  \cellcolor[HTML]{F2E4DE}14.717 &
  {\color[HTML]{808080} 0.90} &
  \cellcolor[HTML]{F7E3D9}23.473 &
  {\color[HTML]{808080} 1.69} &
  \cellcolor[HTML]{F5E4DB}19.049 &
  {\color[HTML]{808080} 1.28} &
  \cellcolor[HTML]{F9E2D7}0.160 &
  {\color[HTML]{808080} 0.05} &
  \cellcolor[HTML]{F6E3DB}-3.550 &
  {\color[HTML]{808080} 0.27} &
  \cellcolor[HTML]{F8E2D8}0.364 &
  {\color[HTML]{808080} 0.01} \\ \cline{2-27} 
 &
  \cellcolor[HTML]{BFBFBF} &
  \cellcolor[HTML]{BFBFBF}\(72   \times 72\) &
  \cellcolor[HTML]{C8EED6}2.899 &
  {\color[HTML]{808080} 0.18} &
  \cellcolor[HTML]{CBEDDC}4.699 &
  {\color[HTML]{808080} 0.32} &
  \cellcolor[HTML]{C9EED7}20.900 &
  {\color[HTML]{808080} 1.54} &
  \cellcolor[HTML]{C4EFCD}0.439 &
  {\color[HTML]{808080} 0.04} &
  \cellcolor[HTML]{C1F0C8}9.837 &
  {\color[HTML]{808080} 0.48} &
  \cellcolor[HTML]{C2EFC9}\textbf{0.703} &
  {\color[HTML]{808080} 0.01} &
  \cellcolor[HTML]{C9EED8}\textbf{1.364} &
  {\color[HTML]{808080} 0.16} &
  \cellcolor[HTML]{CBEDDC}\textbf{3.533} &
  {\color[HTML]{808080} 0.40} &
  \cellcolor[HTML]{C7EFD5}\textbf{1.556} &
  {\color[HTML]{808080} 0.17} &
  \cellcolor[HTML]{CBEDDB}\textbf{0.970} &
  {\color[HTML]{808080} 0.01} &
  \cellcolor[HTML]{C1F0C8}\textbf{10.309} &
  {\color[HTML]{808080} 0.45} &
  \cellcolor[HTML]{C6EED1}\textbf{0.797} &
  {\color[HTML]{808080} 0.01} \\ \cline{3-27} 
 &
  \cellcolor[HTML]{BFBFBF} &
  \cellcolor[HTML]{BFBFBF}\(36 \times 36\) &
  \cellcolor[HTML]{CAEEDB}\textbf{2.934} &
  {\color[HTML]{808080} 0.18} &
  \cellcolor[HTML]{CBEEDC}4.693 &
  {\color[HTML]{808080} 0.32} &
  \cellcolor[HTML]{CCEDDD}\textbf{21.179} &
  {\color[HTML]{808080} 1.52} &
  \cellcolor[HTML]{C9EED6}0.420 &
  {\color[HTML]{808080} 0.04} &
  \cellcolor[HTML]{C7EED3}\textbf{9.700} &
  {\color[HTML]{808080} 0.45} &
  \cellcolor[HTML]{C3EFCC}\textbf{0.702} &
  {\color[HTML]{808080} 0.01} &
  \cellcolor[HTML]{DAE9F8}1.585 &
  {\color[HTML]{808080} 0.20} &
  \cellcolor[HTML]{DAE9F8}4.334 &
  {\color[HTML]{808080} 0.49} &
  \cellcolor[HTML]{DAE9F8}1.816 &
  {\color[HTML]{808080} 0.23} &
  \cellcolor[HTML]{DAE9F8}0.955 &
  {\color[HTML]{808080} 0.01} &
  \cellcolor[HTML]{CCEDDD}10.243 &
  {\color[HTML]{808080} 0.45} &
  \cellcolor[HTML]{CAEDD9}0.796 &
  {\color[HTML]{808080} 0.01} \\ \cline{3-27} 
\multirow{-9}{*}{\begin{tabular}[c]{@{}l@{}}MMRPhys\\ with \AM{}\end{tabular}} &
  \multirow{-3}{*}{\cellcolor[HTML]{BFBFBF}\begin{tabular}[c]{@{}l@{}}RSP: Thermal\\ BVP: RGB\end{tabular}} &
  \cellcolor[HTML]{BFBFBF}\(9 \times 9\) &
  \cellcolor[HTML]{D8EAF5}3.137 &
  {\color[HTML]{808080} 0.19} &
  \cellcolor[HTML]{DAE9F8}4.950 &
  {\color[HTML]{808080} 0.33} &
  \cellcolor[HTML]{DAE9F8}22.397 &
  {\color[HTML]{808080} 1.61} &
  \cellcolor[HTML]{DBE8F7}0.345 &
  {\color[HTML]{808080} 0.05} &
  \cellcolor[HTML]{D6EAF0}9.323 &
  {\color[HTML]{808080} 0.44} &
  \cellcolor[HTML]{DAE9F8}0.686 &
  {\color[HTML]{808080} 0.01} &
  \cellcolor[HTML]{C1F0C9}1.261 &
  {\color[HTML]{808080} 0.13} &
  \cellcolor[HTML]{C1F0C8}\textbf{2.923} &
  {\color[HTML]{808080} 0.27} &
  \cellcolor[HTML]{C2F0CB}1.483 &
  {\color[HTML]{808080} 0.15} &
  \cellcolor[HTML]{C1F0C8}\textbf{0.980} &
  {\color[HTML]{808080} 0.01} &
  \cellcolor[HTML]{DAE9F8}10.151 &
  {\color[HTML]{808080} 0.45} &
  \cellcolor[HTML]{DAE9F8}\textbf{0.793} &
  {\color[HTML]{808080} 0.01} \\ \hline \hline
\end{tabular*}
    \footnotesize
    \begin{flushleft}
    \dag FSAM: Factorized Self-Attention Module \cite{joshi2024factorizephys};  TSFM: Proposed Target Signal Constrained Factorization Module;  \\
    Avg: Average; SE: Standard Error; Cell color scale: Within each comparison group (delineated by double horizontal lines), values are highlighted using 3-color scale (green = best, orange = worst, blue = midpoint; Microsoft Excel conditional formatting). \\
    \end{flushleft}
\end{sidewaystable*}

Please see \cref{tab:results multi-task fold1}, \cref{tab:results multi-task fold2}, and \cref{tab:results multi-task fold3}.


{
\bibliography{tsfm}
\bibliographystyle{plain}
}

\end{document}